\def\paperID{14313}
\def\confName{ICCV}
\def\confYear{2025}

\def\paperTitle{GENCO --- A Unified Neural Solver Embedded in a Development Framework for Steady-State Grid Analysis}

\def\authorBlock{
\makebox[\textwidth][c]{%
\parbox{1.09\textwidth}{\centering
Alban~Puech\textsuperscript{*}$^{1,3}$,
Matteo~Mazzonelli$^{1}$,
Tamara~R.~Govindasamy$^{1}$,
Mangaliso~Mngomezulu$^{1}$,
Héctor~Maeso-García$^{1}$,
Thomas~Tolhurst$^{2}$,
Javad~Bayazi$^{2}$,
Ali~Moeini$^{2}$,
Naomi~Simumba$^{1}$,
Celia~Cintas$^{1}$,
David~Nelischer$^{1}$,
Romeo~Kienzler$^{1}$,
Jonas~Weiss$^{1}$,
Anna~Varbella$^{4}$,
Florian~Dörfler$^{5}$,
Gabriela~Hug$^{3}$,
Martin~Mevissen$^{1}$,
Juan~Bernabé-Moreno$^{1}$,
François~Mirallès$^{2}$,
Hendrik~F.~Hamann$^{6}$,
Etienne~Vos$^{1}$,
Thomas~Brunschwiler\textsuperscript{*}$^{1}$

\vspace{0.6em}

$^{1}$ IBM Research \\ 
$^{2}$ Hydro-Québec Research Institute, Canada \\
$^{3}$ Power Systems Laboratory, ETH Zurich, Switzerland \\
$^{4}$ Reliability and Risk Engineering Laboratory, ETH Zurich, Switzerland \\
$^{5}$ Automatic Control Laboratory, ETH Zurich, Switzerland \\
$^{6}$ Stony Brook University and Brookhaven National Laboratory, USA \\
}
}
}
\newif\ifreview 
\newif\ifarxiv \newcommand{\arxiv}{\arxivtrue}
\newif\ifcamera 
\newif\ifrebuttal 

\newcommand{\maxpfspeedups}{30$\times$\xspace}
\newcommand{\maxopfspeedups}{81$\times$\xspace}
\newcommand{\dcpfspeedupovergenco}{2$\times$\xspace}

\arxiv 

\pdfoutput=1
\documentclass[10pt,twocolumn,letterpaper]{article}
\ifreview \usepackage[review]{iccv} \fi
\ifarxiv \usepackage[pagenumbers]{iccv} \fi
\ifrebuttal \usepackage[rebuttal]{iccv} \fi
\ifcamera \usepackage{iccv} \fi

\usepackage{graphicx}
\usepackage{svg}
\usepackage{amsmath}	
\usepackage{amssymb}	
\usepackage{booktabs}
\usepackage{times}
\usepackage{microtype}
\usepackage{epsfig}
\usepackage{caption}
\usepackage{float}
\usepackage{placeins}
\usepackage{color, colortbl}
\usepackage{stfloats}
\usepackage{enumitem}
\usepackage{tabularx}
\usepackage{xstring}
\usepackage{multirow}
\usepackage{xspace}
\usepackage{url}
\usepackage{subcaption}
\usepackage{xcolor}
\usepackage[hang,flushmargin]{footmisc}
\usepackage{comment}
\usepackage{adjustbox}
\usepackage{pifont}
\usepackage{courier}

\usepackage[acronym,toc,nopostdot]{glossaries}
\makeglossaries

\newacronym{pf}{PF}{Power Flow}
\newacronym{opf}{OPF}{Optimal Power Flow}
\newacronym{se}{SE}{State Estimation}

\newacronym{gnn}{GNN}{Graph Neural Network}
\newacronym{fm}{FM}{Foundation Model}

\newacronym{scada}{SCADA}{Supervisory Control and Data Acquisition}

\ifcamera \usepackage[accsupp]{axessibility} \fi

% ...

\ifarxiv  \fi

\newcommand{\R}[1]{{%
    \textbf{%
        \ifstrequal{#1}{1}{\textcolor{red}{R#1}}{%
        \ifstrequal{#1}{2}{\textcolor{blue}{R#1}}{%
        \ifstrequal{#1}{3}{\textcolor{magenta}{R#1}}{%
        \ifstrequal{#1}{4}{\textcolor{teal}{R#1}}{%
                           \textcolor{cyan}{R#1}%
        }}}}%
    }%
}}

\newcolumntype{C}{>{\centering\arraybackslash}X}
\usepackage{makecell}

\usepackage{tikz}
\usetikzlibrary{arrows.meta,positioning,fit,calc}

\newcommand{\std}[1]{{\scriptsize$\pm${#1}}}
\newcommand{\datakit}{{gridfm-datakit}\xspace}
\newcommand{\graphkit}{{gridfm-graphkit}\xspace}
\newcommand{\genco}{{GENCO}\xspace}

\newcommand{\pfdelta}{PF$\Delta$\xspace}

\newcommand{\cmark}{\checkmark}
\newcommand{\xmark}{\ding{55}}

\definecolor{main}{HTML}{5989cf}    
\definecolor{sub}{HTML}{cde4ff}     
\usepackage{tcolorbox}
\tcbset{sharp corners, colback = white, before skip = 0.2cm, after skip = 0.5cm} 
\tcbuselibrary{breakable}
\newtcolorbox{boxH}{
    breakable,
    colback = sub, 
    colframe = main, 
    boxrule = 0pt, 
    leftrule = 6pt 
}

\newcommand{\G}[1]{\textbf{#1}}

\newcolumntype{Y}{>{\raggedright\arraybackslash}X} 

\usepackage{xr-hyper}

\makeatletter
\newcommand*{\addFileDependency}[1]{
  \typeout{(#1)}
  \@addtofilelist{#1}
  \IfFileExists{#1}{}{\typeout{No file #1.}}
}

\makeatother
\newcommand*{\myexternaldocument}[1]{
    \externaldocument{#1}
    \addFileDependency{#1.tex}
    \addFileDependency{#1.aux}
}

\definecolor{iccvblue}{rgb}{0.21,0.49,0.74}
\usepackage[pagebackref,breaklinks,colorlinks,allcolors=iccvblue]{hyperref}
\usepackage[capitalize]{cleveref}
\crefname{section}{Sec.}{Secs.}
\crefname{table}{Table}{Tables}
\crefname{figure}{Fig.}{Figs.}

\ifarxiv \crefname{appendix}{App.}{Apps.}
\else \crefname{appendix}{Suppl.}{Suppls.} \fi

\unless\ifarxiv \myexternaldocument{_supplementary} \fi

\usepackage[table]{xcolor}
\usepackage{etoolbox}
\usepackage{pgf}

\definecolor{high}{HTML}{76f013}  
\definecolor{low}{HTML}{ec462e}  

\begin{document}

\title{\paperTitle}
\author{\authorBlock}

\twocolumn[{
\maketitle

\begin{center}
\includegraphics[width=0.90\linewidth]{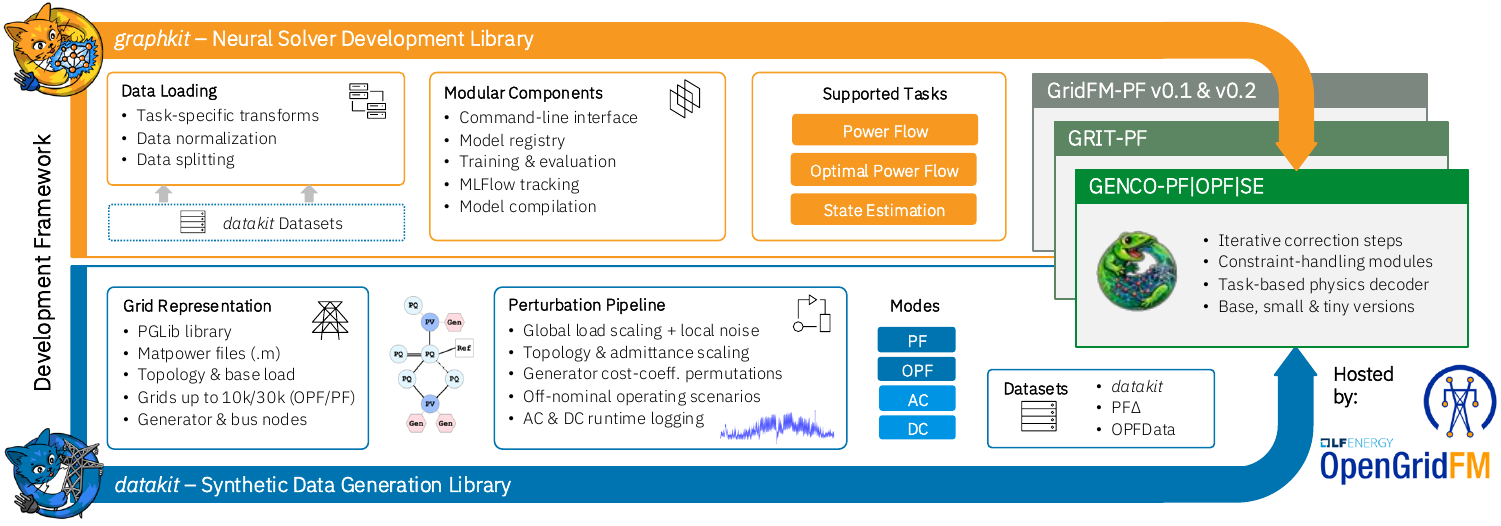}
\captionof{figure}{Representation of \textcolor[HTML]{209138}{\genco}, our GEometric Neural Corrective Optimizer, within the GridFM development framework composed of \textcolor[HTML]{F79B25}{\graphkit} for neural solver training and evaluation, and \textcolor[HTML]{2596BE}{\datakit} for synthetic data generation. All three are released within the Linux Foundation for Energy OpenGridFM project and are available on \href{https://github.com/gridfm}{GitHub}. All models and datasets are available on \href{https://huggingface.co/gridfm}{Hugging Face}.}
\label{fig:motivational_example}
\end{center}

}]

\begingroup
\renewcommand{\thefootnote}{*}
\footnotetext{Corresponding authors: \texttt{\{apu,tbr\}@zurich.ibm.com}}
\endgroup
\setcounter{footnote}{0}
\renewcommand{\thefootnote}{\arabic{footnote}}

\begin{abstract}

\vspace{-0.4em}

Foundation models are transforming business workflows and boosting productivity, yet they remain noticeably absent from some engineering domains such as power system analysis, where strict physical consistency must be enforced.\\

We present \genco (GEometric Neural Corrective Optimizer), a unified neural solver for steady-state transmission grid analysis that handles power flow (PF), optimal power flow (OPF), and state estimation (SE) within a single architecture, eliminating task-specific pipelines by operating on a shared grid representation. To support advances in neural power system solvers, we introduce the open-source GridFM Development Framework (\cref{fig:motivational_example}) that standardizes synthetic data generation and training in a low-code environment. We also release large-scale datasets with millions of PF and OPF scenarios across diverse grid topologies to further support reproducible benchmarking.\\

We evaluate \genco on the latest benchmark datasets (\pfdelta, OPFData), against all current state-of-the-art neural solvers, as well as classical solvers (Newton–Raphson, IPOPT) and real-world Hydro-Québec SCADA data. For large-scale PF, \genco recovers the full AC operating state---including voltage magnitudes and reactive power that DC-PF cannot provide---while matching DC-PF-level active power-balance residuals, with up to \maxpfspeedups speedups over Newton–Raphson at only \dcpfspeedupovergenco the runtime of DC-PF. For OPF, it achieves up to \maxopfspeedups speedups over IPOPT while improving feasibility, optimality, and runtime over DC-OPF. For SE, it is more robust than classical weighted least squares (WLS) to noisy measurements and grid parameters, and it always returns a high-quality estimate even when WLS fails to converge.

Together, the unified architecture and development framework offer a new way to perform large-scale steady-state grid analysis, reduce integration effort, and lower the barrier to entry for power system engineers, marking a key step toward Grid Foundation Models.

\vspace{-5pt}

\end{abstract}
\section{Introduction}
\label{sec:intro}


The growing penetration of variable energy resources, together with changing demand patterns, increases grid uncertainty, broadening the range of operating conditions considered in planning studies and encountered in real-time operations~\cite{doe2024national}. Across these settings, combinations of generation, demand, contingencies, network configurations, and control decisions create a combinatorially large space of potential grid states. Analyzing, optimizing, and monitoring these states require repeated power flow (PF), optimal power flow (OPF), and state estimation (SE) calculations, respectively, placing substantial computational demands on conventional solvers and making efficiency a central challenge~\cite{whitehouse2022climategoals,doe2022cleanelectricity}. Classical AC solvers (e.g., based on Newton-Raphson for PF and interior-point methods for OPF) provide reliable solutions but become computationally expensive at scale. Thus, DC approximations are often used, but sacrifice solution completeness and physical consistency.
\\\\
In this context, machine learning has emerged as a promising approach for building surrogate models of power systems~\cite{benes2024ai,daniel2024ai,smartcities4020029,cremer2024pioneeringroadmapmldrivenalgorithmic,11371605} that can learn complete AC solutions and provide significant speedups. Recent work has demonstrated encouraging gains in accelerating PF and OPF through so-called neural solvers~\cite{DONON2020106547,gnnreviewpowergrid,Khaloie}, particularly using graph neural network-based architectures~\cite{gnnreview,Wu_2021}. However, their practical use remains limited. Existing approaches are often task-specific, restricted to fixed grid sizes or problem formulations, and evaluated on synthetic datasets that do not fully reflect operational variability. For instance, neural PF solvers may degrade under distribution shifts when trained primarily on OPF-feasible states~\cite{lin2024powerflownet,DONON2020106547, NEURIPS2025_d000ef56}, and current neural OPF solvers only rely on fixed generator cost coefficients~\cite{lovett2024opfdatalargescaledatasetsac,opflearn,pglearn,varbella2024powergraph}. In SE, neural approaches remain comparatively underexplored, and achieving robustness to noisy data, missing measurements, and low observability continues to be a key challenge~\cite{10408467,BHUSAL2021106806}.\\

A further limitation is fragmentation. New architectures are trained and tested on custom datasets, specific grids, and different evaluation metrics. This makes comparisons difficult, limits reproducibility, and slows down innovation cycles. Runtime studies are often inconsistent, varying in solver settings, implementations, and hardware, and often compare highly parallelized GPU-accelerated neural solvers against classical solvers running on a single CPU core~\cite{piloto2024, arowolo2025,varbella2024physicsinformedgnnnonlinearconstrained}. Moreover, most neural solvers are developed separately for PF, OPF, or SE, although they share the same network topology, electrical variables, and constraints. This prevents reuse across tasks and limits the transfer of learned grid representations. In addition, limited tooling exists to develop neural solvers, requiring expert knowledge in computer science and power systems, which slows the uptake of the technology.

\subsection*{Contributions}

Motivated by the fragmentation of existing approaches and the transformative impact of foundation models in building broad and fast surrogates for complex systems~\cite{bommasani2021opportunities,mukkavilli2023aifoundationmodelsweather,jakubik2023foundation}, we previously outlined a vision for an AI foundation for the electric power grid~\cite{hamann2024foundation}. This paper represents a first concrete step toward this vision by introducing a unified architecture that supports multiple grid analysis tasks, demonstrates robustness to distribution shifts, and achieves strong data efficiency. This lays the foundation for future Grid Foundation Models (GridFMs) capable of adapting to new tasks and grid topologies without finetuning. \\

Overall, we report our advances in steady-state grid analysis built around two complementary contributions:

\begin{itemize}
    \item \textbf{\genco (GEometric Neural Corrective Optimizer)}, a unified neural solver architecture as a common backbone, that operates on a shared grid representation and scales to large grids of up to 10k buses. It eliminates the need for separate task-specific models, as it supports three core tasks:

    \begin{itemize}
        \item \textit{Power Flow}, including off-nominal operating conditions and high-order contingencies.
        \item \textit{Optimal Power Flow}, supporting variable generator costs.
        \item \textit{State Estimation}, robust to partial observability and inaccurate grid parameters.
    \end{itemize}
    

    \item \textbf{GridFM Development Framework} (\cref{fig:motivational_example}), released under the Apache 2.0 license via the LF Energy \mbox{OpenGridFM} project, designed to enable scalable development and reproducible experimentation with a low barrier to entry. It integrates tools for synthetic data generation, model development, and standardized evaluation in a low-code environment, and consists of:
    \begin{itemize}
        \item \textit{\datakit}, for generating realistic PF/OPF datasets with diverse load and topology perturbations,
        \item \textit{\graphkit}, for training and evaluating neural solvers such as \genco,
        \item \textit{Open datasets} containing 4 million PF/OPF instances across 8 grid topologies, hosted on Hugging Face.
    \end{itemize}
\end{itemize}

Building on this framework, we further evaluate \genco and introduce a unified benchmarking and analysis pipeline:

\begin{itemize}
    \item \textbf{Comprehensive performance, runtime, and scalability analysis} against state-of-the-art neural solvers (PF: GNS~\cite{DONON2020106547}, CANOS-PF~\cite{NEURIPS2025_d000ef56}, PFNet~\cite{lin2024powerflownet}; OPF: HH-MPNN~\cite{arowolo2025}) and classical AC/DC solvers (PowerModels~\cite{powermodels}) on \pfdelta~\cite{NEURIPS2025_d000ef56} and OPFData~\cite{lovett2024opfdatalargescaledatasetsac} benchmarks. 
    \item \textbf{Validation on real-world data} using SCADA measurements from the Hydro-Québec grid. We demonstrate that \genco can leverage synthetic pretraining and limited real-world finetuning to compute complete AC solutions on real-world data.
\end{itemize}

\section{Related Work \& Challenges}

\label{sec:related}

The growing use of machine learning for steady-state grid analysis has produced promising neural solvers for Power Flow (PF), Optimal Power Flow (OPF), and State Estimation (SE). Our work is motivated by four outstanding challenges: (i) unifying grid representations and model architecture across tasks, (ii) balancing physical feasibility with runtime and architectural efficiencies, (iii) improving robustness to realistic grid variations, and (iv) standardizing datasets, metrics, and workflows.

\subsection{From Task-Specific Solvers to Unified Grid Representations and Architectures}

Existing neural solvers are task-specific, which hinders the transfer of innovation; e.g., CANOS~\cite{piloto2024} pioneered the use of supervised models for large-scale OPF, but adapting it to PF required a separate redesign in \pfdelta~\cite{NEURIPS2025_d000ef56}; GNS~\cite{DONON2020106547} introduced message passing as iterative corrections for PF, but extending it to cost minimization for OPF remains open.\\

The three core steady-state tasks considered here (PF, OPF, SE) differ in formulation: PF computes bus voltages and branch flows from injections~\cite{4499318}; OPF optimizes generator setpoints given loads while considering PF equations and operational constraints~\cite{pandya2008,low2014, cain2012}; SE infers system states from sparse measurements~\cite{abur2004power}. All share electrical state variables, grid parameters, and AC physics; yet, current methods use task-specific representations that obscure this structure: PF models use bus-level graphs~\cite{powermodels, lin2024powerflownet}, OPF requires generator-level resolution~\cite{piloto2024}, and SE works with net injections rather than distinguishing generation from load~\cite{10408467}.\\

The consequences are significant. Developing neural solvers is largely an empirical process. Task-specific designs require this effort to be repeated from scratch for each problem, and advances for one task do not transfer to others. More fundamentally, no existing architecture can process the full range of inputs required across tasks and simultaneously handle cost minimization, constraint satisfaction, and robustness to noise -- the three distinct computational demands that arise across tasks. 

\subsection{Feasibility--Optimality--Runtime Tradeoffs}

A central challenge for neural grid solvers is the trade-off between runtime and solution feasibility, and in OPF, optimality. The goal of neural solvers is to reduce computation time by replacing iterative procedures (e.g., Newton–Raphson with roughly quadratic scaling in grid size~\cite{4596138}, and interior-point methods for OPF with higher-order complexity~\cite{cubic}) with a learned direct mapping from inputs to outputs. However, achieving fast inference (ideally linear in graph size) constrains neural solvers' capacity to learn this mapping, which can reduce prediction accuracy and amplify small prediction errors into degraded feasibility and optimality. \\

Existing methods address this through integration of physics and constraints into the model. For PF, GNN-based models evolve from directly regressing numerical solver outputs~\cite{lin2024powerflownet} toward physics-informed architectures, adding power-balance residuals in the loss~\cite{LOPEZGARCIA2023105567}, or predicting only subsets of variables while reconstructing the remaining states through power-flow equations~\cite{DONON2020106547}. These approaches improve physical consistency, but feasibility remains encouraged rather than guaranteed. For OPF, purely regression-based methods have also been proposed~\cite{arowolo2025}, but soft-constrained approaches have become increasingly prevalent, promoting feasibility through penalty terms in the loss function\footnote{Optimality is encoded either in a supervised way, matching pre-computed solutions~\cite{piloto2024}, or in a self-supervised one, minimizing generation costs~\cite{dc3}.}. These penalties may be combined with post-processing corrections to restore feasibility for selected constraints~\cite{piloto2024}, but are sensitive to penalty tuning and often require extensive hyperparameter search\footnote{Primal-dual~\cite{primaldual} and augmented-Lagrangian~\cite{lumina2026} methods adapt penalties dynamically but introduce more complex and potentially unstable training.}. In contrast, hard-constrained architectures enforce feasibility structurally, e.g., via projection or feasibility-seeking layers that map predictions onto the constraint manifold~\cite{NEURIPS2025_3874e2be}. While this guarantees feasibility, it also introduces higher computational costs and greater implementation complexity.\\

This feasibility--runtime trade-off is closely tied to the model architecture. Standard message-passing GNNs face a locality bottleneck: information is exchanged only between neighboring nodes at each layer, so that capturing the influence of electrically distant buses requires deeper models to increase the receptive field, which can suffer from oversmoothing and oversquashing~\cite{gnnreview, liu2022graph}, while also increasing inference cost. Model architectures with a global view, such as full-attention Transformers~\cite{vaswani2017attention, gps} reduce this locality limitation, but scale quadratically with grid size. Sparse, sampled, or hierarchical aggregation schemes improve efficiency, but introduce additional modeling choices and hyperparameters. Overall, neural grid solvers must balance feasibility, optimality, and runtime under the constraints imposed by information propagation.

\subsection{Generalization to Diverse Scenarios}

Robustness and generalization remain a central challenge for neural PF, OPF, and SE solvers. Distribution shifts in load, topology, grid parameters, and measurements can substantially degrade performance outside the training regime~\cite{NEURIPS2025_d000ef56}, unlike classical methods whose limitations stem primarily from numerical issues~\cite{tostado2021solving}.
\\

Neural solvers are largely trained on synthetic datasets with limited variation in operating conditions\footnote{Generating sufficiently large and diverse datasets is computationally expensive due to the need to repeatedly solve PF, OPF, or SE problems. For example, PGLearn required 1245 CPU Core-days for data generation~\cite{pglearn}.}. These existing datasets typically rely on uncorrelated random load perturbations around nominal operating points~\cite{lovett2024opfdatalargescaledatasetsac,varbella2024powergraph,lin2024powerflownet,LOPEZGARCIA2023105567,DONON2020106547}, fixed grid parameters~\cite{opflearn,NEURIPS2025_d000ef56,pglearn,lovett2024opfdatalargescaledatasetsac,varbella2024powergraph}, and limited topology modifications such as N-1/N-2 contingencies. All aforementioned OPF datasets keep generator cost coefficients fixed, effectively restricting learning to a single economic regime with unchanged merit order. Similarly, PF datasets, except \pfdelta, sample from OPF-feasible operating points, limiting exposure to stressed or out-of-nominal operating conditions. Constructing SE datasets is even more challenging because it additionally requires modeling sensor placement, noise, missing data, and bad-data patterns. \\

Recent datasets address some of these limitations: OPFData~\cite{lovett2024opfdatalargescaledatasetsac} and PGLearn~\cite{pglearn} were the first to scale to systems with more than 10k buses; OPF-Learn~\cite{opflearn}, PGLearn~\cite{pglearn}, and PowerGraph~\cite{varbella2024powergraph} introduced more diverse load sampling; and \pfdelta~\cite{NEURIPS2025_d000ef56} extends beyond OPF-feasible operating regions. However, these advances remain fragmented across many different data-generation libraries, which are additionally task-specific and not designed for generating the large-scale datasets required to train foundational models.\\

The importance of training data diversity has been reported in the literature; e.g., \pfdelta~\cite{NEURIPS2025_d000ef56} demonstrates improved generalization under N-1 training, yet generalization to unseen networks remains an open challenge~\cite{yang2026gridsfm,arowolo2025,DONON2020106547}. Despite this, motivations to improve dataset realism remain limited, as more challenging scenarios expose model weaknesses and often reduce reported performance. As a consequence, in the absence of standardized benchmarks, results remain difficult to compare across studies. Finally, another key limitation is the lack of validation between synthetic and real-world performance. While synthetic grid topologies used for data generation capture statistical properties of real network topologies, there is no established measure of (i) similarity between synthetic and real operating conditions, or (ii) transferability of performance from synthetic benchmarks to real grid deployment.

\subsection{Fragmented Benchmarking and Software Workflows}

Despite rapid progress, neural solver development remains fragmented. Beyond differences in data representations (bus- vs. element-level), architectures, and data-generation assumptions, methods often use incompatible input formats and features (e.g., bus-admittance matrices~\cite{varbella2024powergraph} vs. branch-level quantities~\cite{arowolo2025}). They further differ in their output representations and unit conventions, making direct comparisons difficult even for identical tasks.\\

Evaluation protocols are similarly inconsistent. Studies report heterogeneous metrics for feasibility (constraint violation magnitudes~\cite{arowolo2025, NEURIPS2025_3874e2be}, thresholded violation rates~\cite{piloto2024}) and optimality, and often rely on non-physically interpretable regression metrics (e.g., mean squared error)~\cite{varbella2024powergraph}. These are further computed with varying aggregation schemes (mean vs. max) and at different granularities (bus- vs. graph-level), making comparisons sensitive to implementation choices rather than model performance alone.\\

These issues are particularly pronounced in runtime benchmarking against classical solvers. Reported speedups depend on hardware, batch size, parallelization strategy, solver implementation, and processes included in the runtime analysis (e.g., data transfer, preprocessing, graph construction). Existing comparisons contrast highly parallelized neural solvers running on GPUs with single-threaded Python or MATLAB implementations (e.g., pandapower~\cite{pandapower} or MATPOWER~\cite{matpower}) that underutilize CPU hardware~\cite{piloto2024, NEURIPS2025_d000ef56, arowolo2025}. Without standardized evaluation protocols, runtime comparisons between neural and classical AC/DC solvers remain difficult to interpret fairly. These gaps motivate a unified software stack that standardizes data loading, graph construction, feature normalization, solver baselines, feasibility metrics, and runtime measurement protocols across PF, OPF, and SE.\\

\section{Datasets \& Data Generation}
\label{sec:dataset}

In this section, we present the datasets and data-generation frameworks used throughout the paper. As a key contribution, we introduce \datakit (\cref{subsec:datakit}), a publicly available framework to generate the most diverse, realistic, and scalable synthetic datasets for Power Flow (PF), Optimal Power Flow (OPF), and State Estimation (SE). Additionally, we describe two widely used benchmark datasets for PF and OPF (\cref{subsec:benchmark_data}), which we use to compare \genco against existing state-of-the-art models under standard evaluation settings in \cref{subsec:pf} and \cref{subsec:opf}. Finally, we compare the diversity of \datakit-generated data against real SCADA data from Hydro-Québec in \cref{subsec:comparison_data_hq}.

\subsection{Data Generation with \datakit}
\label{subsec:datakit}
We utilize \datakit to generate large PF and OPF datasets. \datakit unifies and advances state-of-the-art synthetic grid data generation methods, uniquely generating realistic off-nominal operating condition samples for PF and diverse generator costs for OPF. We here provide a short overview of the main features of \datakit. For a detailed technical description of the pipeline, runtime analysis, and exhaustive comparisons against other libraries (summarized in \cref{tab:comparison}), we refer to the \datakit technical report~\cite{puech2025gridfmdatakitv1pythonlibraryscalable}.

\begin{table*}[t]
\centering
\small
\resizebox{\textwidth}{!}{
\begin{tabular}{l c c c c c c c c}
& & \multicolumn{3}{c}{Load Variations} & & & & \\
\cmidrule(lr){3-5}
Library & Grid Size &
\makecell{Spatial\\Correlation} &
\makecell{from Real\\Profiles} &
Diverse$^*$ &
\makecell{Higher-order N-k ($k>2$)\\Topology Variations} &
\makecell{Admittance\\Variations} &
\makecell{Generator Profile\\Variations} &
\makecell{Off-Nominal\\Operating Scenarios} \\
\midrule

\multicolumn{9}{c}{\textbf{Power Flow (PF) Datasets}} \\
\midrule
\datakit-pf~\cite{puech2025gridfmdatakitv1pythonlibraryscalable} & 30K & \cmark & \cmark & \cmark & \cmark & \cmark & \cmark & \cmark \\
\pfdelta~\cite{NEURIPS2025_d000ef56} & 2K & \xmark & \xmark & \cmark & \xmark & \xmark & \cmark & \cmark \\
PFNet~\cite{lin2024powerflownet} & 6K & \cmark & \xmark & \xmark & \xmark & \cmark & \xmark & \xmark \\
\midrule

\multicolumn{9}{c}{\textbf{Optimal Power Flow (OPF) Datasets}} \\
\midrule
\datakit-opf~\cite{puech2025gridfmdatakitv1pythonlibraryscalable} & 10K & \cmark & \cmark & \cmark & \cmark & \cmark & \cmark & --- \\
OPFData (CANOS)~\cite{piloto2024, lovett2024opfdatalargescaledatasetsac} & 14K & \cmark & \xmark & \xmark & \xmark & \xmark & \xmark & --- \\
OPF-Learn~\cite{opflearn} & 118 & \xmark & \xmark & \cmark & \xmark & \xmark & \xmark & --- \\
PGLearn~\cite{pglearn} & 24K & \cmark & \xmark & \cmark & \xmark & \xmark & \xmark & --- \\
PowerGraph~\cite{varbella2024powergraph} & 118 & \cmark & \cmark & \xmark & \xmark & \xmark & \xmark & --- \\
\bottomrule
\end{tabular}}
\caption{Comparison of PF and OPF data generation libraries. $^*$Diverse load variations denote load scenarios that extend beyond uniform perturbations around nominal operating points and beyond simple global scaling of all loads. OPF datasets do not contain off-nominal scenarios, as OPF solutions have to satisfy all operational constraints (e.g., thermal and voltage limits).}
\label{tab:comparison}
\end{table*}

\subsubsection{Grid Support and Scalability}
\datakit uses PowerModels.jl~\cite{powermodels} with distributed Julia~\cite{julia} runtimes through JuliaCall~\cite{juliacall} to scale to networks of up to 30{,}000 buses for PF and 10{,}000 buses for OPF, supporting MATPOWER~\cite{matpower} (e.g., all PGLib grids~\cite{pglib}), PSS$^{\circledR}$E~\cite{siemens_psse_v34}, and ENTSO-E's CGMES~\cite{entsoe_cim_network_model_management_guide} formats. Importantly, the library can be used to generate data for real grids, as we demonstrate for the $\sim$1{,}200-bus Hydro-Qu\'ebec (HQ1200) network.\\

The generation is highly parallelized; as shown in \cref{tab:scalability}, generating $\sim$200,000 samples for the IEEE 118-bus system requires under 20 minutes for PF and roughly 2 hours for OPF\footnote{using 20 cores of an AMD EPYC 7763 CPU @ 2.45 GHz.}. Moreover, the convergence rate exceeds 97\% across all grid sizes for PF thanks to the use of robust solvers and our load sampling technique, compared to approximately 75\%~\cite{hitandrun} for methods utilizing hit-and-run sampling that uniformly explore the feasible load space, such as OPF-Learn~\cite{opflearn} and \pfdelta~\cite{NEURIPS2025_d000ef56}.
\begin{table}
\centering
\caption{Time required to generate approximately 200,000 samples with N-1 topology perturbations for different grid sizes. The reported number of samples corresponds to successfully converged samples. Refer to~\cite{puech2025gridfmdatakitv1pythonlibraryscalable} for details on runtimes.}\label{tab:scalability}
\resizebox{1.0\linewidth}{!}{%
\begin{tabular}{lccc}
\hline
Grid name & \makecell{Number of\\converged samples [-]} & 
\makecell{CPU\\Core-hours [-]} & 
\makecell{Convergence\\rate [\%]} \\
\hline
\multicolumn{4}{c}{\textbf{Power Flow (PF)}} \\
\hline
IEEE 24 bus  & 199,540  & 2.69      & 99.77     \\
IEEE 118 bus & 199,339  & 6.44      & 99.67     \\
GOC 2k bus    & 198,858  & 247.55    & 99.43     \\
GOC 10k bus   & 199,880  & 1,384.24  & 99.94     \\
\hline
\multicolumn{4}{c}{\textbf{Optimal Power Flow (OPF)}} \\
\hline
IEEE 24 bus  & 190,387  & 21.33     & 95.19     \\
IEEE 118 bus & 197,769  & 46.10     & 98.88     \\
GOC 2k bus    & 198,308  & 1,103.67  & 99.15     \\
GOC 10k bus   & 195,920  & 3,628.01  & 97.96     \\
\hline
\end{tabular}}
\end{table}

\subsubsection{Load Scenarios}
Public grid datasets rarely provide bus-level load trajectories due to the sensitive nature of operational data. Consequently, synthetic datasets typically rely on a single nominal operating point and generate scenarios through artificial perturbations. Existing approaches either randomly perturb the base load independently at each bus~\cite{lin2024powerflownet, piloto2024, lovett2024opfdatalargescaledatasetsac}, or sample loads randomly in the feasible space~\cite{NEURIPS2025_d000ef56,opflearn}, which may fail to preserve realistic spatial and temporal load correlations. To address these limitations, we implement a hybrid perturbation strategy that combines realistic system-level demand evolution with local bus-level diversity. At time $t$, a global scaling factor \texttt{ref}$_t$ is derived from real aggregated profiles provided in \datakit, e.g., ERCOT's 2025 hourly system-level load published by the EIA~\cite{EIA_ERCOT_2025}, or other user-provided aggregated load profiles. For nominal loads $p_i$ and $q_i$ at bus $i$, the perturbed load is:
\begin{equation}
\label{eq:scaling}
    \tilde{p}_{i,t} = p_i \cdot \texttt{ref}_t \cdot \epsilon^p_{i,t}, \quad \tilde{q}_{i,t} = q_i \cdot \texttt{ref}_t \cdot \epsilon^q_{i,t}
\end{equation}
where $\epsilon^p_{i,t}, \epsilon^q_{i,t} \sim \mathcal{U}(1 - \sigma, 1 + \sigma)$ are independent local perturbation factors sampled from a uniform distribution over $[1-\sigma,1+\sigma]$, introducing per-bus variability, while $\texttt{ref}_t$ captures the temporal evolution of the aggregated demand.

\subsubsection{Topology and Admittance Perturbations}
Unlike all other libraries that are limited to N-1~\cite{lin2024powerflownet, lovett2024opfdatalargescaledatasetsac, piloto2024, opflearn, pglearn, varbella2024powergraph} or up to N-2~\cite{NEURIPS2025_d000ef56} contingencies, \datakit supports arbitrary N-k topology perturbations through either exhaustive enumeration or rejection sampling (to avoid islanding). Admittance variations are introduced by scaling branch resistances and reactances by random factors sampled from a uniform distribution.

\subsubsection{Generator Setpoints and Data Modes}
\label{subsubsec:gen_setpoints}

\paragraph{OPF mode.}
Generator setpoints are computed by solving AC-OPF on the perturbed topology. To ensure generalizability across market conditions, \datakit permutes or randomly scales generator cost coefficients prior to solving, introducing dispatch diversity, absent in fixed-cost libraries.

\paragraph{PF mode.}
Base generator setpoints are obtained by solving AC-OPF on the base topology (without perturbations, but accounting for load, admittance, and generator cost variations). Then, N-k topology perturbations are applied, and AC-PF is solved \textit{without} re-optimizing dispatch, producing realistic off-nominal operating states that may violate OPF constraints (e.g., line overloads, voltage limits). This yields a balanced mix of points within and outside normal operating limits, reflecting realistic system behavior where OPF determines generator dispatch under nominal conditions and unexpected changes may lead to violations.

\subsection{Benchmark Datasets}
\label{subsec:benchmark_data}
To benchmark \genco against existing state-of-the-art neural solvers (\cref{subsec:pf} and~\ref{subsec:opf}), we additionally leverage two widely used benchmark datasets: \pfdelta~\cite{NEURIPS2025_d000ef56} for PF and OPFData~\cite{lovett2024opfdatalargescaledatasetsac} for OPF. While these datasets enable direct comparison with existing neural solvers under standard evaluation settings, they remain limited in several important aspects. In particular, they do not provide DC-PF or DC-OPF solutions, preventing direct comparison with classical DC solvers under the same evaluation settings. More importantly, their limited scenario diversity prevents studying model generalization under diverse topology perturbations, varying generator costs, contingency scenarios, or admittance perturbations. Therefore, \datakit, which addresses these limitations, is used for all other experiments.
\\\\
\textbf{\pfdelta}~\cite{NEURIPS2025_d000ef56} is a dataset of solved PF problems with load scenarios sampled via hit-and-run to explore the feasible load space, permuted generator costs, N-1 and N-2 topology perturbations, and scenarios generated by solving OPF excluding selected inequality constraints to obtain off-nominal operating conditions. Unlike \datakit, it does not offer real load profiles, admittance variations, or higher-order topology perturbations; but it enables comparison against state-of-the-art PF models (GNS~\cite{DONON2020106547}, CANOS-PF~\cite{piloto2024}, PFNet~\cite{lin2024powerflownet}).
\\\\
\textbf{OPFData} is a widely used~\cite{arowolo2025, hedgeopf2025,piloto2024, lumina2026, yang2026gridsfm} dataset for OPF with N-1 topology perturbations. Despite being one of the largest publicly available OPF datasets, it has various limitations: (i) load scenarios are generated by independently scaling each load with a uniform factor in $[0.8, 1.2]$, limiting load diversity and realism; and (ii) fixed generator cost parameters, which restrict generator setpoint diversity by effectively forcing low-cost generators to operate at capacity across most scenarios. However, it enables direct comparison of \genco with HH-MPNN~\cite{arowolo2025}, the currently best-performing open-source model using this dataset.

\subsection{Validation of Synthetic Data Against Hydro-Québec Real-World SCADA Dataset}
\label{subsec:comparison_data_hq}

\begin{figure}
\centering
\includegraphics[width=0.7\linewidth]{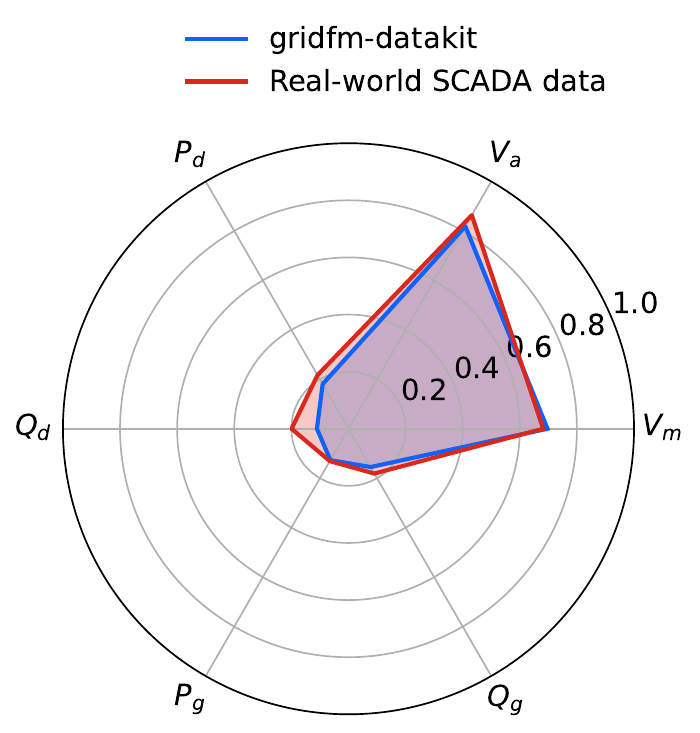}
\caption{Normalized mean feature entropy of real-world SCADA PF data from HQ1200 and synthetic PF data generated for HQ1200 using \datakit. Values close to one indicate higher diversity, with one corresponding to a uniform distribution. Features are: $V_m$/$V_a$: bus voltage magnitudes/angles; $P_g$/$Q_g$: active/reactive power generation; $P_d$/$Q_d$: active/reactive power demand.} 
\label{fig:spider_hq1200}
\end{figure}

A key trade-off in data generation lies between covering a broad range of operating conditions and avoiding unrealistic samples that can destabilize training or distract from the regimes that matter. To assess the diversity of the data generated with \datakit, we compare it against real-world SCADA measurements from the Hydro-Québec (HQ) transmission network\footnote{Beyond the validation against real-world SCADA data presented here, a broader data diversity comparison with existing PF and OPF datasets is provided in the \datakit technical report~\cite{puech2025gridfmdatakitv1pythonlibraryscalable}.}, referred to as HQ1200, using normalized Shannon entropy\footnote{Standard deviation overestimates variability when features take few discrete values (e.g., $P_g$ switching between zero and its upper bound). In contrast, Shannon entropy remains low when samples concentrate in a small region of the admissible space and increases only when the empirical distribution spreads across multiple populated regions.} across key grid features, following~\cite{puech2025gridfmdatakitv1pythonlibraryscalable, hedgeopf2025}. Values near zero indicate little variability, whereas values approaching one correspond to broad and nearly uniform exploration of the feature’s admissible range.\\

Real topologies and PF data were extracted from SCADA at 30-minute intervals over the year 2024, forming a dataset of network snapshots that comprises approximately 1,200 nodes and 1,600 branches after aggregating the full HQ transmission network ($\sim$2,600 nodes). These snapshots are the output of the state estimator, and each power-flow solution has been verified for convergence with the PSS$^{\circledR}$E software~\cite{siemens_psse_v34}. We compute the normalized Shannon entropy on a set of 15,000 samples and compare it with the entropy obtained from the same number of samples generated with \datakit on a representative HQ1200 base topology in \cref{fig:spider_hq1200}. Only the base topology and grid parameters are given as input to \datakit, which then generates load, topology, admittance, and generator cost scenarios. \\

The results indicate that the entropy of the synthetic data closely matches that of the real-world data, except for reactive power demand, which is slightly more diverse in the real data. This gap is consistent with reduced power factor variability in \datakit, resulting from the joint scaling of active and reactive demands through a global scaling factor (while only the local scaling factors differ between active and reactive loads; see \cref{eq:scaling}).\\

\begin{boxH}
\paragraph{Takeaways.}
\begin{itemize}
\item \datakit enables fast and scalable generation of diverse, realistic PF and OPF datasets, generating 200,000 IEEE 118-bus samples in less than 20 minutes for PF and less than 2 hours for OPF using 20 CPU cores, while scaling to 10k-bus systems for OPF and 30k-bus systems for PF.
\item \datakit addresses key limitations of existing PF and OPF datasets by generating realistic off-nominal PF operating conditions and OPF scenarios with diverse generator cost coefficients.
\item The diversity of \datakit-generated data closely aligns with that of real Hydro-Québec SCADA data on the HQ1200 network.
\end{itemize}
\end{boxH}

\section{\genco Model Architecture}

\label{sec:model}

\begin{figure*}
    \centering
    \includegraphics[width=\linewidth]{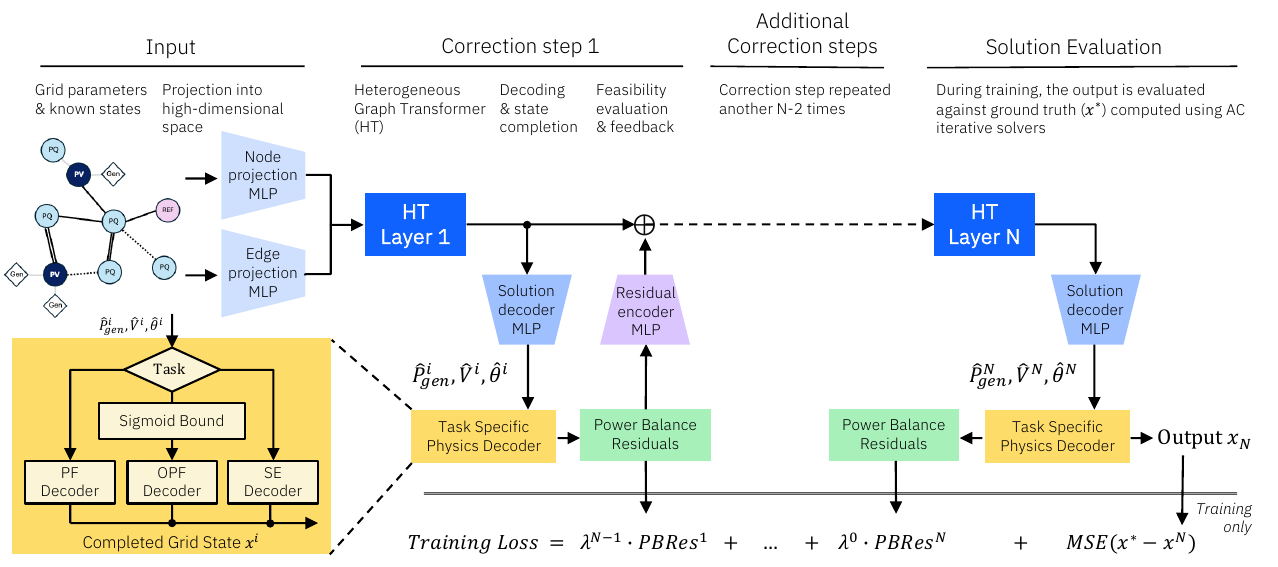}
    \caption{Overview of \genco.}
    \label{fig:architecture}
\end{figure*}

In this section, we provide an overview of our model, \genco, which is depicted in \cref{fig:architecture}. The architecture comprises a Multi-Layer Perceptron (MLP)-based input projection layer (\cref{subsec:input}) that projects grid parameters and known states into a high-dimensional latent space, followed by $N$ iterative correction steps (\cref{subsec:correction}). At each step $i$, a heterogeneous Graph Transformer (HT) layer, implemented using the \texttt{HeteroConv}~\cite{Fey_Lenssen_2019, Fey_etal_2025} and \texttt{TransformerConv}~\cite{shi2021maskedlabelpredictionunified} operators, is followed by a Solution Decoder MLP that decodes the primary variables $(\hat{P}_{\text{gen}}^i, \hat{V}^i, \hat{\theta}^i)$. The decoded variables can optionally be bounded by a sigmoid transformation to enforce box constraints before passing through a task-specific physics decoder to obtain a complete intermediate solution containing all injections and voltage states. Power Balance Residuals ($\mathrm{PBRes}^i$) are then computed and fed back through a Residual Encoder MLP, enabling feasibility-aware refinement of the latent representation of the intermediate solution. Finally, the model is trained against ground-truth solutions from classical methods---obtained by Newton--Raphson for PF, interior point methods for OPF, and Weighted Least Squares for SE---by minimizing a loss that combines a supervised distance term to the ground-truth solution with physics-informed residuals accumulated across all correction steps to encourage early feasibility (\cref{subsec:eval}).

\subsection{Input}
\label{subsec:input}

\subsubsection{Grid Parameters and Known States}

The input contains all grid parameters and known operating-state variables as
a heterogeneous directed graph. The graph has two node types—buses and generators—and two edge types: directed branch edges (one per
direction per physical branch) and featureless bus--generator connectivity
edges. We denote by $\mathcal{N}$ the set of buses, $\mathcal{G}$ the set of
generators, and $\mathcal{E}'$ the set of directed branch edges (i.e., one
edge per direction per physical branch). The slack bus is denoted by REF. Bus features encode active and reactive loads, voltage angles and magnitudes, reactive power generation, as well as shunt admittances, voltage limits, and bus-type
indicators (PQ, PV, REF); branch edge features encode admittance parameters, tap ratios,
and thermal and angle limits; generator features encode active and reactive generation,
generation limits, and quadratic dispatch cost coefficients. Variables that
are unknown and must be solved are zeroed out in the input, while known
variables are passed through unchanged. 

\subsubsection{Projection into High-Dimensional Space}

The input projection layer lifts the raw grid features into a shared
high-dimensional latent space. Separate two-layer feed-forward networks
project bus, generator, and branch edge features independently into latent representations
of dimension $H$:
\begin{equation}
    \label{eq:input_proj}
  h_{\text{bus}}^0 \in \mathbb{R}^{|\mathcal{N}| \times H}, \quad
  h_{\text{gen}}^0 \in \mathbb{R}^{|\mathcal{G}| \times H}, \quad
  h_{\text{edge}} \in \mathbb{R}^{|\mathcal{E}'| \times H}.
\end{equation}
Edge embeddings $h_{\text{edge}}$ are computed once and held constant; only bus
and generator embeddings are updated across correction steps.

\subsection{Correction Steps}
\label{subsec:correction}

After projecting the inputs into a higher-dimensional space, \genco operates a series of $N$ correction steps to update the latent representation of a candidate solution for the tasks it solves. This solution is then decoded, evaluated for feasibility of the power balance equations, and updated according to a feedback message.

\subsubsection{Latent Solution Update with an HT Layer}
The GNN layer propagates information between electrically connected components of the grid to produce updated latent representations of the system state at each correction step. An HT layer with $N_h$ attention heads updates bus and generator embeddings by aggregating messages from electrically connected neighbors, using separate attention parameters for each relation type (bus-to-bus, gen-to-bus, bus-to-gen). For bus-to-bus edges, branch embeddings $h_{\text{edge}}$ (computed from the branch admittance and thermal and angle limits) condition the attention mechanism so that electrical coupling strengths modulate information flow. After the GNN update at correction step $i$, each bus and generator has a latent embedding of dimension $N_h \cdot H$:
\begin{equation}
\label{eq:transformer}
z_{\text{bus}}^i \in \mathbb{R}^{|\mathcal{N}| \times (N_h H)}, \qquad
z_{\text{gen}}^i \in \mathbb{R}^{|\mathcal{G}| \times (N_h H)}.
\end{equation}

Thus, for a model with $N$ correction steps, there is one updated bus and generator embedding per node \emph{per step}, encoding both the grid topology and the feasibility feedback injected at the end of the previous correction step (as discussed in \cref{subsec:feedback}).

\subsubsection{Decoding and Solution Completion}
\label{subsubsec:decoding}

At each correction step $i$, a Solution Decoder MLP (with weights shared across all steps) maps the latent embeddings to primary variables
$(\hat{P}_{\mathrm{gen}}^i,\hat{V}^i,\hat{\theta}^i)$.
These variables are optionally projected onto box constraints and then passed through a task-specific physics decoder to construct a complete intermediate solution.

\paragraph{Primary variable decoding.}

A bus-level Solution Decoder MLP maps the bus latent embeddings
$z_{\mathrm{bus}}^i$
to reconstructed voltage magnitudes and angles:
\begin{equation}
\hat{V}^i \in \mathbb{R}^{|\mathcal{N}|},
\qquad
\hat{\theta}^i \in \mathbb{R}^{|\mathcal{N}|}.
\end{equation}

For OPF, a generator-level Solution Decoder MLP maps
$z_{\mathrm{gen}}^i$
to reconstructed active power generations:
\begin{equation}
\hat{P}_{\mathrm{gen}}^i \in \mathbb{R}^{|\mathcal{G}|}.
\end{equation}

For PF and SE, only the voltage state variables are decoded.
All known quantities (e.g., loads and specified generation in PF) are passed through unchanged.

\paragraph{Box constraints.}

For OPF, voltage magnitudes and active generator powers are constrained to their feasible intervals through a sigmoid projection:
\begin{equation}
\hat{x}^i
=
\underline{x}
+
(\overline{x}-\underline{x})
\,\sigma(\tilde{x}^i),
\end{equation}
where $\tilde{x}^i$ denotes the unconstrained decoder output,
$\sigma(\cdot)$ is the sigmoid function,
and $(\underline{x},\overline{x})$ are the lower and upper bounds. No box constraints are imposed for PF or SE.

\paragraph{Physics decoder.}

Given $(\hat{V}^i,\hat{\theta}^i)$, the physics decoder reconstructs
network-implied electrical quantities using the branch admittance model. This applies across all PF, OPF, and SE tasks.
\\\\
Each directed branch edge
$e \in \mathcal{E}'$
is associated with a source bus
$s(e)$
and a target bus
$t(e)$.
The active and reactive power flows leaving the source bus $s(e)$ through edge $e$ are computed as
\begin{align}
\hat{P}_{e}^i
&=
\hat{P}_{e}(\hat{V}^i,\hat{\theta}^i),
\\
\hat{Q}_{e}^i
&=
\hat{Q}_{e}(\hat{V}^i,\hat{\theta}^i).
\end{align}

The corresponding \emph{power-flow-implied injections} are then obtained by summing all outgoing
branch flows:
\begin{align}
\label{eq:inj}
\hat{P}_{\mathrm{inj},k}^i
&=
\sum_{e\,:\,s(e)=k}
\hat{P}_{e}^i,
\\
\hat{Q}_{\mathrm{inj},k}^i
&=
\sum_{e\,:\,s(e)=k}
\hat{Q}_{e}^i.
\end{align}

\paragraph{Recovery of generation variables for OPF.}

Reactive generator powers at PV and REF buses are analytically recovered from the reactive power balance equation:
\begin{equation}
\label{eq:reactive_power_balance}
\hat{Q}_{g,j}^i
=
\hat{Q}_{\mathrm{inj},j}^i
+
Q_{d,j}
-
Q_{\mathrm{shunt},j}(\hat{V}^i),
\end{equation}
where
$Q_{d,j}$ is the reactive demand and
$Q_{\mathrm{shunt},j}(\hat{V}^i)$
denotes the shunt reactive injection computed from the reconstructed voltage.

\paragraph{Recovery of generation variables for PF.}

Reactive generation
$\hat{Q}_{g,j}^i$
is similarly recovered at PV and REF buses using \cref{eq:reactive_power_balance}.
At the REF bus, bus-level active generation
$\hat{P}_{g,\mathrm{REF}}^i$
is recovered from the active power balance equation.
\\\\
A key property of the model is that whenever a variable is analytically recovered from a balance equation,
the corresponding residual component is structurally zero
(\cref{subsec:feedback}). For SE, no generation variables are recovered as the model directly predicts net nodal injections.

\paragraph{Intermediate solution.}

The intermediate solution at correction step $i$ is defined as
\begin{equation}
x^i =
\begin{cases}
[
\hat{V}^i,
\hat{\theta}^i,
\hat{P}_{\mathrm{gen}}^i,
\hat{Q}_{g}^i,
\hat{P}_{\mathrm{flow}}^i,
\hat{Q}_{\mathrm{flow}}^i
]
& \text{(OPF)},
\\[6pt]
[
\hat{V}^i,
\hat{\theta}^i,
\hat{P}_{g,\mathrm{REF}}^i,
\hat{Q}_{g}^i,
\hat{P}_{\mathrm{flow}}^i,
\hat{Q}_{\mathrm{flow}}^i
]
& \text{(PF)},
\\[6pt]
[
\hat{V}^i,
\hat{\theta}^i,
\hat{P}_{\mathrm{inj}}^i,
\hat{Q}_{\mathrm{inj}}^i,
\hat{P}_{\mathrm{flow}}^i,
\hat{Q}_{\mathrm{flow}}^i
]
& \text{(SE)}.
\end{cases}
\end{equation}

For SE, injections are not decomposed into generation and load components.

\subsubsection{Feasibility Evaluation and Feedback}
\label{subsec:feedback}

The feasibility evaluation quantifies the deviation of the intermediate solution $x^i$ from satisfying the nodal power balance equations\footnote{Only active and reactive power-balance residuals are used in the intermediate feedback at each correction step. Other OPF constraints (e.g., thermal limits, branch angle differences, and reactive generation bounds) are not included in this feedback signal. Extending the mechanism to incorporate all constraint types would require combining heterogeneous residuals with additional learned mappings or carefully tuned weightings, substantially increasing hyperparameter complexity, and is left for future work.}, and provides a feedback signal for the next correction step.

\paragraph{Feasibility evaluation using power balance residuals.}
For each bus $k$, we define the per-bus power balance residual vector at correction step $i$ as
\begin{equation}
\mathrm{PBRes}_k^i :=
\begin{bmatrix}
\mathrm{PBRes}_{P,k}^i \\
\mathrm{PBRes}_{Q,k}^i
\end{bmatrix}.
\end{equation}

The active and reactive components are given by:
\begin{align}
\mathrm{PBRes}_{P,k}^i
&:=
\hat{P}_{\mathrm{bus},k}^i
-
\hat{P}_{\mathrm{inj},k}^i
+
P_{\mathrm{shunt},k}(\hat{V}^i),
\nonumber\\
\mathrm{PBRes}_{Q,k}^i
&:=
\hat{Q}_{\mathrm{bus},k}^i
-
\hat{Q}_{\mathrm{inj},k}^i
+
Q_{\mathrm{shunt},k}(\hat{V}^i).
\label{eq:PBRes}
\end{align}
They are computed by comparing (i) the \emph{power-flow-implied injections} $\hat{P}_{\mathrm{inj},k}^i$ and $\hat{Q}_{\mathrm{inj},k}^i$, obtained by summing branch flows (\cref{eq:inj}), and (ii), the \emph{net bus injections}, defined as the difference between bus-level generation and load:
\begin{align}
\label{eq:bus-inj}
\hat{P}_{\mathrm{bus},k}^i &= \hat{P}_{g,k}^i - P_{d,k}, \\
\hat{Q}_{\mathrm{bus},k}^i &= \hat{Q}_{g,k}^i - Q_{d,k}.
\end{align}

For OPF, bus-level generation $\hat{P}_{g,k}$ is obtained by summing the active power generation of all generators connected to bus $k$.\\

Some residual components are structurally zero because the corresponding generation variables are analytically recovered from the power balance equations and therefore satisfy them by construction (e.g., reactive generation at PV buses) (\cref{subsubsec:decoding}). In SE, residuals are evaluated only at buses with available measurements of both generation and load.

\paragraph{Feedback.}
A Residual Encoder MLP maps the residual vector back into the latent space and injects it into the bus embedding before the next GNN layer:

\begin{equation}
h_{\mathrm{bus},k}^{i+1}
=
z_{\mathrm{bus},k}^i
+
\operatorname{MLP}'_{\mathrm{bus}}
\left(
\mathrm{PBRes}_k^i
\right).
\end{equation}

This closes the correction loop: at each step, the model receives an explicit
physics-grounded infeasibility feedback that enables iterative refinement toward
feasible solutions across the $N$ correction steps.

\subsection{Final Solution Evaluation and Training Loss}
\label{subsec:eval}

The training objective combines supervised learning on the final solution with
intermediate self-supervised feasibility regularization through the power balance residuals.

The final prediction is $\hat{x} := x^N$, and the training loss is defined as
\begin{equation}
\label{eq:loss}
\mathcal{L}
=
\mathcal{L}_{\mathrm{supervised}}(x^N, x^*)
+
\sum_{i=1}^{N}
\lambda^{N-i}
\,
\overline{\|\mathrm{PBRes}^i\|}_2,
\end{equation}
where $x^*$ denotes the ground-truth solution obtained from the corresponding classical solver,
$\mathcal{L}_{\mathrm{supervised}}$ is a task-specific supervised loss, and $\overline{\|\mathrm{PBRes}^i\|}_2$ denotes the mean $\ell_2$ norm of the per-bus power balance residual vectors $\mathrm{PBRes}_k^i$ across all buses at correction step $i$. The factor $\lambda^{N-i}$, with $\lambda<1$, is an exponentially increasing weight that places greater emphasis on later correction steps.
The supervised loss is instantiated differently for each task:
MSE on bus- and generator-level variables for OPF,
MSE on bus-level variables for PF,
and MAE on bus- and edge-level variables for SE.\\

\subsection{Training \genco with gridfm-graphkit}

\label{subsec:graphkit}

\genco is implemented and trained using \graphkit\footnote{\url{https://github.com/gridfm/gridfm-graphkit}},
our PyTorch Lightning–based framework for developing neural solvers and grid foundation models. Training and evaluation are configured via YAML files and run through a CLI. The library is optimized for large-scale training. It supports multi-GPU execution, multi-grid training, fine-tuning, and model compilation.\\

\graphkit decouples architecture design from data engineering, training, and evaluation by providing a unified heterogeneous graph structure as a common backbone, shared across PF, OPF, and SE tasks. It handles graph construction, feature normalization, and masking of task-dependent variables, ensuring a consistent representation across models (\cref{fig:motivational_example}). Since new architectures are integrated by defining a forward pass over this standardized graph, it enables straightforward addition and testing of new neural solvers. Further, training components such as losses, optimizers, checkpointing, and logging are implemented in \graphkit. The framework includes power system-specific modules, including physics decoders (\cref{subsubsec:decoding}), power-flow computation, and evaluation of constraint violations and optimality gaps, reducing the need for domain-specific implementation. It also enables direct comparison with classical solvers using precomputed solutions from \datakit.\\

The main benefit brought to the community by \graphkit is benchmarking consistency: data loading, splitting, and evaluation metrics are standardized across models, so that differences arise only from architecture specifications in configuration files while evaluation protocols remain fixed.\\

\begin{boxH}
\paragraph{Takeaways.}
\begin{itemize}
\item \genco is a unified iterative correction architecture that maps grid states to PF/OPF/SE solutions via $N$ correction steps in a high-dimensional latent space, learned from precomputed solutions.
\item Each iteration decodes full physical states and is refined using power-balance residuals, providing physics-grounded infeasibility feedback (self-supervision).
\item The architecture combines learned message passing with sparse attention with a task-specific physics decoder that performs solution completion via power-flow equations and enforces OPF box constraints through sigmoid projections.
\end{itemize}
\end{boxH}

\section{Results}
\label{sec:experiments}

We evaluate \genco against classical and state-of-the-art neural solvers for power flow (PF) in \cref{subsec:pf}, optimal power flow (OPF) in \cref{subsec:opf}, and state estimation (SE) in \cref{subsec:SE}. We then study runtime and accuracy scaling with grid size and compare against classical AC and DC PF/OPF solvers in \cref{subsec:scaling}. Next, we assess generalization to topology perturbations, out-of-distribution operating conditions, and unseen grids in \cref{subsec:generalization}. Finally, we validate \genco on real SCADA data from the Hydro-Québec grid in \cref{subsec:hq1200}.

Throughout this section, we consider three \genco variants that differ only in their hidden dimension $H$, which determines the size of the input representations after projection (\cref{eq:input_proj}) and the latent space (\cref{eq:transformer}): \genco Base ($H=48$), Small ($H=24$), and Tiny ($H=12$), with 20.1M, 5.0M, and 1.3M parameters, respectively. All other architectural components are kept identical across variants. Unless stated otherwise, ``\genco'' refers to \genco Base.

\subsection{Power Flow}
\label{subsec:pf}

We evaluate \genco on Power Flow (PF) using PF$\Delta$~\cite{NEURIPS2025_d000ef56}, a benchmark of 12 tasks designed to assess neural PF solvers under varying topologies, data regimes, and operating conditions. We compare \genco against CANOS-PF (adapted from~\cite{piloto2024}), GNS-S~\cite{DONON2020106547}, and PFNet~\cite{lin2024powerflownet}, on the subset of tasks listed in \cref{tab:tasks-summary}, for which results on the IEEE 118-bus system were reported by the authors. Extensive comparisons against classical AC and DC solvers are presented in \cref{subsec:scaling}, as \pfdelta does not provide DC-PF solutions nor classical solver runtimes. CANOS-PF is a soft-constrained supervised model trained with a combination of an $\ell_2$ loss to precomputed AC power flow solutions and a penalty on power-balance residuals. PFNet is a purely supervised model trained with an $\ell_2$ loss on precomputed AC power flow solutions. GNS-S is a self-supervised method that iteratively minimizes power-balance violations through residual-based updates of nodal features.\\

\begin{table}
\centering
\small
\renewcommand{\arraystretch}{1.0}
\begin{tabularx}{\linewidth}{lX}
\toprule
\textbf{Task} & \textbf{Training setup} \\
\midrule
\multicolumn{2}{l}{\textbf{Generalization to topology perturbations}} \\
\midrule
1.1 & Train on (N) only. \\
1.2 & Train on (N) and (N{-}1). \\
1.3 & Train on (N), (N{-}1), and (N{-}2). \\
\midrule
\multicolumn{2}{l}{\textbf{Data efficiency}} \\
\midrule
2.1 & Baseline setting (same as 1.3; 54k training samples). \\
2.2 & \textit{Skipped; no results reported by the authors.} \\
2.3 & Same training composition as 2.1, with $3\times$ less training data (18k samples). \\
\midrule
\multicolumn{2}{l}{\textbf{Generalization to different grid sizes}} \\
\midrule
3.1 & Train on IEEE 118; evaluate on the unseen IEEE 57 and GOC 500 grids. \\
\midrule
\multicolumn{2}{l}{\textbf{Training on challenging PF cases}} \\
\midrule
4.1 & Train on 90\% feasible and 10\% C2I samples. \\
4.2 & Train on 50\% feasible, 10\% C2I, and 40\% approaching-infeasibility samples. \\
4.3 & Train on 20\% C2I and 80\% approaching-infeasibility samples only. \\
\bottomrule
\end{tabularx}
\caption{PF$\Delta$ tasks evaluated in this work. The tasks assess (i) generalization to topology perturbations (Tasks 1.x), (ii) data efficiency (Tasks 2.x), (iii) generalization to different grid sizes (Task 3.1), and (iv) the effect of training on challenging PF cases (Tasks 4.x). Each task trains a separate neural solver on the IEEE 118-bus system and is evaluated on (N), (N{-}1), (N{-}2), and close-to-infeasible (C2I) test cases for IEEE 118, except Task 3.1, which additionally evaluates on the unseen IEEE 57- and GOC 500-bus systems. Tasks 4.1--4.3 vary the composition of the training set by progressively replacing feasible samples with C2I and approaching-infeasibility samples. C2I and approaching-infeasibility samples correspond to operating points at and immediately preceding the steady-state stability limit, respectively.}
\label{tab:tasks-summary}
\end{table}

\begin{figure*} \centering \includegraphics[width=\linewidth]{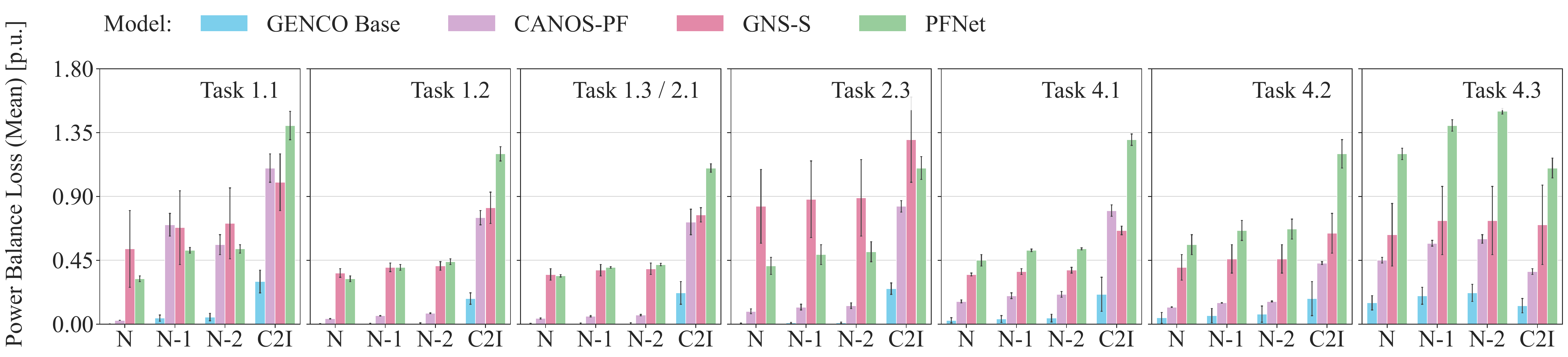} 
\caption{Mean power-balance loss (averaged across all buses of all samples) for \genco, CANOS-PF, GNS-S, and PFNet across PF$\Delta$ tasks on the IEEE 118-bus system. Error bars in the figure correspond to the standard deviation over three seeds. C2I denotes close-to-infeasible cases at the steady-state stability limit, where classical solvers face challenges~\cite{NEURIPS2025_d000ef56}.}
\label{fig:mean-pb} \end{figure*}

We report mean power-balance loss as in~\cite{NEURIPS2025_d000ef56}, which is computed for each bus $k$ as the $\ell_2$ norm of the nodal power-balance residuals (defined in \cref{eq:PBRes}), i.e., $
\|\mathrm{PBRes}_k\|_2
=
\sqrt{\mathrm{PBRes}_{P,k}^2 + \mathrm{PBRes}_{Q,k}^2},
$
and averaged over all buses across all PF samples within a given test set, with values expressed in p.u. on a 100 MVA base, i.e., \(x_{\mathrm{p.u.}} = x / S_{\mathrm{base}}\), where \(S_{\mathrm{base}} = 100\,\mathrm{MVA}\) is the power base.\\

We report \genco's performance for all tasks except Task 3.1 (\cref{table:all-cases-mean}) in \cref{fig:mean-pb}. The results indicate that \genco achieves substantially lower errors and variance than all baselines across all tasks and topology variants. We complement the figure and additionally provide a relative measure of error magnitude, expressing the mean power-balance loss as a percentage of the mean bus apparent power injection\footnote{For a bus $k$, the apparent power is $|S_k| = \sqrt{P_{\mathrm{bus},k}^2 + Q_{\mathrm{bus},k}^2}$.}. In the baseline setting, where \genco is trained on (N), (N{-}1), and (N{-}2) topologies and evaluated on the corresponding test sets (Tasks 1.3/2.1), the loss amounts to only 0.28\%, 0.53\%, and 0.69\% of the mean apparent power under (N), (N{-}1), and (N{-}2) conditions, respectively (with mean bus apparent powers of 0.95, 0.87, and 0.86 p.u.). Among the baselines, only CANOS-PF remains within the single-digit percentage range (4.2\%, 6.4\%, and 7.3\%), whereas all other methods exhibit errors corresponding to approximately 40--50\% of the mean apparent power, indicating substantial violations of power-balance constraints and very limited physical consistency for these solvers.\\

Examining individual task groups provides further insight into \genco's behavior: Tasks~1.1--1.3 show that adding (N{-}1) contingencies to the training set is sufficient for \genco to generalize to both (N{-}1) and (N{-}2) topologies with essentially no loss in accuracy compared to (N). Even training only on (N) topologies already yields strong performance. In contrast, the baseline solvers, particularly CANOS-PF, exhibit substantially larger residuals on (N{-}1) and (N{-}2) topologies when such perturbations are absent from the training set. Tasks~2.1 and~2.3 further demonstrate that \genco maintains low error when the training set is reduced by a factor of three, while keeping the training composition and test sets unchanged. While CANOS-PF experiences only a modest degradation, the error of GNS-S nearly doubles between Tasks~2.1 and~2.3. Task~4.1 indicates that replacing 10\% of feasible training samples with close-to-infeasible (C2I) samples provides only marginal improvements on C2I test cases while slightly degrading performance on the remaining datasets compared to Tasks~1.3/2.1. Tasks~4.2 and~4.3 show that progressively increasing the proportion of challenging training samples, from 50\% to 100\%, further improves performance on C2I cases, particularly in terms of standard deviation, but comes at the expense of increased errors on the other datasets.\\

\cref{fig:max-pb} (Appendix) shows that \genco also achieves substantially lower \emph{maximum} power-balance errors than all baselines across (N), (N{-}1), and (N{-}2) topologies. On close-to-infeasible cases, however, GNS-S exhibits lower worst-case errors, potentially due to its self-supervised training objective, which may provide greater robustness to out-of-distribution operating conditions. Nevertheless, all methods exhibit very large worst-case residuals, often exceeding several times the mean apparent power\footnote{A more detailed analysis of the distribution of these residuals would be required to determine what fraction of test samples exhibit such large errors, which we leave for future work.}.\\

\begin{table}
    \centering
    \small
    \resizebox{\columnwidth}{!}{
    \renewcommand{\arraystretch}{1.2}
    \begin{tabular}{c|c|c|c|c|c} \hline 
        
        \multicolumn{2}{c|}{\textbf{Experiment}} & \multicolumn{4}{|c}{\textbf{Power Balance Loss (Mean) [p.u.]}}\\ \hline 
        
        \textbf{Case} & 
        \textbf{Model} & \textbf{N} & \textbf{N-1} 
        & \textbf{N-2} & \textbf{Close-to-infeasible} \\ \hline

        \multirow{5}{*}{57} 
        & \genco Base
        & \G{4.9e-1}\std{1.9e-1}
        & \G{5.0e-1}\std{2.07e-1}
        & \G{4.8e-1}\std{1.7e-1}
        & \G{5.8e-1}\std{2.2e-1} \\
        & PFNet
        & 2.3\std{0.4}
        & 2.3\std{0.4}
        & 2.3\std{0.4}
        & 2.4\std{0.4} \\
        & CANOS-PF
        & 1.7\std{0.2}
        & 1.8\std{0.2}
        & 1.8\std{0.2}
        & 1.8\std{0.2} \\
        & GNS-S
        & 3.3e1\std{1.4e1}
        & 8.8e1\std{3.9e1}
        & 1.5e2\std{0.7e2}
        & 8.0e-1\std{2.0e-1} \\
        & \textcolor{gray}{NR}
        & \textcolor{gray}{1.1e-6\std{0.0e-6}}
        & \textcolor{gray}{1.2e-6\std{0.0e-6}}
        & \textcolor{gray}{1.1e-6\std{0.0e-6}}
        & \textcolor{gray}{1.3e-6\std{0.0e-6}} \\ \hline

        \multirow{5}{*}{118} 
        & \genco Base
        & \G{2.7e-3}\std{2.1e-3}
        & \G{4.5e-3}\std{3.4e-3}
        & \G{5.8e-3}\std{4.3e-3}
        & \G{2.2e-1}\std{0.8e-1} \\
        & PFNet
        & 3.4e-1\std{0.08e-1}
        & 4.0e-1\std{0.05e-1}
        & 4.2e-1\std{0.09e-1}
        & 1.1\std{0.03} \\
        & CANOS-PF
        & 4.0e-2\std{0.4e-2}
        & 5.6e-2\std{0.5e-2}
        & 6.3e-2\std{0.5e-2}
        & 7.2e-1\std{0.9e-1} \\
        & GNS-S
        & 3.5e-1\std{0.4e-1}
        & 3.8e-1\std{0.4e-1}
        & 3.9e-1\std{0.4e-1}
        & 7.7e-1\std{0.5e-1} \\
        & \textcolor{gray}{NR}
        & \textcolor{gray}{3.7e-6\std{0.0e-6}}
        & \textcolor{gray}{3.2e-6\std{0.0e-6}}
        & \textcolor{gray}{3.3e-6\std{0.0e-6}}
        & \textcolor{gray}{4.6e-6\std{0.0e-6}} \\ \hline

        \multirow{5}{*}{500} 
        & \genco Base
        & \G{8.7}\std{5.9}
        & \G{8.7}\std{5.9}
        & \G{8.7}\std{5.9}
        & \G{8.7}\std{5.8} \\
        & PFNet
        & 8.3e1\std{3.9e1}
        & 8.1e1\std{3.8e1}
        & 8.4e1\std{4.0e1}
        & 8.9e1\std{3.8e1} \\
        & CANOS-PF
        & 2.3e1\std{0.6e1}
        & 2.3e1\std{0.6e1}
        & 2.3e1\std{0.6e1}
        & 2.6e1\std{0.6e1} \\
        & GNS-S
        & 2.4e1\std{0.7e1}
        & 2.4e1\std{0.7e1}
        & 2.4e1\std{0.7e1}
        & 2.4e1\std{0.7e1} \\
        & \textcolor{gray}{NR}
        & \textcolor{gray}{1.4e-5\std{0.0e-5}}
        & \textcolor{gray}{1.4e-5\std{0.0e-5}}
        & \textcolor{gray}{1.3e-5\std{0.0e-5}}
        & \textcolor{gray}{1.6e-5\std{0.0e-5}} \\ \hline
    \end{tabular}}
\caption{Mean power balance loss for Task 3.1, where neural solvers are trained on IEEE 118 and evaluated on IEEE 57, IEEE 118, and GOC 500 under (N), (N{-}1), (N{-}2), and close-to-infeasible conditions. Errors denote the standard deviation across three random seeds. NR denotes AC power flow solved using Newton--Raphson (highlighted in light gray). \textbf{Bold} values indicate the best performance among learning-based methods.}
\label{table:all-cases-mean}
\end{table}

Finally, Task~3.1 evaluates models trained on IEEE 118 on the unseen IEEE 57 and GOC 500 grids (\cref{table:all-cases-mean}). \genco achieves the lowest mean power-balance loss on both unseen grids, although all models suffer from substantial degradation under cross-grid evaluation. Relative to the mean apparent power averaged over (N), (N{-}1), and (N{-}2), \genco's mean error is approximately equal to the average apparent power on IEEE 57 and approximately 14$\times$ larger on GOC 500, while the baselines show even larger errors.\\ 

\begin{boxH}
\paragraph{Takeaways.}
\begin{itemize}
\item When used to solve Power Flow, \genco achieves very low \emph{mean} power-balance residuals (\textless 1\% of the mean apparent power), lower than all state-of-the-art models on the PF$\Delta$ dataset.
\item It is more data-efficient and more robust to close-to-infeasible cases than competing methods.
\item All baseline models and \genco exhibit large worst-case errors and degraded performance on unseen grids, with power-balance residuals reaching multiples of the mean apparent power in both cases, highlighting the need for more robust solvers and more sophisticated pretraining.\end{itemize}
\end{boxH}

\subsection{Optimal Power Flow}
\label{subsec:opf}

\begin{table*}
\caption{%
  Constraint violations and optimality gap comparison between \genco Base and HH-MPNN.
  \textbf{Bold} indicates the better result per metric per system.
  $Q_g$ violations are structurally zero for HH-MPNN (reactive limits enforced via sigmoid activation).%
}
\label{tab:opf_results}
\centering
\small
\setlength{\tabcolsep}{4pt}
\begin{tabular}{@{}llllllll@{}}
\toprule
& & Optimality & \multicolumn{2}{c}{Thermal limits} & \multicolumn{2}{c}{Power balance} & React.\ gen. bounds \\
\cmidrule(lr){3-3}\cmidrule(lr){4-5}\cmidrule(lr){6-7}\cmidrule(lr){8-8}
System & Model
  & Gap (\%)
  & $S_{ij}(+)$ [MVA]
  & $S_{ij}(-)$ [MVA]
  & $\mathrm{PBRes}_{P}$ [MW]
  & $\mathrm{PBRes}_{Q}$ [MVar]
  & $Q_g$ [MVar] \\
\midrule
\multirow{2}{*}{IEEE 14}
  & HH-MPNN & \G{0.01}           & 0.00                     & 0.00                     & \G{2.00e-4}              & 2.70e-2                  & \G{0.00} \\
  & \genco Base   & 0.05\std{0.01}      & 0.00                     & 0.00                     & 8.27e-4\std{1.13e-4}     & \G{3.83e-4}\std{5.24e-5} & 9.18e-3\std{3.36e-4} \\
\midrule
\multirow{2}{*}{IEEE 30}
  & HH-MPNN & 0.18               & 5.20e-3                  & \G{3.00e-4}                  & 1.31e-2                  & 7.60e-3                  & \G{0.00} \\
  & \genco Base   & \G{0.03}\std{0.00}  & \G{1.30e-3}\std{2.14e-4} & 1.28e-3\std{2.05e-4} & \G{1.05e-3}\std{7.22e-5} & \G{7.29e-4}\std{2.19e-5} & 2.68e-3\std{3.03e-4} \\
\midrule
\multirow{2}{*}{IEEE 57}
  & HH-MPNN & \G{0.00}            & \G{0.00}                 & \G{0.00}                 & \G{2.10e-2}              & 1.21e-1                  & \G{0.00} \\
  & \genco Base   & 0.12\std{0.02}     & 5.01e-7\std{2.66e-7}     & 1.14e-6\std{1.06e-6}     & 5.15e-2\std{1.32e-2}     & \G{2.98e-2}\std{6.67e-3} & 1.23e-2\std{4.95e-3} \\
\midrule
\multirow{2}{*}{IEEE 118}
  & HH-MPNN & 0.24                & 1.37e-1                  & 1.38e-1                  & 5.11e-2                  & 1.99e-1                  & \G{0.00} \\
  & \genco Base   & \G{0.16}\std{0.02} & \G{4.37e-3}\std{7.11e-4} & \G{4.39e-3}\std{7.03e-4} & \G{2.19e-2}\std{1.31e-2} & \G{5.14e-3}\std{2.70e-3} & 2.02e-3\std{4.03e-4} \\
\midrule
\multirow{2}{*}{GOC 500}
  & HH-MPNN & \G{0.02}           & 5.80e-2                  & 5.60e-2                  & \G{1.72e-2}              & 1.09e-1                  & \G{0.00} \\
  & \genco Base   & 0.30\std{0.04}      & \G{1.91e-3}\std{2.52e-4} & \G{1.88e-3}\std{2.55e-4} & 7.02e-2\std{8.35e-3}     & \G{2.60e-2}\std{1.60e-3} & 4.57e-4\std{1.06e-4} \\
\midrule
\multirow{2}{*}{GOC 2000}
  & HH-MPNN & \G{0.00}            & 2.80e-3                  & 2.80e-3                  & \G{1.03e-2}              & 2.65e-2                  & \G{0.00} \\
  & \genco Base   & 0.17\std{0.01}      & \G{2.17e-5}\std{1.09e-6} & \G{2.45e-5}\std{1.43e-6} & 5.87e-2\std{6.04e-3}     & \G{2.23e-2}\std{2.78e-3} & 1.64e-3\std{3.33e-4} \\
\bottomrule
\end{tabular}
\\
\raggedright\footnotesize
$\pm$ denotes standard deviation over three seeds.
\end{table*}

We evaluate \genco on the OPFData~\cite{lovett2024opfdatalargescaledatasetsac} benchmark against the Hybrid Heterogeneous Message Passing Neural Network (HH-MPNN) proposed in~\cite{arowolo2025}, which is the latest baseline available for this dataset at the time of writing. We use the dataset version containing N-1 topology perturbations and follow exactly the same train/validation/test splits as in \cite{arowolo2025}: 270,000 samples for training (90\%), 15,000 for validation (5\%), and 15,000 for testing (5\%). Experiments are conducted on six benchmark systems ranging from 14 to 2000 buses. All \genco experiments are repeated over three random seeds while keeping identical data splits to ensure a fair comparison with \cite{arowolo2025}, which only reports results for one seed.
\\\\

\cref{tab:opf_results} reports both optimality and feasibility metrics. Optimality gap (\%) is computed as the mean relative objective gap with respect to the IPOPT solution over the test set. Feasibility metrics report absolute mean constraint violations: branch angle-difference violations $\theta_{ij}$ are zero for all solvers and considered cases and are therefore omitted, apparent power flow limit violations $S_{ij}(+)$ and $S_{ij}(-)$ in MVA for both branch directions, active and reactive power-balance residuals $\mathrm{PBRes}_{P}$ and $\mathrm{PBRes}_{Q}$ in MW and MVar, respectively (\cref{eq:PBRes}), and reactive generation limit violations $Q_g$ in MVar. All violations are averaged over all buses and lines across all test-set instances.
\\\\
The results highlight a trade-off between optimality and feasibility. On the smaller 14- to 118-bus systems, both solvers achieve highly feasible and optimal solutions overall. Moreover, \genco shows consistently lower reactive power-balance residuals $\mathrm{PBRes}_{Q}$, often by one to two orders of magnitude compared to HH-MPNN.
\\\\
On larger systems (500- and 2000-bus systems), HH-MPNN typically achieves lower objective gaps, whereas \genco attains substantially smaller branch thermal and reactive power-balance violations. For example, on the 2000-bus system, HH-MPNN reaches a near-zero optimality gap (potentially due to reduced constraint satisfaction), while \genco reduces branch flow violations from approximately $2.8 \times 10^{-3}$ to $2.3 \times 10^{-5}$~MVA (averaged over both directions).
\\\\
\genco consistently achieves lower $\mathrm{PBRes}_{Q}$ residuals because reactive generation $Q_g$ is reconstructed analytically from the reactive power-balance equations using the decoded voltage state, rather than being predicted directly. This physics-based completion makes reactive power-balance residuals structurally zero at PV and reference buses. This design trades reactive power-balance feasibility against explicit enforcement of reactive generation limits. Unlike \cite{arowolo2025}, \genco does not project reconstructed $Q_g$ onto generator limits using sigmoid activations, which can lead to small $Q_g$ violations. However, these remain negligible in practice: the largest mean violation is below 0.036\% of the mean reactive power limits.
\\\\
Finally, in addition to achieving an extremely small optimality gap (0.3\% in the worst case), we provide context for the magnitude of the reported violations by normalizing inequality constraints by their mean limits and equality constraints by the corresponding mean net injections.\footnote{A more informative metric would normalize each violation by its constraint limit before averaging. However, these quantities are not reported in~\cite{arowolo2025}, and the variability in constraint limits (including very small bounds) can make such metrics sensitive to outliers. We therefore report mean violations relative to mean constraint limits for inequality constraints, and use mean mismatches for equality constraints, to provide a sense of scale. We leave a more complete analysis for future work.}. The largest mean thermal violation to mean thermal limit ratio is obtained for IEEE 118, where violations represent 0.0018\% of the mean limits (0.0556\% for HH-MPNN on the same case). For the power-balance mismatch relative to mean net active injection, the worst case for \genco is GOC 2000 with 0.176\% (0.082\% for HH-MPNN on IEEE 118). For reactive power-balance mismatch relative to mean net reactive injection, the worst case is IEEE 57, with 0.340\% (1.38\% for HH-MPNN on IEEE 57).\\


\begin{boxH}
\paragraph{Takeaways.}

\begin{itemize}

\item For Optimal Power Flow, evaluated on OPFData, \genco achieves \textbf{competitive optimality} (worst-case gap $\leq0.3\%$) while substantially improving feasibility, with all \textbf{mean normalized feasibility violations remaining below $0.35\%$} across all grids.
\end{itemize}
\end{boxH}

\subsection{State Estimation}
\label{subsec:SE}

We evaluate \genco on State Estimation (SE), the task of recovering bus voltage magnitudes and phase angles from noisy, incomplete, and corrupted measurements. Classical state estimators formulate SE as a nonlinear inverse problem, most commonly solved using Weighted Least Squares (WLS), which relies on an accurate grid model and a sufficiently informative (i.e. locally observable) measurement set. In contrast, \genco learns a direct mapping from measurements to grid states by combining data-driven priors with physics-informed corrections, as described in \cref{sec:model}. 
We first evaluate its robustness to sparse, noisy, and outlier-corrupted measurements, as well as cases where WLS fails to converge, before studying its resilience to inaccuracies in the underlying grid model caused by admittance parameter errors.

\paragraph{Experimental setup.} GENCO models are trained on datasets generated via \datakit, comprising roughly 250{,}000 scenarios per grid, with varying N-k topology perturbations ($k=5$ for IEEE 14/30/57 and $k=10$ for IEEE 118) and admittance perturbations, on the standard IEEE 14, 30, 57, and 118-bus systems. In the absence of standardized datasets and procedures for generating synthetic SCADA measurements, we define our measurement-generation protocol and evaluation settings as follows. We generate SCADA measurements from the true bus voltage magnitudes $V$, bus power injections $P_{\text{inj}}, Q_{\text{inj}}$, and branch flows $P_{\text{flow}}, Q_{\text{flow}}$ by applying three different types of perturbations; first, we randomly mask a fraction $p_{\text{mask}}$ of the available measurements to emulate limited measurement coverage, with the resulting number of measurements expressed as a multiple of the number of buses $|\mathcal{N}|$; second, we apply multiplicative Gaussian noise ($\sigma_V = 0.01$~p.u., $\sigma_{P,Q} = 0.02$~p.u.) to all measurements, independently perturbing each measured voltage magnitude, active/reactive power injection, and active/reactive branch flow; third, we create outliers by adding a bias of three standard deviations to a fraction $p_{\text{outlier}}$ of measurements. We consider arbitrary measurement locations, but do not enforce sensor placement according to power-system observability rules. To mitigate this simplification, we use measurement densities above typical observability requirements (at least $5|\mathcal{N}|$ measurements, compared to the commonly used $4|\mathcal{N}|$ rule of thumb\footnote{\url{https://pandapower.readthedocs.io/en/latest/estimation.html}}, and $3.2 \text{ to } 4.6|\mathcal{N}|$ measurements used by Hydro-Québec~\cite{4596318}) and report results only on scenarios that remain observable under WLS. Voltage angles are never measured. To train GENCO models, we use $p_{\text{mask}} = 0.2$ and $p_{\text{outlier}} = 0.1$ and minimize the loss shown in \cref{eq:loss}, with $\mathcal{L}_{\text{supervised}}$ being the mean $\ell_1$ norm between true and predicted bus and edge variables.


\paragraph{Baseline and evaluation.} We compare \genco against a Weighted Least Squares (WLS) baseline solved with the Gauss--Newton (GN) algorithm and outlier rejection, using the pandapower implementation~\cite{pandapower}. WLS receives the standard deviations of the measurements, giving it an advantage over \genco, which must implicitly infer measurement uncertainty from the training distribution. We evaluate each scenario using the bus-averaged mean absolute error. In rare under-determined scenarios, WLS can converge to a solution substantially different from the reference state while remaining consistent with the measurements, producing large errors that disproportionately affect the mean. We therefore use the median across scenarios of the bus-averaged mean absolute error as our primary metric, reducing the influence of such cases without requiring an arbitrary threshold to identify and exclude them. The baseline is evaluated only when the measurement-function Jacobian is full rank; comparisons are thus reported on that subset. Results for cases where the baseline does not converge are shown in \cref{fig:non-observable} in \cref{app:se}. \

\begin{figure}
    \centering
    \includegraphics[width=\linewidth]{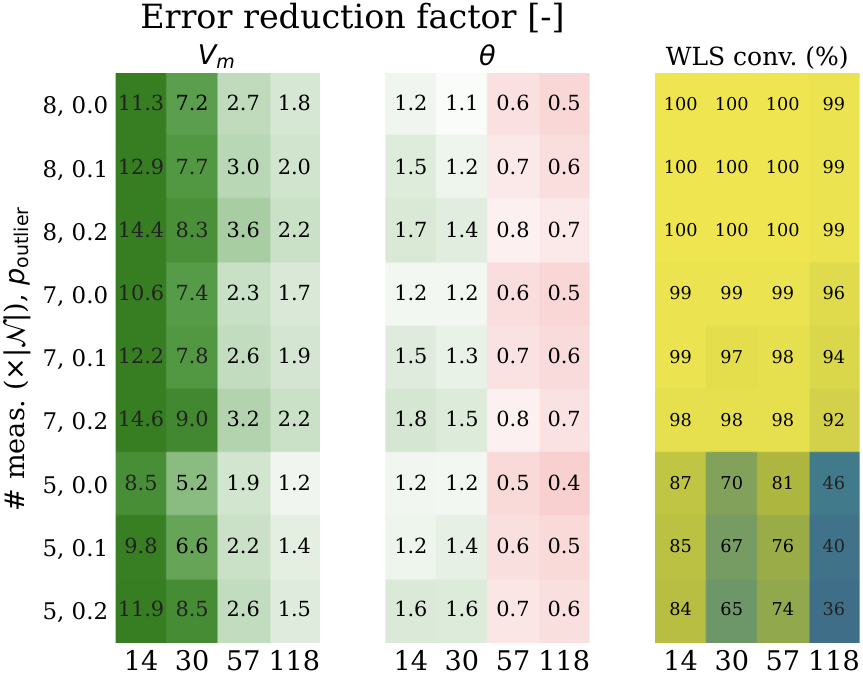}
    \caption{State-estimation accuracy for varying numbers of noisy measurements and outlier probabilities. The first two panels show the error ratio of \genco relative to the WLS baseline (ratio of their median mean absolute errors) per variable: values above $1$ (green) indicate that \genco is more accurate, whereas values below $1$ (red) favor the baseline. The last column represents the percentage of scenarios in which the baseline converges; \genco always returns an estimate. Columns correspond to the IEEE 14-, 30-, 57-, and 118-bus grids.}
    \label{fig:base-case}
\end{figure}

\paragraph{Robustness to sparse, noisy, and outlier measurements.}
We first evaluate \genco against WLS with varying numbers of measurements and outlier rates in \cref{fig:base-case}. One GENCO model is trained for each grid and evaluated across all measurement densities and outlier levels. Overall, \genco reconstructs states more accurately than WLS on the small grids---its median voltage magnitude error is up to an order of magnitude lower on the IEEE 14- and 30-bus cases---with a degradation on the larger grids. 
\\

As measurements become sparser and more corrupted, \genco remains robust, increasingly outperforming WLS on small grids and closing the gap with pandapower on larger grids. Note that the reported accuracy is computed only over scenarios where WLS converges, the number of which shrinks as measurements become sparser. \genco provides a complete estimator: unlike WLS, it returns an estimate even when the measurement configuration is insufficient for convergence. To assess the performance in cases where no WLS reference exists, \cref{fig:non-observable} reports the ratio between the errors of the non-converged and the converged subsets. Averaged across all sparsity levels, outlier rates, and grid-size settings, this ratio is $\approx 1.11$---a $\sim 10\%$ error increase on the cases where WLS yields no output at all. \genco thus degrades gracefully rather than failing abruptly, making it the more dependable estimator under marginal observability. Further improvements are required on larger grids, where \genco achieves the greatest gains in convergence robustness but falls short of WLS in accuracy.

\paragraph{Robustness to inaccurate grid parameters.}

\begin{figure}
    \centering
    \includegraphics[width=\linewidth]{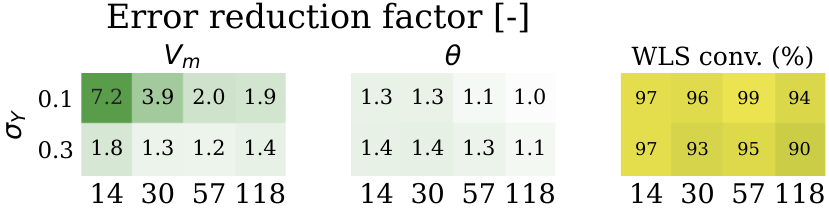}
    \caption{Robustness to inaccurate admittance parameters, evaluated across increasing admittance-noise levels $\sigma_Y$ (rows). The estimators are given only the base admittance matrix, before perturbation. The first two panels show the error ratio of \genco relative to the WLS baseline (ratio of median mean absolute errors) per state variable; values above $1$ (green) indicate that \genco is more accurate. The last panel depicts the convergence rate of the WLS solver; \genco always returns an estimate. Columns are the IEEE 14-, 30-, 57-, and 118-bus grids.} \label{fig:perturbation}
\end{figure}

While base grid parameters (e.g., admittance) are known to operators, they can drift throughout the day due to, e.g., temperature variations, which affect electrical resistance and the thermal expansion of transmission lines, resulting in line sag modulating the shunt capacitance. Consequently, the true grid parameters differ from the nominal parameters used by the state estimator. In \cref{fig:perturbation}, we evaluate \genco under grid parameter errors by perturbing the admittance with multiplicative factors sampled from a uniform distribution between $[1-\sigma_Y, 1+\sigma_Y]$. During training, measurements were consistent with the grid parameters, whereas during evaluation, the test set contains measurements that are generated using perturbed parameters, while the estimator receives the nominal admittance. Because \genco treats the grid physics as a soft prior rather than a hard constraint, it absorbs these discrepancies and keeps improving over the baseline as $\sigma_Y$ grows. In comparison, WLS enforces the now-incorrect model equations, causing the outlier-rejection procedure to incorrectly flag valid measurements and making non-convergence increasingly likely.\\

By only adding an auxiliary decoder to the architecture discussed in \cref{sec:model}, \genco can simultaneously estimate bus states and grid parameters. We demonstrate this approach by perturbing the branch resistance/reactance ($R$, $X$) parameters with multiplicative uniform noise on $[1-\sigma', 1+\sigma']$ (half-width $\sigma'_R = \sigma'_X = 0.2$
) prior to inference. \genco successfully denoises the input parameters, achieving a $30$--$50\%$ reduction in noise magnitude across all tested IEEE grids. \\ 

\begin{boxH}
\paragraph{Takeaways.}
\begin{itemize}
\item SE is a task where \genco can surpass classical estimators under challenging measurement conditions. It is strongest exactly where WLS is limited by its reliance on an accurate grid model and sufficient observability:
\begin{itemize}
\item It achieves \textbf{better estimates than WLS under sparse and corrupted measurements}, although scaling this advantage to larger grids remains future work.
\item It \textbf{always returns an estimate}, maintaining good accuracy in scenarios where WLS fails to converge.
\item By treating the grid model as a soft prior rather than a hard constraint, it remains \textbf{robust to inaccurate grid parameters}.
\end{itemize}
\item The same architecture naturally extends to joint state and parameter estimation, recovering system states while denoising branch parameters.
\end{itemize}
\end{boxH}

\subsection{Runtime and Performance Scaling}
\label{subsec:scaling}

The promise of neural solvers is to reduce time-to-solution compared with classical methods for batched PF and OPF problems. We therefore evaluate solver runtime and residuals across batches of PF and OPF instances and grid sizes, reflecting realistic utility workloads where large collections of scenarios are repeatedly solved for applications such as planning, time-series forecasting, contingency analysis, and uncertainty quantification, rather than a single isolated operating point. Since the datasets used in \cref{subsec:pf} and~\ref{subsec:opf} do not include DC solutions, we use \datakit datasets, which provide solutions obtained with all classical AC-PF/AC-OPF and DC-PF/DC-OPF solvers implemented through PowerModels~\cite{powermodels}. This enables comparison against the PowerModels implementations of AC-PF (Newton--Raphson), AC-OPF (IPOPT with MUMPS), DC-PF, and DC-OPF\footnote{A fully controlled comparison of runtime and residuals with other neural solvers from the literature remains challenging because prior work uses different benchmarking protocols and heterogeneous hardware and inputs.}. All GENCO models are trained from scratch for each task and grid.

\subsubsection{Batch Runtime Analysis Methodology}

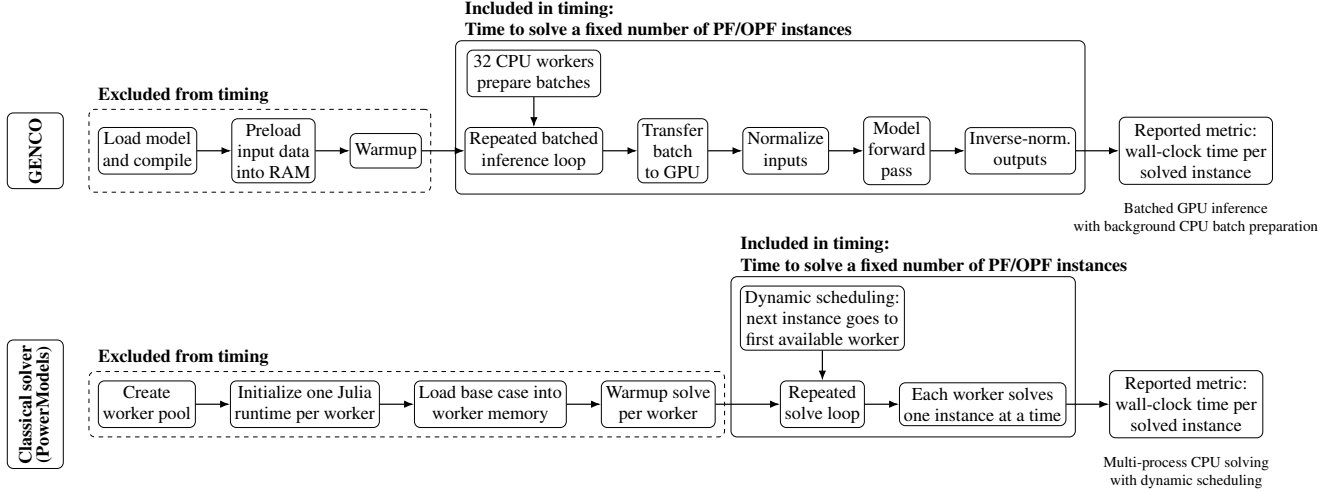
\begin{figure*}[t]
    \centering
    \resizebox{\textwidth}{!}{%
    \begin{tikzpicture}[
        >=Latex,
        font=\large,
        node distance=5mm and 8mm,
        every node/.style={align=center},
        box/.style={
            draw,
            rounded corners,
            minimum width=1.3cm,
            minimum height=8mm,
            inner sep=3pt,
            fill=white
        },
        sidebox/.style={
            draw,
            rounded corners,
            minimum width=3.1cm,
            minimum height=8mm,
            inner sep=3pt,
            fill=white
        },
        titlebox/.style={
            draw,
            rounded corners,
            font=\large\bfseries,
            minimum width=1.3cm,
            minimum height=9mm,
            inner sep=3pt,
            fill=white
        },
        metricbox/.style={
            draw,
            rounded corners,
            minimum width=1.3cm,
            minimum height=10mm,
            inner sep=4pt,
            fill=white
        },
        flow/.style={-Latex, thick},
        note/.style={font=\normalsize, align=center},
        stage/.style={font=\large\bfseries, align=left}
    ]
    
    \node[titlebox] (GENCOTitle) {\rotatebox{90}{\genco}};
    \node[titlebox, below=35mm of GENCOTitle]
(pmTitle) {\rotatebox{90}{\shortstack{Classical solver\\(PowerModels)}}};

    \node[box, right=8mm of GENCOTitle] (gLoad) {Load model\\and compile};
    \node[box, right=of gLoad] (gPreload) {Preload \\input data \\into RAM};
    \node[box, right=of gPreload] (gWarmup) {Warmup};
    \node[box, right=10mm of gWarmup] (gLoop) {Repeated batched\\inference loop};
    \node[sidebox, above=7mm of gLoop] (gCpu) {32 CPU workers\\prepare batches};
    \node[box, right=of gLoop] (gTransfer) {Transfer \\batch\\to GPU};
    \node[box, right=of gTransfer] (gNorm) {Normalize \\inputs};
    \node[box, right=of gNorm] (gForward) {Model\\forward \\pass};
    \node[box, right=of gForward] (gInverse) {Inverse-norm.\\outputs};
    \node[metricbox, right=10mm of gInverse] (gMetric) {Reported metric:\\wall-clock time per\\solved instance};
    \node[note, below=3mm of gMetric] (gNote) {Batched GPU inference\\with background CPU batch preparation};
    
    \node[box, right=8mm of pmTitle] (pPool) {Create\\worker pool};
    \node[box, right=of pPool] (pInit) {Initialize one Julia\\runtime per worker};
    \node[box, right=of pInit] (pCase) {Load base case into\\worker memory};
    \node[box, right=of pCase] (pWarmup) {Warmup solve\\per worker};
    \node[box, right=15mm of pWarmup] (pLoop) {Repeated\\solve loop};
    \node[sidebox, above=7mm of pLoop] (pSched) {Dynamic scheduling:\\next instance goes to\\first available worker};
    \node[box, right=of pLoop] (pSolve) {Each worker solves\\one instance at a time};
    \node[metricbox, right=10mm of pSolve] (pMetric) {Reported metric:\\wall-clock time per\\solved instance};
    \node[note, below=3mm of pMetric] (pNote) {Multi-process CPU solving\\with dynamic scheduling};
    
    \draw[flow] (gLoad) -- (gPreload);
    \draw[flow] (gPreload) -- (gWarmup);
    \draw[flow] (gWarmup) -- (gLoop);
    \draw[flow] (gLoop) -- (gTransfer);
    \draw[flow] (gTransfer) -- (gNorm);
    \draw[flow] (gNorm) -- (gForward);
    \draw[flow] (gForward) -- (gInverse);
    \draw[flow] (gInverse) -- (gMetric);
    \draw[flow] (gCpu.south) -- (gLoop.north);
    
    \draw[flow] (pPool) -- (pInit);
    \draw[flow] (pInit) -- (pCase);
    \draw[flow] (pCase) -- (pWarmup);
    \draw[flow] (pWarmup) -- (pLoop);
    \draw[flow] (pLoop) -- (pSolve);
    \draw[flow] (pSolve) -- (pMetric);
    \draw[flow] (pSched.south) -- (pLoop.north);
    
    \node[draw, dashed, rounded corners, inner sep=6pt, fit=(gLoad)(gPreload)(gWarmup)] (gExcluded) {};
    \node[stage, above left=0mm and -1mm of gExcluded.north west, anchor=south west] {Excluded from timing};
    \node[draw, rounded corners, inner sep=6pt, fit=(gLoop)(gCpu)(gTransfer)(gNorm)(gForward)(gInverse)] (gIncluded) {};
    \node[stage, above left=0mm and -1mm of gIncluded.north west, anchor=south west] {Included in timing: \\Time to solve a fixed number of PF/OPF instances};
    
    \node[draw, dashed, rounded corners, inner sep=6pt, fit=(pPool)(pInit)(pCase)(pWarmup)] (pExcluded) {};
    \node[stage, above left=0mm and -1mm of pExcluded.north west, anchor=south west] {Excluded from timing};
    \node[draw, rounded corners, inner sep=6pt, fit=(pLoop)(pSched)(pSolve)] (pIncluded) {};
    \node[stage, above left=0mm and -1mm of pIncluded.north west, anchor=south west] {Included in timing: \\Time to solve a fixed number of PF/OPF instances};
    
    \end{tikzpicture}
    }
    \caption{Benchmark pipelines used for runtime comparison. Top: for \genco, we measure steady-state throughput of batched GPU inference after model setup, sample preloading, and warmup. Bottom: for classical solvers, we measure steady-state throughput of a dynamically scheduled multi-process CPU solver pool after worker initialization, case loading, and warmup. Both report wall-clock time per solved instance, but they exploit parallelism differently: \genco through batching on a single GPU, and the classical solvers through multiple independent worker processes.}
    \label{fig:benchmark_pipeline_diagram}
    \end{figure*}

To ensure a fair runtime comparison, we evaluate both solvers under their best achievable throughput configurations while controlling for factors unrelated to the solution process. In particular, four aspects need to be carefully designed: (i) the evaluation metric and parallelization mode, (ii) benchmark scenarios, (iii) timing scope, and (iv) compute hardware. \Cref{fig:benchmark_pipeline_diagram} summarizes the benchmark pipelines and timing boundaries; complete implementation details and parameters are provided in \cref{app:runtime_details}.

\paragraph{Evaluation metric and parallelization mode.}
We measure the \emph{amortized per-instance runtime}, defined as the total wall-clock time required to solve a large batch of PF/OPF instances divided by the number of solved instances. 
Neural and classical approaches exploit different forms of parallelism: \genco performs batched GPU inference, whereas the classical solvers implemented through PowerModels exploit CPU parallelism through multiple workers, with each worker solving instances sequentially on a single CPU core.
We thus carefully design our experiments so that both solver classes are evaluated at their maximum throughput: For each grid and solver, we select the batch size or number of workers that minimizes our runtime metric, ensuring that neither approach is evaluated below its optimal throughput configuration\footnote{Note that previous runtime benchmarks of neural solvers evaluated classical baselines without the CPU parallelism used in our experiments, which can favor neural solvers in runtime comparisons~\cite{NEURIPS2025_d000ef56, arowolo2025, piloto2024}.}.

\paragraph{Benchmark scenarios.}
We focus on batched scenario analysis in which evaluated scenarios are typically obtained in memory by applying small changes (e.g., to topology or load) to a given base network, reflecting use cases such as time-series analysis, contingency screening, and sensitivity studies. A challenge is that classical solver runtime depends on scenario difficulty, through, e.g., the number of iterations required for convergence, whereas neural solvers' inference is largely insensitive to scenario difficulty. The runtime of classical solvers is also affected by failed solves, for which the runtime depends on the maximum iteration limit. To avoid favoring neural solvers through harder or non-convergent scenarios, and to make the experiment reproducible without choosing a scenario distribution, the timed main protocol repeatedly solves the fixed base case of each grid (already resident in memory). To complement this analysis, \cref{app:runtime_details} reports a second \emph{from-disk} protocol for heterogeneous grid snapshots, such as archived scenarios or externally generated scenarios, where each scenario is an independent \datakit scenario loaded from disk and data loading is part of the measured workload.\\

To ensure a reliable throughput estimate, all solvers are evaluated on a sufficiently large number of repeated cases to reduce measurement noise from non-deterministic hardware effects, such as OS scheduling, CPU frequency fluctuations, and background processes. The number of repetitions is chosen to provide stable timing measurements (with at least 20~s of total solve time in the best-performing configuration) while remaining computationally tractable, as this evaluation is repeated across batch sizes and CPU worker counts to identify the optimal configuration.

\paragraph{Timing scope.}
We measure steady-state solver throughput rather than end-to-end pipeline latency, excluding one-time infrastructure overheads unrelated to the repeated PF/OPF solution process. These overheads depend on implementation choices and system factors rather than the numerical efficiency of the solver.
Thus, for a fair comparison, we exclude some components for both neural and classical solvers, i.e., worker initialization (PowerModels solver workers and \genco data-loading workers), compilation and warmup (Julia compilation and model compilation), and initial disk-to-RAM loading of the base network.

\paragraph{Compute hardware.}
A fair comparison requires CPU and GPU hardware from the same node generation and performance class\footnote{Energy consumption and hardware cost are also relevant metrics but are difficult to compare fairly.}. In this study, we benchmark the solvers using two high-end devices, an AMD EPYC 9634 CPU with 84 cores and an NVIDIA H100 SXM5 GPU with 80GB HBM3 memory, both with a node size of 5nm and release date of late 2022. \\

\subsubsection{PF Scaling with Grid Size: Runtime and Accuracy}
 
We first study the trade-off between runtime and accuracy of the different solvers for PF.\\

\begin{figure}
    \centering
    \includegraphics[width=\linewidth]{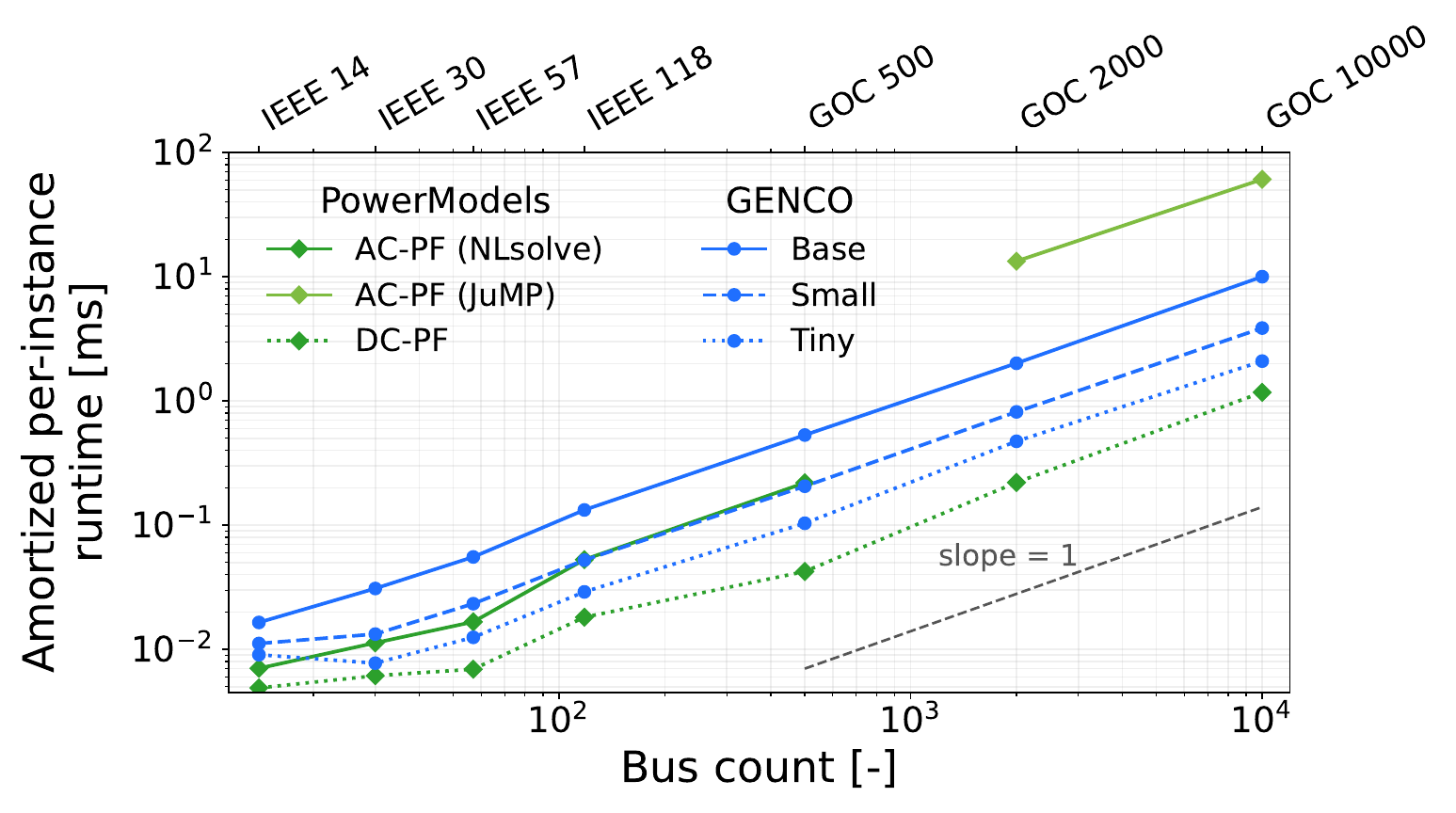}
    \caption{Amortized per-instance runtime versus grid size for \genco and PowerModels AC-PF and DC-PF at the best batch size or worker count. Scaling exponents fitted on the four largest grids (two for AC-PF) are comparable across all methods, ranging from 0.94 to 0.97.}
    \label{fig:gridsize_runtime_pf}
\end{figure}

\begin{figure}
    \centering
    \includegraphics[width=1\linewidth]{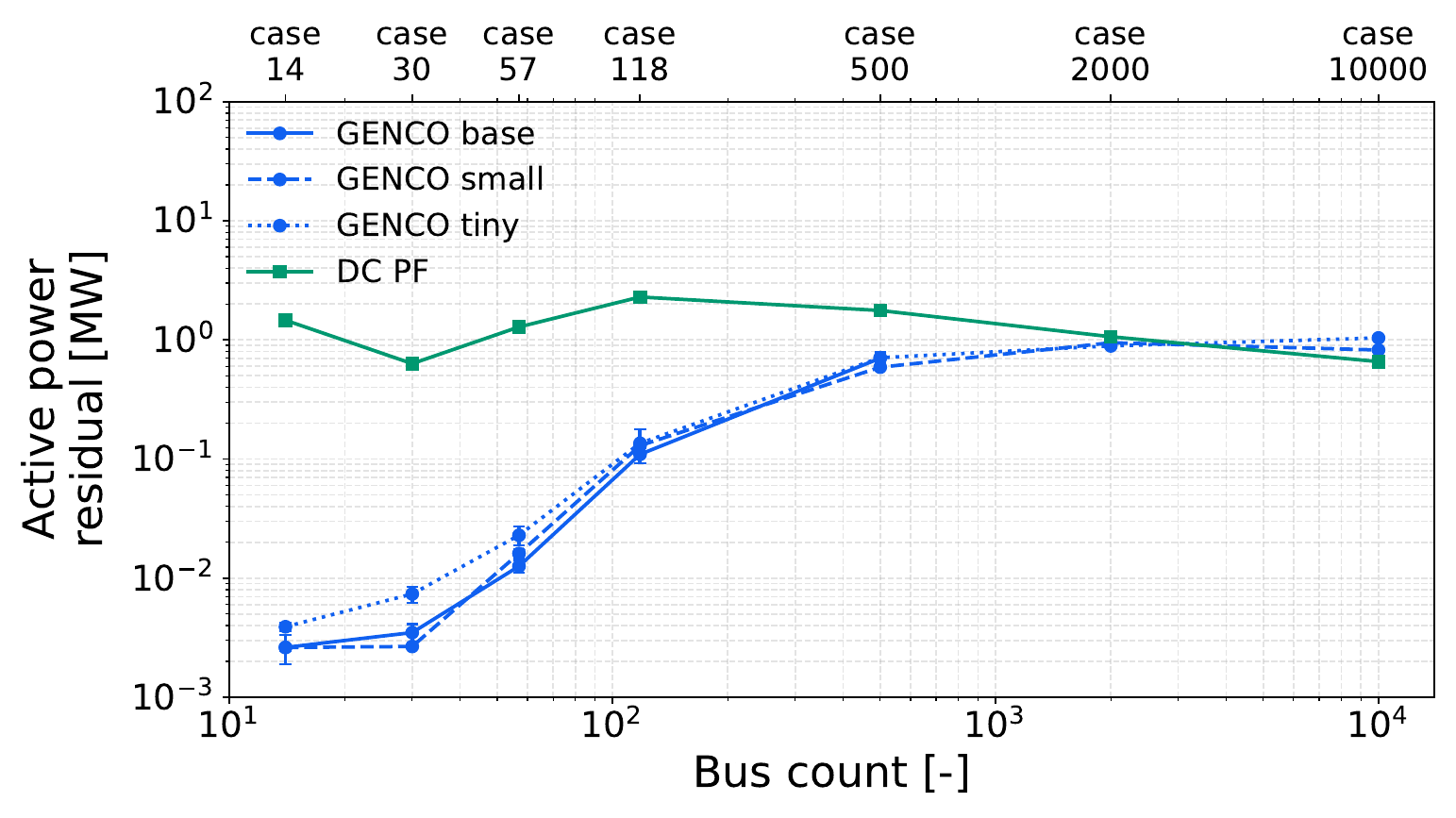}
    \caption{Mean active power-balance residuals versus grid size for \genco and PowerModels DC-PF.}
   \label{fig:gridsize_vs_residuals_pf}
\end{figure}

\paragraph{Runtime and accuracy for different grid sizes.}
The amortized per-instance runtime of \genco, AC-PF, and DC-PF as a function of bus count at the optimal batch size (\genco) or worker count (for the PowerModels-based classical solvers) is depicted in \cref{fig:gridsize_runtime_pf}. The runtime of all \genco versions (Base, Small, and Tiny) scales close to linear with grid size. For AC-PF, we selected the fastest converging solver available in PowerModels: the NLsolve-based solver for grids up to 500 buses and the JuMP-based solver for larger grids, as indicated by the disconnected lines. For large grids, the slopes of all solvers are similar, while for small networks AC-PF and DC-PF runtimes are nearly constant as multiprocessing overhead dominates. Active power-balance residuals are shown in \cref{fig:gridsize_vs_residuals_pf}. PowerModels AC-PF residuals ($10^{-11}$--$10^{-8}$~MW) are omitted for readability, while DC-PF remains near 1~MW. \genco increases from about $10^{-3}$~MW on the smallest IEEE grids to about 1~MW on GOC~2,000 and GOC~10,000. Its residuals are substantially lower than for DC-PF through GOC~500, slightly lower on GOC~2,000, and comparable to or slightly above for GOC~10,000, where Small outperforms Tiny; Base is evaluated only through GOC~500.\\

\paragraph{\genco's utility for PF.} \cref{tab:pf_selected_genco_tradeoff} summarizes the PF speed--accuracy trade-off, focusing on \genco Tiny because it provides the highest throughput while retaining useful solution accuracy: For a learned PF surrogate to be compelling, it must first provide a meaningful speedup over AC-PF, whose solutions are more accurate; modest gains may not offset the cost of training, maintaining, and deploying the neural solver. This criterion is not met on the small grids up to 500 buses, and the approximately $2\times$ speedup on GOC~500 remains marginal. In contrast, \genco Tiny is approximately $28$--$29\times$ faster than AC-PF on GOC~2,000 and GOC~10,000, making the learned approach increasingly attractive at scale. On these grids, \genco provides a good alternative to DC-PF: unlike DC-PF, it predicts a complete AC state, including voltage magnitudes and reactive generation, and therefore supports downstream voltage- and reactive-power-based analyses. For this reason, \genco can remain useful even when it is moderately slower or has slightly larger active power-balance residuals than DC-PF. On GOC~2,000 and GOC~10,000, it achieves similar residuals while being approximately $2\times$ slower than DC-PF. Overall, \genco Tiny is compelling for large-scale PF, where it substantially outperforms AC-PF and provides a complete AC state at a modest slowdown relative to DC-PF; on grids through GOC~500, AC-PF remains preferable.

\begin{table}
\centering
\scriptsize
\setlength{\tabcolsep}{3pt}
\renewcommand{\arraystretch}{1.1}
\begin{tabular}{@{}l c c c >{\raggedright\arraybackslash}p{0.35\linewidth}@{}}
\toprule
Grid
& \shortstack{Speedup\\vs AC-PF}
& \shortstack{DC-PF/\genco\\residual}
& \shortstack{Speedup\\vs DC-PF}
& Utility \\
\midrule
  14 & $0.8\times$ & $374.0\times$ & $0.5\times$ & 
  \multirow{5}{=}{None. Slower than AC-PF, which should be preferred since it provides more accurate solutions.} \\
  30 & $1.5\times$ & $85.8\times$ & $0.8\times$ &  \\
  57 & $1.3\times$ & $56.1\times$ & $0.6\times$ &  \\
 118 & $1.8\times$ & $16.9\times$ & $0.6\times$ &  \\
 500 & $2.1\times$ & $2.49\times$ & $0.4\times$ &  \\
\midrule
2000 & $28.2\times$ & $1.19\times$ & $0.5\times$ &
\multirow{6}{=}{\raggedright
Much faster than AC-PF; provides complete solutions with near-DC-PF active power-balance residuals at only 2× the runtime of DC-PF.} \\
10000 & $29.1\times$ & $0.63\times$ & $0.6\times$ & \\
& & & & \\
& & & & \\
& & & & \\
& & & & \\
\bottomrule
\end{tabular}
\caption{Scaling performance summary for \genco on Power Flow. Speedups $>$1 mean faster; the Utility column summarizes the recommended use of \genco relative to AC-PF and DC-PF for each grid-size. Results are for \genco Tiny for all grid sizes as it provides the best speedups and reasonable performance.}
\label{tab:pf_selected_genco_tradeoff}
\end{table}

\subsubsection{OPF Scaling with Grid Size: Runtime and Accuracy}

\begin{figure}
    \centering
    \includegraphics[width=\linewidth]{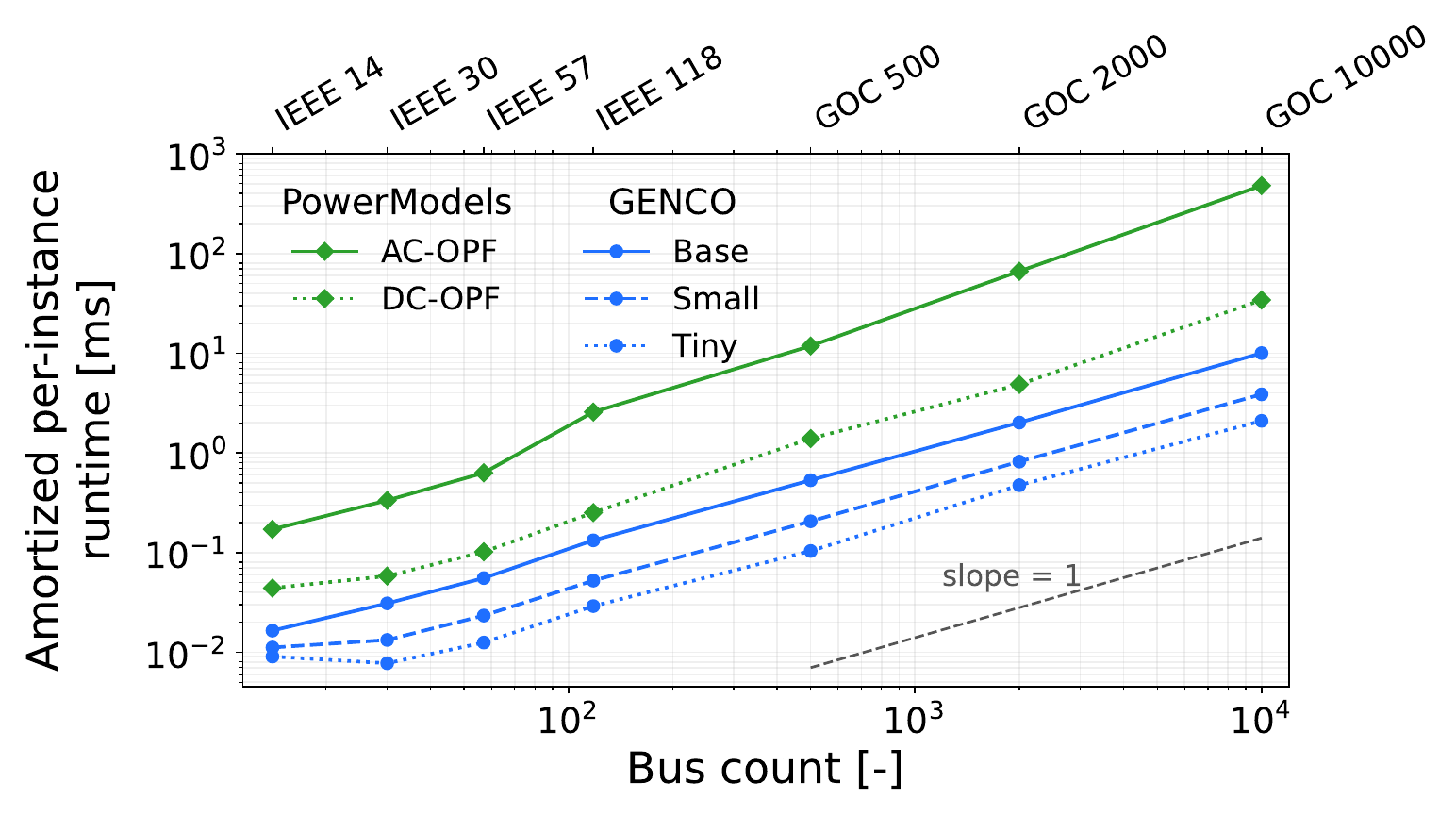}
    \caption{Amortized per-instance runtime versus grid size for \genco and PowerModels AC-OPF and DC-OPF solvers. Results are reported for the optimal batch size for \genco and the best worker count for PowerModels. Four-largest-grid scaling exponents are 1.18 for AC-OPF, 1.09 for DC-OPF, and 0.97/0.97/0.98 for \genco Base/Small/Tiny.}
    \label{fig:gridsize_runtime_opf}
\end{figure}

The amortized per-instance runtime as a function of grid size for AC-OPF, DC-OPF, and \genco is shown in \cref{fig:gridsize_runtime_opf}. The runtime values for \genco are independent of the task and thus identical to \cref{fig:gridsize_runtime_pf}. However, OPF is substantially harder to solve for classical solvers, resulting in a runtime advantage for \genco across all grid sizes.\\


\paragraph{\genco's utility for OPF.} The performance of \genco Small (which offers the best tradeoff between runtime, optimality, and feasibility) across grid sizes is summarized in \cref{tab:opf_selected_genco_tradeoff}. \genco is substantially faster than AC-OPF and DC-OPF on every grid, reaching an $81.1\times$ speedup over AC-OPF on GOC~2,000. It also produces more feasible and cost-effective solutions than DC-OPF, although these gains generally decrease with grid size. Finally, \genco predicts quantities unavailable from DC-OPF, including voltage magnitudes and reactive power setpoints. Overall, \genco offers a compelling alternative to DC-OPF. Detailed optimality and constraint-violation results for \genco Base and Small are reported in \cref{tab:opf_scaling} in the Appendix.\\

\begin{table}
\centering
\scriptsize
\setlength{\tabcolsep}{4pt}
\renewcommand{\arraystretch}{1.2}
\begin{tabular}{@{}l c c c c@{}}
\toprule
Grid
& \shortstack{Speedup\\vs AC-OPF}
& \shortstack{Opt. gap reduction\\vs DC-OPF}
& \shortstack{Violation reduction\\vs DC-OPF}
& \shortstack{Speedup\\vs DC-OPF} \\
\midrule
  14 & $15.3\times$ & $55.2\times$ & $86.0\times$ & $3.94\times$ \\
  30 & $25.1\times$ & $31.8\times$ & $72.4\times$ & $4.38\times$ \\
  57 & $27.0\times$ & $6.28\times$ & $4.67\times$ & $4.36\times$ \\
 118 & $49.1\times$ & $2.85\times$ & $2.40\times$ & $4.81\times$ \\
 500 & $57.4\times$ & $3.20\times$ & $10.3\times$ & $6.76\times$ \\
2000 & $81.1\times$ & $1.85\times$ & $6.70\times$ & $5.94\times$ \\
\bottomrule
\end{tabular}
\caption{Performance scaling summary for \genco Small on Optimal Power Flow; Speedups $>$1 mean faster; optimality and feasibility improvements are ratios of DC-OPF to \genco mean metrics.}
\label{tab:opf_selected_genco_tradeoff}
\end{table}

\begin{boxH}
\paragraph{Takeaways.}

\begin{itemize}


\item For PF, \genco starts providing a valuable alternative to DC-PF for large grids ($\geq$2000 buses) by predicting the full AC operating state, including voltage magnitudes and reactive power, while achieving active power-balance residuals comparable to DC-PF, with up to \maxpfspeedups speedup over Newton--Raphson and only a \dcpfspeedupovergenco runtime relative to DC-PF.

\item For OPF, \genco is compelling across all evaluated grid sizes: Small is $15$--$81\times$ faster than AC-OPF and $4$--$7\times$ faster than DC-OPF, while reducing DC-OPF optimality gaps by $1.9$--$55\times$ and feasibility violations by $2.4$--$86\times$.

\end{itemize}
\end{boxH}

\subsection{Robustness \& Generalization}

\label{subsec:generalization}

A key requirement for any power-system solver is reliable performance across (1) diverse topologies, (2) operating conditions, and (3) network sizes. Classical solvers are well established across these dimensions, with their main limitation being occasional convergence failures in challenging cases~\cite{tostado2021solving}. In contrast, robustness across these dimensions remains a central challenge for neural solvers, which typically require retraining when topology or operating regimes change. Achieving high accuracy across diverse scenarios is therefore essential for the scalability and practical deployment of neural solvers. We evaluate \genco along three axes: generalization to topology perturbations beyond the training distribution (\cref{subsubsec:top_perturbation}), robustness under stressed operating conditions such as thermal overloads and voltage violations (\cref{subsubsec:robustness}), and transfer to previously unseen grids (\cref{subsubsec:transfer}). 

\subsubsection{Generalization to Topology Perturbations}

\label{subsubsec:top_perturbation}

\paragraph{Experiment setup.} \genco is evaluated under contingency levels that match or exceed those seen during training to assess robustness to topology perturbations. Specifically, \genco Base is trained on 140,000 AC-PF samples\footnote{Using \datakit, we generated 200,000 samples, of which 176,415 converged; 80\% of converged samples were used for training.} from the Texas 2000-bus system with up to \textbf{N-2} contingencies, corresponding to the removal of up to two branches and/or generators. The test set contains 9,000 cases for each contingency level from N-1 to N-20. Components are removed uniformly across the grid, and islanding cases are filtered out. Since perturbations are applied after computing the base dispatch (\cref{subsubsec:gen_setpoints}), the resulting operating points may exhibit voltage violations, branch overloads, large angle differences, and reactive-power limit violations.

\begin{figure}
\centering
\includegraphics[width=\linewidth]{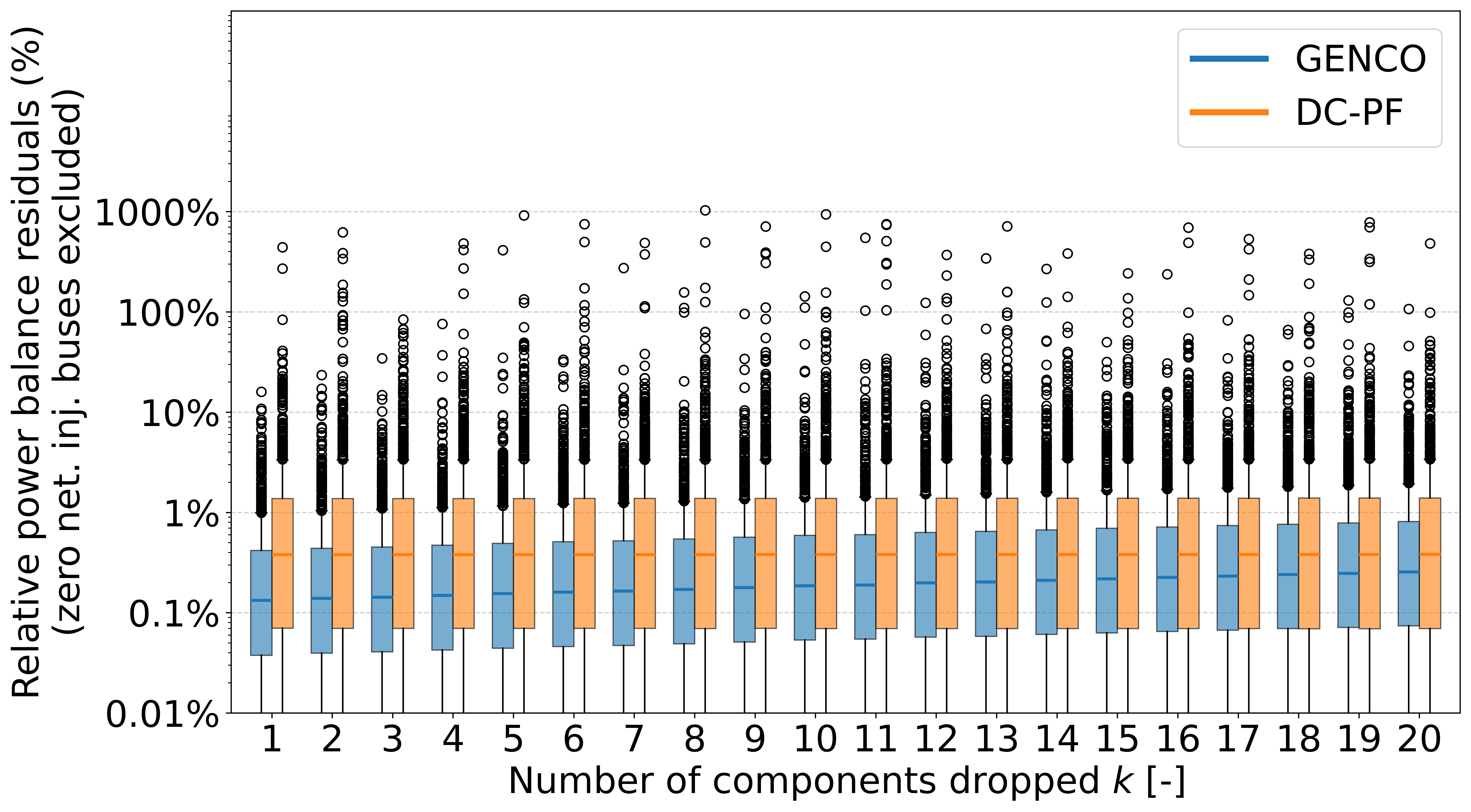}
\caption{Bus-level active power-balance residuals (excluding zero net injection buses) for \genco and DC-PF under N-$k$ contingencies ($k=1,\dots,20$) on the Texas 2000-bus system. \genco is trained only on N-2 perturbations. Whiskers indicate 1.5$\times$IQR.}
\label{fig:performance_k_1_20}
\end{figure}

\paragraph{Results.} We evaluate bus-level relative active power-balance residuals, as opposed to absolute residuals, which mask large errors at low-injection buses. The bus-level relative active power-balance residuals are defined as the active power-balance residuals divided by the magnitude of the net bus injection\footnote{We filter out buses with zero net injection for which such a metric cannot be computed. These represent ~25\% of all buses across the entire dataset, and we report absolute metrics separately for them in \cref{appendix:zeroinj}}. This normalization approach makes the metric comparable across buses with vastly different injection levels. \genco consistently outperforms DC-PF across all contingency levels, including those beyond the training distribution, i.e., for N-k with $k > 2$ (\cref{fig:performance_k_1_20}). Residuals increase with outage count, but degradation remains gradual: \genco achieves 2.85$\times$ lower median residuals than DC-PF for N-1 contingencies and 1.5$\times$ lower median residuals for N-20.\\

\begin{figure}
\centering
\includegraphics[width=0.7\linewidth]{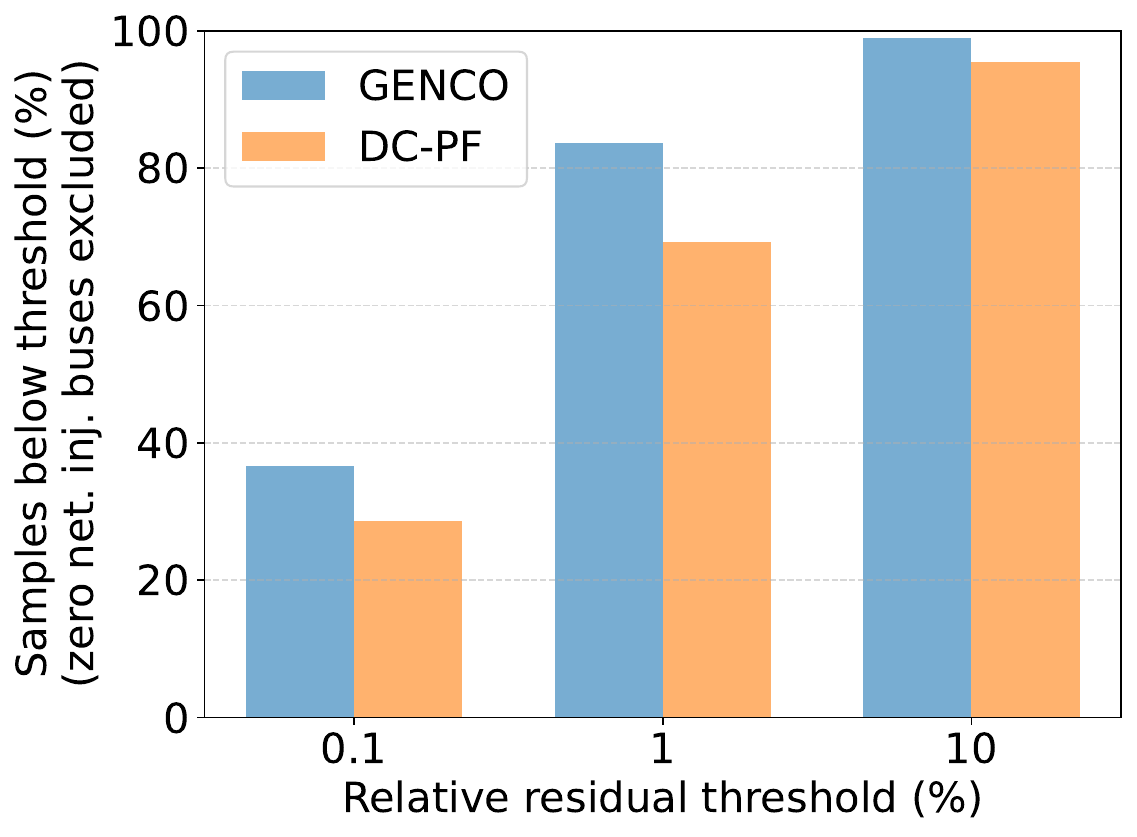}
\caption{Percentage of samples below relative active power-balance residual thresholds at $k=10$ (excluding zero net injection buses). \genco achieves a larger share of low-residual predictions than DC-PF across all reported thresholds.}
\label{fig:threshold_share_relative_k10}
\end{figure}

Outlier residuals displayed in \cref{fig:performance_k_1_20} are high for both DC-PF and \genco (between 9.2 and 9.8 percent of the buses across all contingency orders and solvers). Thus, we report the fraction of buses with residuals below practically relevant thresholds for $k=10$ in \cref{fig:threshold_share_relative_k10}: \genco outperforms DC-PF at all levels, including 1\% (83.65\% vs 69.30\%). The same trend holds for absolute thresholds (\cref{fig:abs_threshold_k10}, \cref{appendix:top}), e.g., at 1~MW (97.69\% vs 85.78\%). Further details about the 1\% threshold fraction across contingencies can be found in \cref{fig:rel_threshold_vs_k} in \cref{appendix:top}.

\subsubsection{Robustness to Out-of-Operating-Limit Scenarios}
\label{subsubsec:robustness}

\paragraph{Experiment setup.} Next, we evaluate GENCO's performance on operating points that violate thermal or voltage limits. These out-of-operating-limit elements are rare: only $0.0071$\% of branches are overloaded, corresponding to 1,979 out of a total of 28,052,640 branches in the dataset. Thus, such cases are underrepresented during model training relative to branches operating within nominal limits. For this experiment, we use the same \genco Base model as in the previous experiment, trained and tested on the N-2 contingency dataset.

\begin{figure}
\centering
\includegraphics[width=0.9\linewidth]{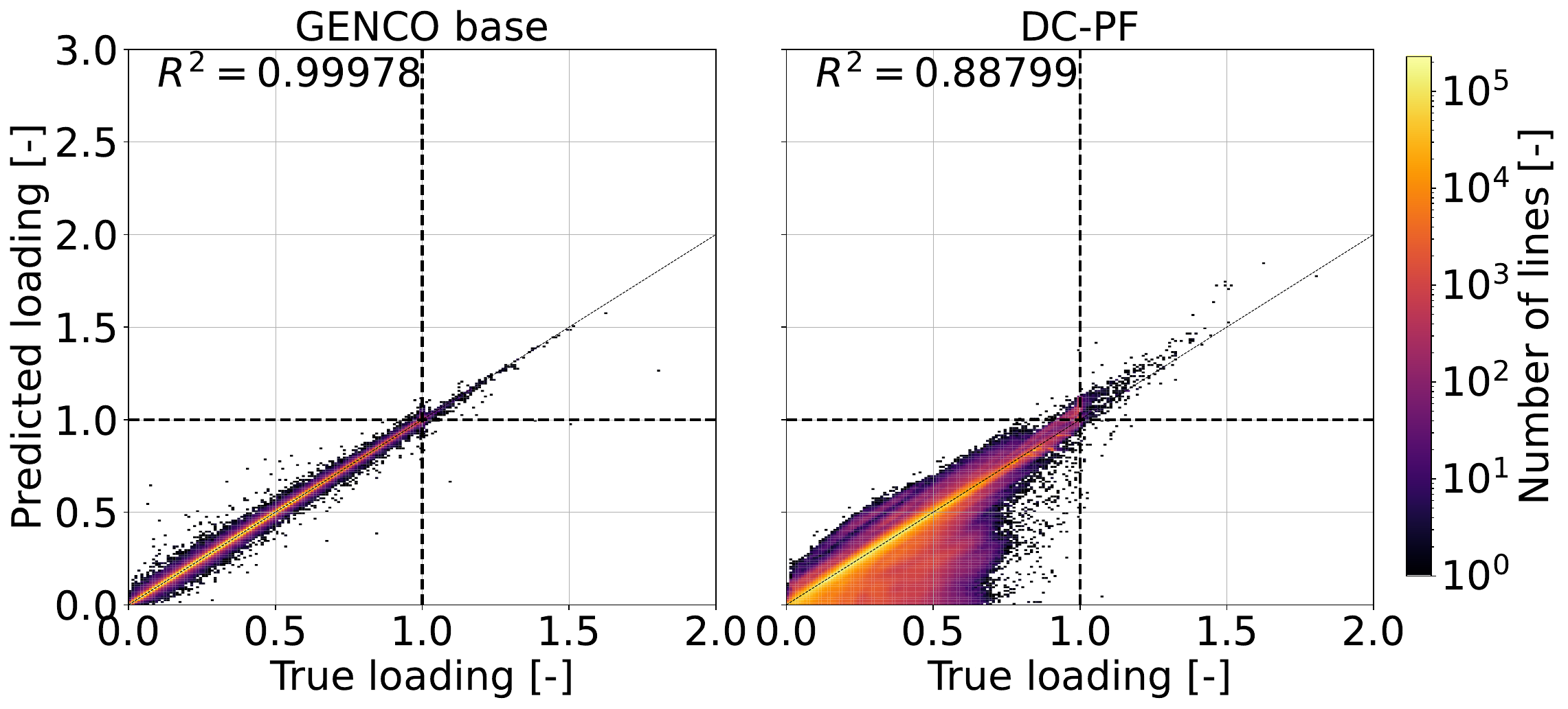}
\caption{Relative branch loading predictions obtained with \genco base (left) and DC-PF (right) compared to AC-PF loadings. Values above 1.0 indicate thermal overloads.}
\label{fig:dc_loading}
\end{figure}

\paragraph{Results.} The comparison of \genco and DC-PF branch loading predictions against AC-PF is depicted in \cref{fig:dc_loading}. Branch loadings are normalized by their thermal limits, such that values above 1.0 correspond to overloads. While DC-PF exhibits substantial dispersion, \genco remains closely aligned with the diagonal across both normal and overloaded operating regimes, as quantified by the coefficient of determination.\\

\begin{figure}
\centering
\includegraphics[width=0.9\linewidth]{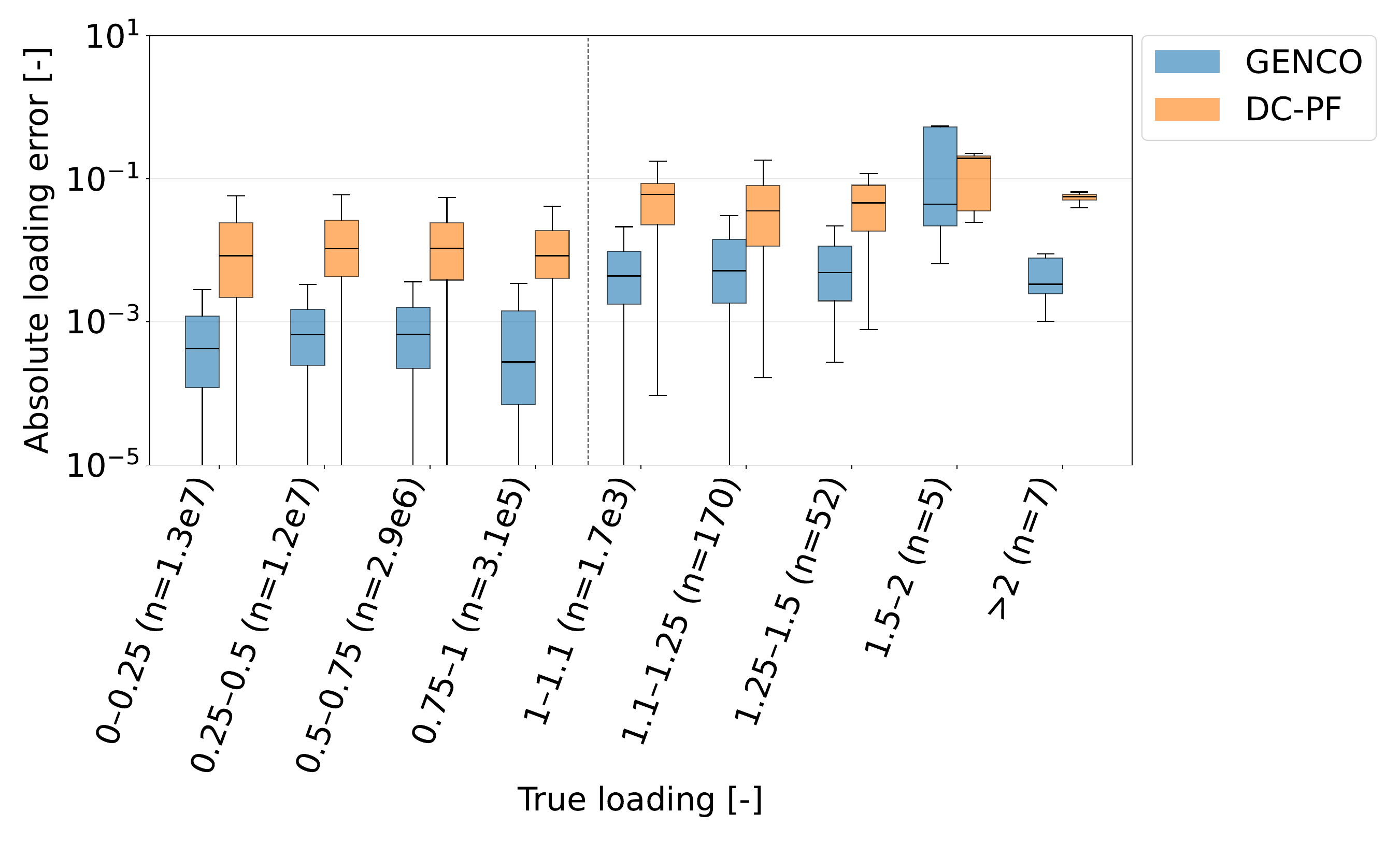}
\caption{Absolute branch loading prediction error as a function of the true loading level. $n$ indicates the number of test samples in each bin.}
\label{fig:loading_boxplot}
\end{figure}

To further assess performance near and beyond thermal limits, \cref{fig:loading_boxplot} reports absolute loading errors as a function of the true loading level. \genco consistently outperforms DC-PF across all loading bins, including overload conditions that are strongly underrepresented in the training data. For lines with true loadings between 1.0 and 1.1, \genco achieves a mean absolute error of 0.004, enabling accurate identification of overloaded components. \\

Since DC-PF does not predict voltage magnitudes, voltage-related results are reported separately in \cref{appendix:voltage_violations}. There, we show that \genco accurately captures both normal and out-of-limit voltage regimes despite the scarcity of voltage violations in the training data. These results indicate that explicit oversampling of out-of-limit operating points is not required for accurate overload prediction.\\

\subsubsection{Transfer to Unseen Grids and Data Efficiency}
\label{subsubsec:transfer}
\begin{figure} 
\centering 
\includegraphics[width=0.85\linewidth]{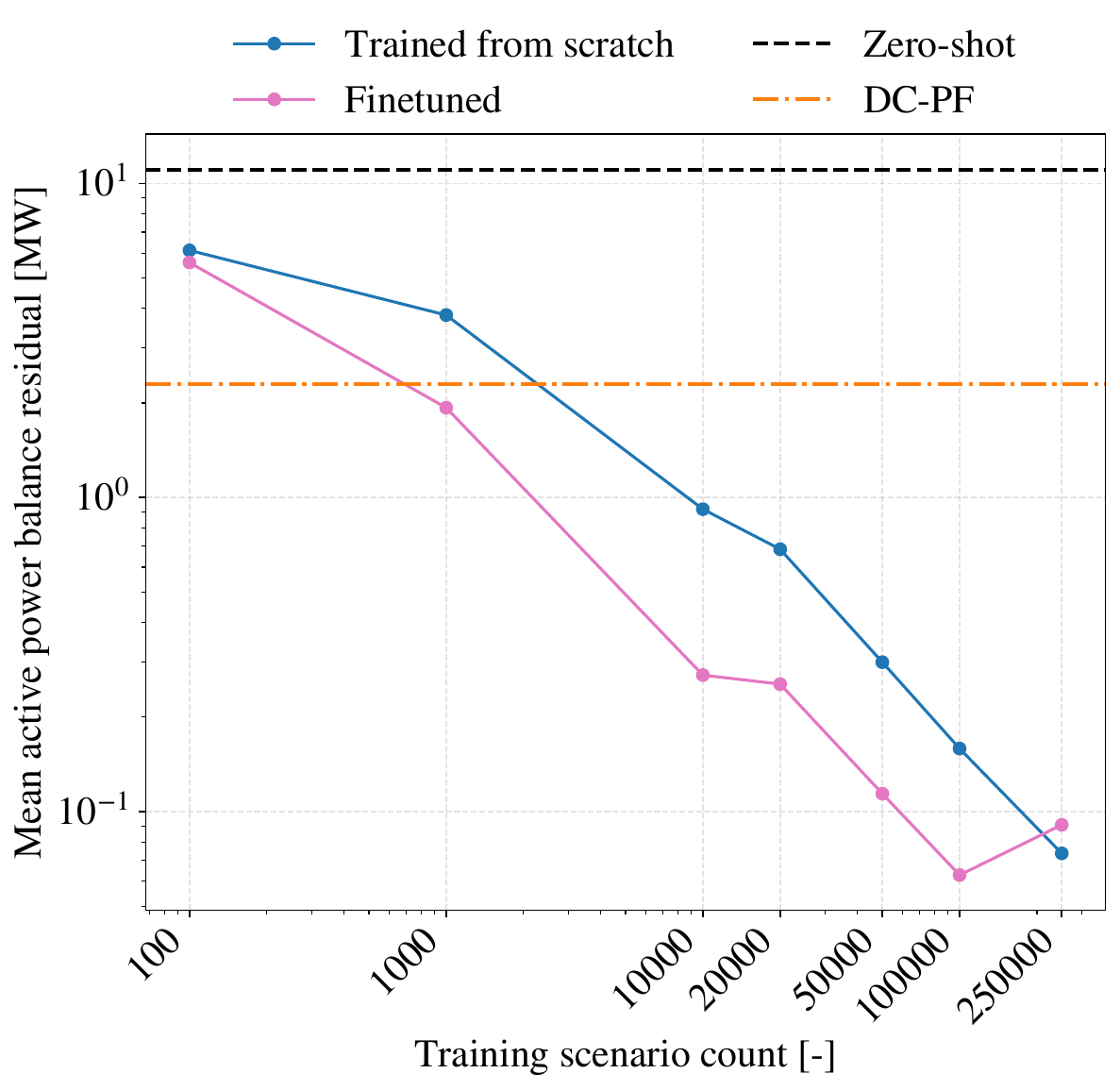} 
\caption{Mean active power-balance residuals on IEEE 118 for fine-tuned pretrained models and models trained from scratch using datasets of varying size. DC-PF and the zero-shot pretrained model (without fine-tuning) are shown as baselines.} 
\label{fig:transfer} 
\end{figure}

An important objective for neural PF solvers is rapid adaptation to entirely new grids using only a limited number of grid-specific samples. This capability would reduce deployment costs, since acquiring AC-PF datasets and training neural solvers from scratch for every new topology is time-consuming and expensive. We therefore investigate the transferability of pretrained \genco models to unseen grids under different levels of fine-tuning. To increase topology diversity during pretraining, we generate subgrids from existing grids and evaluate transfer to unseen topologies.

\paragraph{Experiment setup.}
Only around 50 different grids for which PF converges are available in PGLib. To increase the number of pretraining topologies, we extract subgrids from base grids using Metropolis-Hastings random walks. Power flows on cut lines are aggregated into dummy generators or loads depending on their sign. Using this approach, we generate 100 subgrids from 10 PGLib base cases, with sizes ranging from 100 to 1000 buses. We generate 9,000 PF samples per subgrid using \datakit, yielding 900,000 scenarios. Validation is performed on 10 unseen subgrids. \\ 

We then fine-tune the pretrained GENCO model on the IEEE 118-bus system using varying numbers of grid-specific samples. The resulting fine-tuned models are compared with models trained from scratch using the same datasets\footnote{The only difference between the two models is their weight initialization: the fine-tuned model starts from pretrained weights, whereas the scratch model uses random initialization.}.

\paragraph{Results.}
The active power-balance residuals are shown in \cref{fig:transfer}, together with those obtained with DC-PF and with the pretrained GENCO used in a zero-shot manner, i.e., without any fine-tuning on the test grid. The pretrained model has poor zero-shot generalization, with an average residual of 11.07 MW compared with 2.30 MW for DC-PF. However, with only 1,000 grid-specific samples, fine-tuning already outperforms DC-PF (1.93 MW), whereas training from scratch remains worse (3.81 MW). With 10,000 samples, fine-tuning further reduces the residual to 0.27 MW, compared with 0.92 MW when training from scratch. At 250,000 samples, scratch training surpasses fine-tuning, which slightly degrades under extensive fine-tuning.\\

Overall, these results show that topology-diverse pretraining can accelerate adaptation to unseen grids when grid-specific data are limited. However, with sufficient data from a single grid, training from scratch can eventually outperform pretraining. Further work is needed to reduce the number of fine-tuning samples required for true few-shot adaptation.\\

\begin{boxH}
\paragraph{Takeaways.}

\begin{itemize}

\item \genco generalizes to contingency levels up to N-20 despite training only on N-2, achieving 2.85$\times$ lower median residuals than DC-PF for N-1 contingencies and 1.5$\times$ lower median residuals for N-20.

\item \genco remains robust under out-of-operating-limit scenarios despite training on fewer than $0.01$\% such out-of-limit elements, outperforming DC-PF for line-loading prediction and detecting voltage violations beyond the capabilities of DC-PF.

\item A pretrained \genco transfers effectively to new grids, with only 1,000 grid-specific samples sufficient to outperform DC-PF on IEEE 118, although zero-shot generalization remains an open challenge.

\end{itemize}
\end{boxH}

\subsection{Validation on Real Data from the Hydro-Québec Grid}
\label{subsec:hq1200}

Finally, we evaluate \genco on the real Hydro-Qu\'ebec HQ1200 transmission network, extending the validation beyond the synthetic IEEE/GOC topologies. All experiments are done for the PF task. HQ1200 comprises 1,200 buses and differs markedly from these benchmarks through its V-shaped topology, with long corridors linking northern generation to southern load. These long-range dependencies make HQ1200 a non-trivial test for \genco.\\

As we only have a limited amount of SCADA data samples---roughly one year of measurements at 30-minute resolution---we proceed in two stages.
First, we pretrain \genco on synthetic HQ1200 operating points generated with \datakit (\cref{subsec:comparison_data_hq}), and compare \genco's performance against GRIT-HQ and classical AC/DC power flow solvers (\cref{subsubsec:hq_synthetic}).
Second, we fine-tune the pretrained \genco model on varying amounts of SCADA scenarios, and compare against \genco trained from scratch, on the same SCADA splits (\cref{subsubsec:hq_scada}).

\subsubsection{Pretraining on synthetic data}
\label{subsubsec:hq_synthetic}

First, we compare \genco Tiny against Hydro-Qu\'ebec Research Institute's GRIT-HQ model. Unlike the more general-purpose \genco architecture, GRIT-HQ is specialized for power flow on bus-level graphs. It is based on the GRIT architecture~\cite{ma2023graph}, uses random-walk structural encodings, and scales linearly with graph size; details are given in \cref{appendix:hqgrit}. Both models are trained on 200{,}000 synthetic HQ1200 samples (80/10/10 train/validation/test), with three independent seeds.\\

Mean active and reactive power-balance residuals are reported in \cref{tab:hq1200_results}. GRIT-HQ exhibits higher active power-balance residuals than DC-PF, whereas \genco reduces active power-balance residuals by approximately $2.5\times$ relative to DC-PF and substantially improves reactive power-balance residuals compared with GRIT-HQ. The resulting residual magnitudes of \genco are small compared to balancing and remedial actions routinely deployed by Hydro-Québec~\cite{EU2017_2195,EU2017_1485_SOGL}. This result is also consistent with IEEE grids of comparable size (\cref{fig:gridsize_vs_residuals_pf}), indicating robustness to a markedly different topology without architectural changes.

\begin{table}\centering\small
\resizebox{0.8\columnwidth}{!}{%
\renewcommand{\arraystretch}{1.2}
\begin{tabular}{c|c|c} \hline
    \multicolumn{1}{c|}{\textbf{Model}} &
    \multicolumn{2}{c}{\textbf{Power Balance Residual}} \\ \hline
    &
    \textbf{Active [MW]} &
    \textbf{Reactive [MVar]} \\ \hline
    \textbf{\genco Tiny}
    & \textbf{1.1 \scriptsize $\pm$ 0.3}
    & \textbf{0.5 \scriptsize $\pm$ 0.2} \\
    GRIT-HQ
    & 3.87 \scriptsize $\pm$ 0.02
    & 4.03 \scriptsize $\pm$ 0.44 \\
    DC-PF
    & 2.89
    & --- \\
    \rowcolor[gray]{0.92}
    AC-PF
    & 2.94e-9
    & 1.81e-9 \\ \hline
\end{tabular}}
\caption{Mean active and reactive power-balance residuals on the synthetic HQ1200 dataset.
Errors denote standard deviation over three seeds.
Bold values indicate the best learning-based method.}
\label{tab:hq1200_results}
\end{table}

\subsubsection{Fine-tuning on real SCADA data}
\label{subsubsec:hq_scada}

Next, we explore whether \genco can perform well on real HQ data under limited training on SCADA data.
Starting from the HQ1200 pretrained checkpoint of \cref{subsubsec:hq_synthetic}, we fine-tune on real SCADA scenarios using $N\in\{100,10^3,10^4,1.5\times10^4\}$ labeled training samples, and compare against \genco solvers trained from scratch on the same training samples. For the evaluation, we use a held-out set of 1{,}702 SCADA samples. In addition, we report the zero-shot performance of the pretrained \genco model, without SCADA fine-tuning, as well as the performance of DC-PF on the same evaluation set. All settings are repeated over three seeds; we report mean active power-balance residuals with sample standard deviation.\\

\begin{figure}
\centering
\includegraphics[width=0.85\columnwidth]{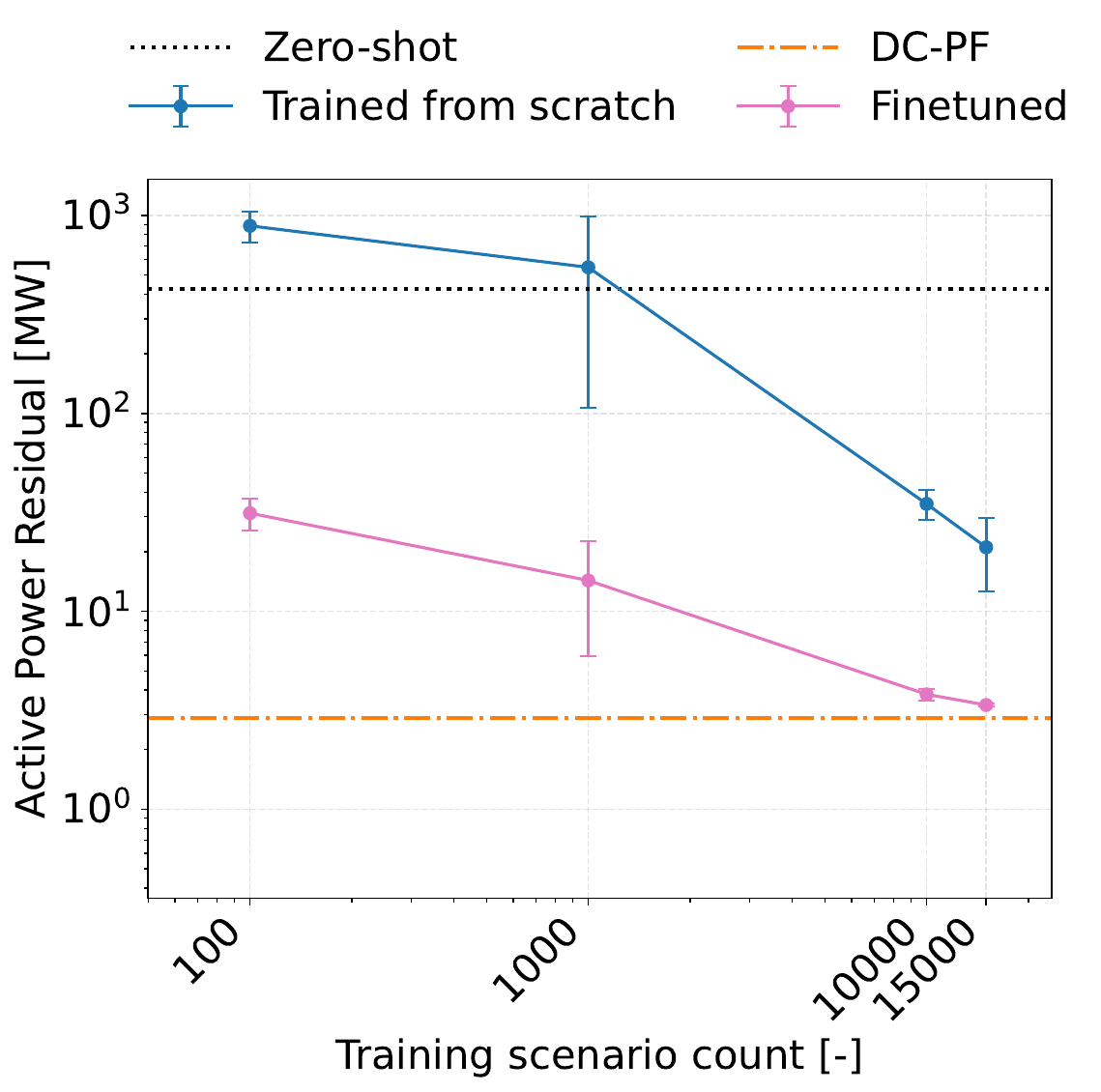}
\caption{Mean active power-balance residuals on HQ SCADA versus number of training scenarios, comparing fine-tuning of the synthetically pretrained model against training from scratch (mean $\pm$ sample std over three seeds).
The dotted line denotes zero-shot performance of the pretrained model.}
\label{fig:hq_scada_finetune}
\end{figure}

\cref{fig:hq_scada_finetune} shows active power-balance residuals versus the number of SCADA training scenarios. Zero-shot transfer remains poor, indicating that synthetic pretraining alone is insufficient for the model to perform well on real SCADA data. Fine-tuning rapidly closes this gap: with 15,000 training samples, fine-tuning achieves $3.36\pm0.05$~MW, an order of magnitude below training from scratch ($21.1\pm8.4$~MW) and approaching the DC-PF residual of $2.90$~MW on the same evaluation set. We note that the continued improvement with increasing amounts of SCADA samples also suggests that further real-data fine-tuning could close the remaining gap. Training from scratch also exhibits substantially higher variance at low sample counts. These results show that synthetic pretraining on HQ1200 enables data-efficient adaptation of \genco to real SCADA measurements.\\

\begin{boxH}
\paragraph{Takeaways.}
\begin{itemize}
\item \genco successfully transfers to a real grid and is evaluated on real SCADA data: after fine-tuning on 15,000 Hydro-Québec SCADA samples, it achieves a mean active power-balance residual close to that of DC-PF ($3.36$~MW vs. $2.90$~MW).
\item Pretraining on synthetic HQ1200 scenarios generated with \datakit enables substantially more data-efficient adaptation to real SCADA measurements, reducing the residual by an order of magnitude relative to training from scratch with the same 15,000 SCADA samples.
\end{itemize}
\end{boxH}

\section{Discussion}
\label{sec:discussion}

In this section, we offer perspectives on \genco's performance and discuss its practical deployment. Future research directions for \genco and the GridFM Development Framework are provided in \cref{appendix:future_work}.

\begin{figure}
    \centering
    \includegraphics[width=1.0\linewidth]{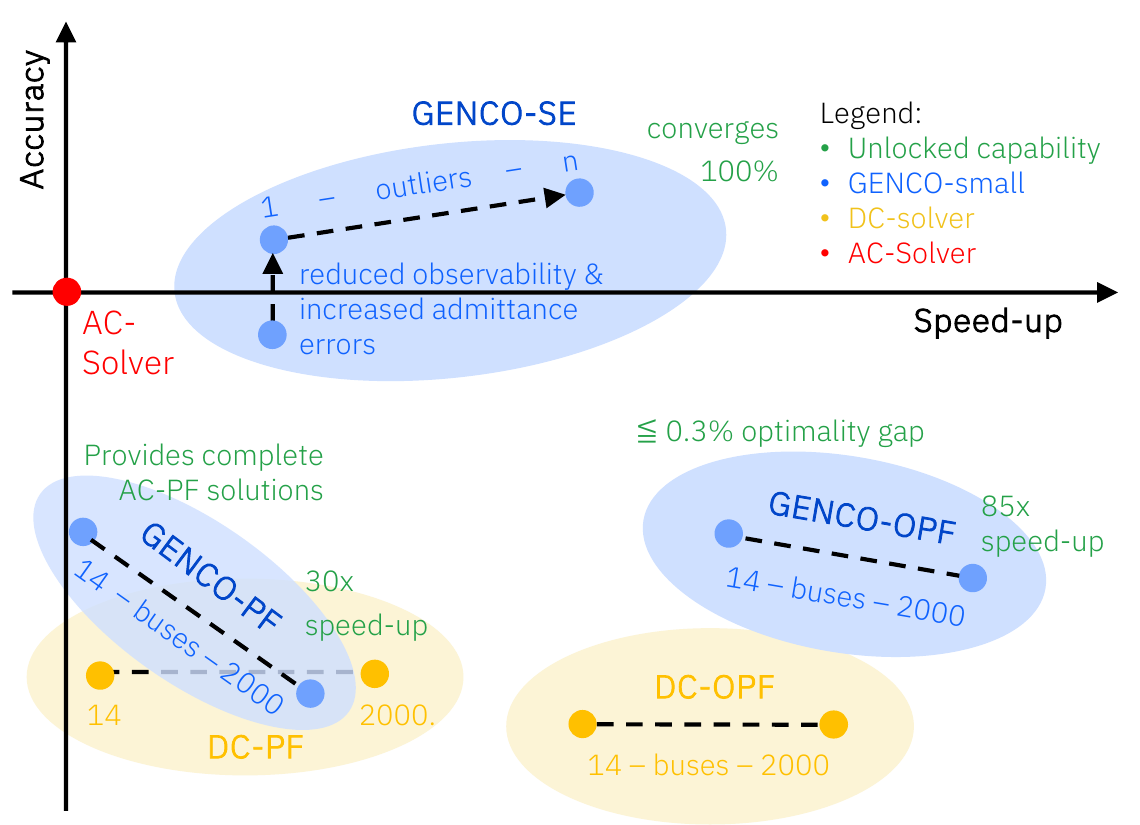}
    \caption{Semi-quantitative performance summary comparing \genco and classical DC solvers relative to classical AC solvers. Specific \genco capabilities are highlighted explicitly.}
    \label{fig:summary_plot}
\end{figure}

\paragraph{Performance of \genco.}
\genco's performance results are summarized in \cref{fig:summary_plot}, which illustrates semi-quantitatively the computational speed and solution accuracy of DC solvers and \genco relative to classical AC solvers. 
\genco offers a middle ground: In contrast to AC solvers and DC approximations, \genco sustains high throughput while recovering the full set of grid variables, including voltage magnitudes and reactive power.  
For PF on grids with $\geq 2000$ bus nodes, runtime improvements of \maxpfspeedups w.r.t. AC-PF at residuals comparable to DC solvers (\cref{tab:pf_selected_genco_tradeoff}) were achieved. For OPF, the speedups are larger, reaching \maxopfspeedups for case~2000 w.r.t. AC-OPF (\cref{fig:gridsize_runtime_opf}) at optimality gaps $\leq 0.3\%$, and feasibility within 0.35\% of the mean bounds on the OPFData benchmark (\cref{tab:opf_results}).
For SE, \genco can even outperform the weighted least squares (WLS) method (\cref{fig:perturbation}). Its runtime advantage grows with the number of outliers, as classical SE requires additional iterations to process them. Further, \genco always returns a high-quality estimate even when WLS fails to converge. 
Compared to task-specific machine learning models, \genco reached state-of-the-art performance across all three steady-state grid tasks, at a fraction of development time and costs, due to the unified architecture acting as a common backbone, sharing model and learning parameters, so hyperparameter optimization is required only once, rather than repeated for every task. Further task-specific optimization remains possible but was not required to achieve competitive performance.

\paragraph{Deployment of \genco.}
Our validation of \genco-PF on the real-world HQ1200 grid topology (\cref{tab:hq1200_results}) and SCADA data (\cref{fig:hq_scada_finetune}) indicates that pretraining \genco on synthetic \datakit data, followed by fine-tuning on approximately 15k data points (one year of SCADA data at 30-minute intervals), is important for achieving similar accuracy to DC solvers. We also showed that pretraining on a large number of different grids allows reaching similar accuracy as DC-PF with very few samples (\cref{subsubsec:transfer}). Further, we envision a safeguarded hybrid workflow in which \genco rapidly evaluates all scenarios, while power-balance residuals and violation checks identify critical cases for re-evaluation with a classical AC solver. With appropriate screening criteria, this substantially increases throughput while preserving numerical reliability, as most scenarios avoid a full AC solve.

\paragraph{Standardized development and benchmarking.}
With the release of the GridFM Development Framework---\datakit, \graphkit, and the accompanying datasets (\cref{fig:motivational_example})---we establish an end-to-end pipeline for reproducible neural solver development and benchmarking across the three steady-state grid analysis tasks. A new model only needs to be implemented as a single class in \graphkit to take advantage of the shared data-loading, preprocessing, training, and evaluation routines. The widely used \pfdelta and OPFData benchmarks enable baseline comparison against state-of-the-art models, whereas \datakit data allows robustness assessments with user-defined perturbations including OPF scenarios with variable generator costs and topology variants beyond N-2. In addition, robustness assessments of neural solvers require more than reporting only mean or median errors. Instead, the full error distribution must be reported, as supported by \graphkit. 

\paragraph{Fair runtime analysis methodology.}
In \cref{subsec:scaling}, we proposed a protocol for a fair comparison of neural solvers against one another and against classical solvers, based on batched inference with fully utilized GPUs and CPUs of the same hardware generation. To our knowledge, all prior studies compare GPU inference against a single classical-solver thread on one core, which does not reflect the capacity of modern multi-core processors and the typical batch processing in power systems. For runtime evaluation, the model must first be integrated into \graphkit. The user can then specify whether data loading from disk should be included in the reported runtime or isolated from the solver execution. The benchmarking script automatically initializes the classical and neural solvers, determines suitable worker counts and batch sizes, executes the benchmark runs, and reports the resulting runtime metrics.

\section{Conclusion}
\label{sec:conclusion}

Our unified neural solver and open-source GridFM Development Framework represent two technical contributions: (i) an end-to-end workflow that enables fair and reproducible neural solver development and benchmarking, and (ii) \genco, a state-of-the-art neural solver that unifies the three fundamental steady-state grid analysis tasks within a single shared representation. Together, these contributions enable the power-system-AI community to compare models on a common foundation and accelerate innovation.\\ 

These contributions directly address the four key challenges for neural solvers identified in \cref{sec:related}:\\

First, we responded to the challenge of a \textit{unified grid representation and architecture} through a shared graph representation in which task-specific differences in the input features are covered by masking: irrelevant variables are hidden, while unknown and decision variables are exposed for inference, yielding a fixed input representation across tasks. The heterogeneous graph structure, neural architecture, normalization scheme, and AC-physics enforcement remain unchanged, with only lightweight, non-learnable task-specific physics decoders added. Consequently, the same model and hyperparameters serve PF, OPF, and SE, enabling direct transfer of architectural improvements and tuning effort while reducing the time and compute cost required for model training.\\

Second, \genco addresses the \textit{runtime--feasibility--optimality--completeness trade-off} through physics-based solution completion and iterative correction with power-balance feedback that scales linearly with grid size, while preserving efficient batched inference. This places \genco between classical AC and DC solvers: it approaches the speed of DC methods while recovering the full set of AC variables, enabling fast approximate AC solutions. On large grids, \genco achieves up to \maxpfspeedups speedups over Newton--Raphson for PF while attaining residual levels comparable to DC-PF. For OPF, it provides up to \maxopfspeedups speedups over IPOPT while maintaining an optimality gap of $\leq 0.3\%$ on the OPFData benchmark. For SE, it consistently returns high-quality estimates, including in cases where WLS fails to converge.\\

Third, \genco's \textit{generalization} is evaluated with \datakit, which extends existing sampling methods with realistic load profiles, higher-order topology perturbations beyond N-2, and admittance variations, adding out-of-limit operating points for PF, generator cost variations for OPF, and sparse, noisy, and mismatched measurements for SE. \genco generalizes to contingencies up to N-20 with up to $3\times$ lower median residuals than DC-PF, and remains robust to out-of-operating-limit scenarios, outperforming DC-PF on line-loading prediction while also detecting voltage violations. Most importantly, we assess \genco on the real-world HQ1200 grid topology and its SCADA-derived states, achieving similar performance to DC-PF. Topology-agnostic zero-shot transfer remains unsolved: the pretrained \genco requires fine-tuning on roughly $1{,}000$ grid-specific samples to outperform DC-PF on a new grid.\\

Finally, we address \textit{workflow and evaluation fragmentation} along three axes: (i) we enable harmonized comparisons across datasets by evaluating \genco on \pfdelta and OPFData and providing converters that map existing benchmarks into \datakit, while releasing standardized PF, OPF, and SE datasets on Hugging Face; (ii) we introduce a method for fair runtime benchmarking on fully utilized CPUs and GPUs, based on amortized accounting for hardware and parallelization effects; and (iii) we release the Linux Foundation OpenGridFM libraries (\datakit and \graphkit) as a shared backbone that enables model development and utilization across PF, OPF, and SE tasks, in a low-code environment. 

\begin{table}
\centering
\small
\caption{Author contributions.}
\label{tab:contributions}
\begin{tabularx}{\linewidth}{>{\bfseries}l X}
\hline
Contribution & Authors \\
\hline
Conceptualization &
Alban Puech, Matteo Mazzonelli, Jonas Weiss, François Mirallès, Hendrik F.~Hamann, Etienne Vos, Thomas Brunschwiler \\
\hline
Experiments &
Alban Puech, Matteo Mazzonelli, Tamara R.~Govindasamy, Mangaliso Mngomezulu, Héctor Maeso García, Thomas Tolhurst, Javad Bayazi, Ali Moeini, Naomi Simumba, David Nelischer, Etienne Vos, Thomas Brunschwiler \\
\hline
Software Engineering &
Thomas Tolhurst, Javad Bayazi, Celia Cintas, Romeo Kienzler \\
\hline
Writing &
Alban Puech, Tamara R.~Govindasamy, Mangaliso Mngomezulu, François Mirallès, Hendrik F.~Hamann, Etienne Vos, Thomas Brunschwiler \\
\hline
Editing \& Reviewing &
Alban Puech, Florian Dörfler, Gabriela Hug, Martin Mevissen, Juan Bernabé-Moreno, François Mirallès, Hendrik F.~Hamann, Etienne Vos, Thomas Brunschwiler \\
\hline
Supervision &
Alban Puech, Jonas Weiss, Anna Varbella, Florian Dörfler, Gabriela Hug, Etienne Vos, Thomas Brunschwiler \\
\hline
Project Management &
Alban Puech, Jonas Weiss, Martin Mevissen, Juan Bernabé-Moreno, François Mirallès, Hendrik F.~Hamann, Etienne Vos, Thomas Brunschwiler \\
\hline
Sponsoring &
Juan Bernabé-Moreno, François Mirallès, Hendrik F.~Hamann \\
\hline
\end{tabularx}
\end{table}

\section*{Acknowledgments}

This work was partially supported by the U.S. Department of Energy’s Office of Critical Minerals and Energy Innovation (CMEI), Integrated Energy Systems Office (IESO), under the project titled “Artificial Intelligence for Energy Dominance,” Award Number 54805. We thank Panagiotis Grontas for proofreading the paper, and Mohammad Atif, Guang Zhao, Benedikt Wahl, Riccardo Ghetti, Srihith Bharadwaj Burra, Tilman Bockhacker, Olayiwola Arowolo, and Antonio Alcántara for their feedback and contributions to \graphkit and \datakit. We also thank Matteo Baù, Marcus Freitag, Johannes Schmude, Le Xie, and Thomas Theis for their valuable comments, insights, and discussions. Finally, the authors acknowledge valuable discussions with the GridFM community.

\section*{Author Contributions}
Author contributions are summarized in \cref{tab:contributions}.

\section*{Declaration of Interests}
The authors declare no competing interests.

\section*{Declaration of Generative AI and AI-Assisted Technologies}
During the preparation of this work, the authors used ChatGPT and Claude to find more concise reformulations and to condense paragraphs. After using these tools, the authors reviewed and edited the content as needed and take full responsibility for the content of the publication.

{\small
\bibliographystyle{ieeenat_fullname}
\bibliography{11_references}
}

\onecolumn
\appendix

\section{\textit{Supplementary Material for \cref{subsec:pf}: Results -- Power Flow}}

\subsection{Additional Results on \pfdelta}

\begin{figure}[H]
    \centering
    \includegraphics[width=1\columnwidth]{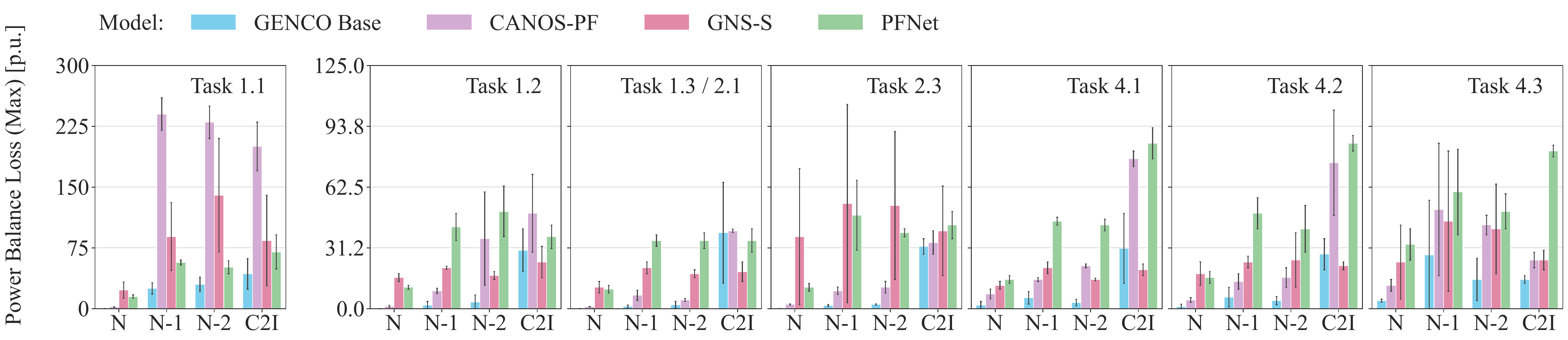}
\caption{Max power-balance loss (computed across all buses of all samples) for \genco, CANOS-PF, GNS-S, and PFNet across PF$\Delta$ tasks on the IEEE 118-bus system. Error bars in the figure correspond to the standard deviation over three seeds. C2I denotes close-to-infeasible cases at the steady-state stability limit, where classical solvers face challenges~\cite{NEURIPS2025_d000ef56}.}
    \label{fig:max-pb}
\end{figure}

\section{\textit{Supplementary Material for \cref{subsec:SE}: Results -- State Estimation}}
\label{app:se}

\begin{figure}[H]
    \centering
    \includegraphics[width=0.7\columnwidth]{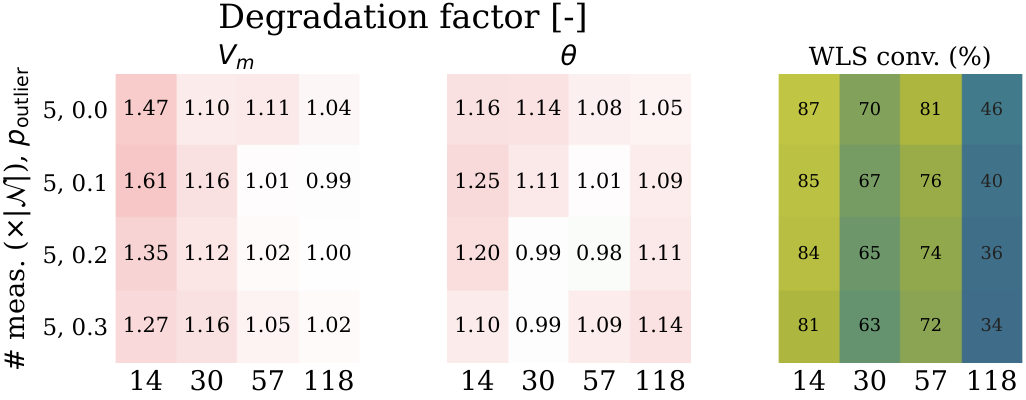}
    \caption{Degradation in scenarios where WLS did not converge (at a sparse coverage of $5|\mathcal{N}|$ measurements). The first two panels report, for $V_m$ and $\theta$, the ratio of \genco's error on scenarios where the baseline did not converge to its error on scenarios where it did (rows sweep the outlier probability $p_{\text{outlier}}$). Values near $1$ mean \genco is almost as accurate on the hard, non-converged cases as on the easy ones, i.e.\ it degrades gracefully rather than failing outright; values above $1$ indicate higher error on the hard set. The last panel shows the WLS convergence rate. Columns are the IEEE 14-, 30-, 57-, and 118-bus grids.}

    \label{fig:non-observable}
\end{figure}

\section{\textit{Supplementary Material for \cref{subsec:scaling}: Results -- Runtime and Performance Scaling}}
\label{app:runtime_details}

\subsection{Experimental Details}

\paragraph{Grids and sample counts.}
We evaluate small IEEE~14/30/57/118 and large GOC~500/2{,}000/10{,}000 grids. All solvers process the same number of instances per network, respectively:
\[
4\times10^{6},\;
3\times10^{6},\;
2\times10^{6},\;
2\times10^{6},\;
5\times10^{5},\;
5\times10^{4},\;
10^{4}.
\]
Each benchmark run at the selected optimal batch size lasts approximately 20\,s or longer, making initialization operations a small fraction of the measured runtime while keeping the full search for the best batch size or number of workers tractable (runs with suboptimal batch sizes or worker counts can take much longer). All runtime experiments consumed 37.5 H100 GPU-hours for \genco and 343 CPU-node-hours for the PowerModels-based classical solvers.

\paragraph{\genco.}
Inference uses FP32 GPU execution with \texttt{torch.compile(mode="reduce-overhead")}. Each job uses one exclusive H100 80\,GB GPU, 40 CPU cores, and 128\,GB of host RAM; this was sufficient because only $10^4$ samples are resident in memory at once. Batch sizes are powers of two from 64 to 16{,}384 on IEEE grids and from 16 to 16{,}384 on GOC grids. Inference uses 32 DataLoader workers.

\paragraph{Classical solvers implemented through PowerModels.}
AC-PF uses PowerModels' NLsolve-based Newton solver through GOC~500 and a more robust JuMP model on GOC~2{,}000/10{,}000. AC-OPF and DC-OPF use IPOPT with MUMPS; IPOPT uses tolerance $10^{-6}$ and \texttt{max\_iter}=100. Each job is assigned all 84 physical CPU cores of one node, with one Basic Linear Algebra Subprogram (BLAS) thread per worker. The worker sweep is $p\in\{24,40,\ldots,216\}$ (with increments of 16). Reproducing the experiments requires at least 256\,GB for small grids and 960\,GB for large grids. Dispatch batch size is 32 (workers get 32 scenarios to solve at once) on small grids and 1 on large grids to amortize scheduling overhead where solves are cheap without delaying dynamic load balancing on expensive cases.

\subsection{Batch-Size and Worker-Count Sweeps}
\label{appendix:sweep}

\begin{figure}[H]
    \centering
    \includegraphics[width=0.5\columnwidth]{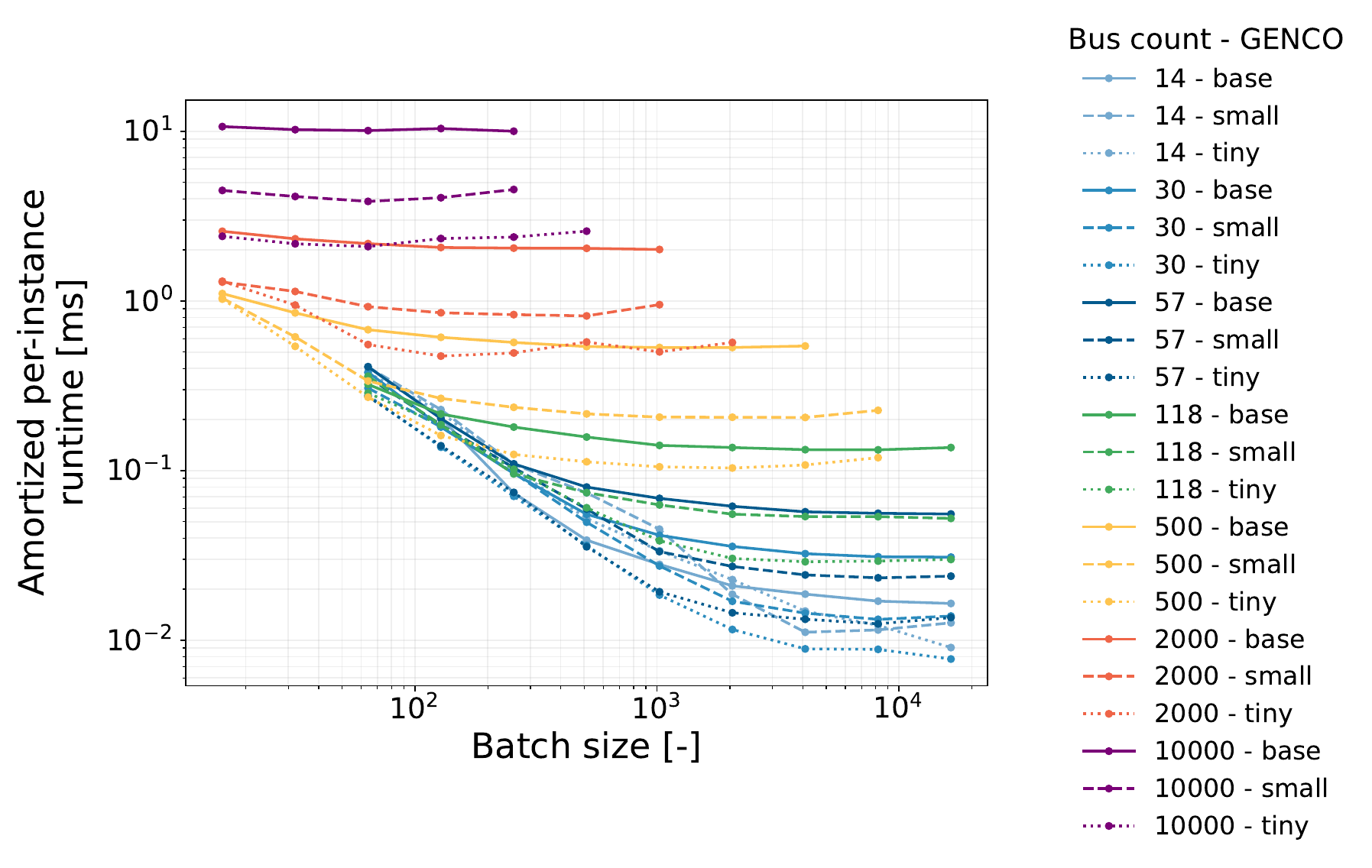}
    \caption{\genco wall-clock time per instance versus batch size across grids and model scales.}
    \label{fig:batchsize_vs_runtime_appendix}
\end{figure}

\begin{figure}[H]
    \centering
    \includegraphics[width=0.7\columnwidth]{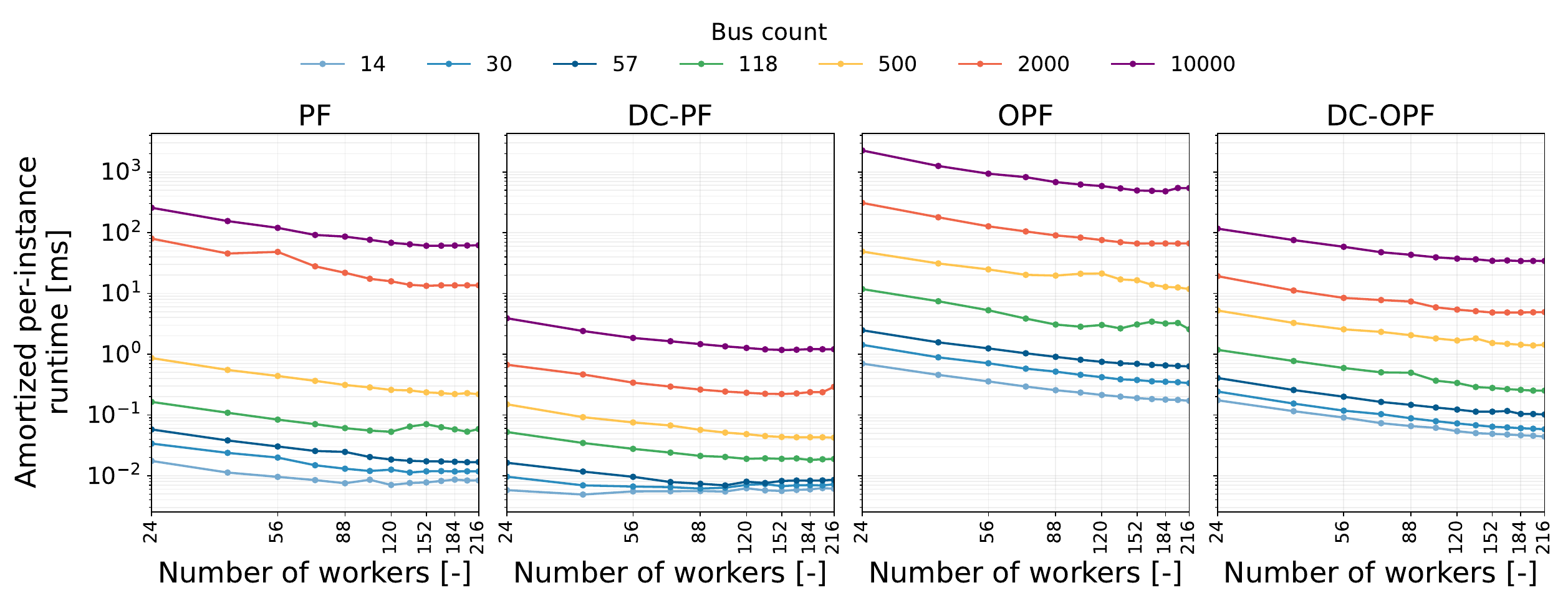}
    \caption{Wall-clock time per instance versus worker count for the PowerModels-based classical AC-PF, DC-PF, AC-OPF, and DC-OPF solvers across grids using the in-memory protocol.}
    \label{fig:workers_vs_runtime_appendix}
\end{figure}

\genco's wall-clock time per instance versus batch size for all grids and Base/Small/Tiny models is depicted in \cref{fig:batchsize_vs_runtime_appendix}. Wall-clock time per instance versus worker count for the PowerModels-based classical solvers is shown in \cref{fig:workers_vs_runtime_appendix}. Larger batches or worker pools reduce wall-clock time per instance until saturation; larger models and grids saturate earlier.

\subsection{Runtime Analysis With and Without Data Loading}
\label{app:s1_vs_s2}

\paragraph{Protocols.}

The main results in \cref{subsec:scaling} use an \emph{in-memory protocol} designed to represent repeated scenario analysis around a reference operating point. The \emph{from-disk protocol} complements this setup by considering grid snapshots in which each scenario is an independent input loaded from disk. The two protocols therefore capture two distinct use cases: repeated in-memory perturbations around a base operating point, and solving heterogeneous scenarios retrieved from storage.

\paragraph{Practical differences.}

In the \emph{in-memory protocol}, both solver classes use preprocessed, parsed, and memory-resident scenarios: GENCO cycles over $10^4$ already loaded tensors in RAM, while each PowerModels worker uses an already parsed base case. The number and content of the tensors do not affect solve runtime; the large pool is used only to avoid potential caching effects. In contrast, in the \emph{from-disk protocol}, both solver classes read scenarios from disk: GENCO reads each processed sample, while PowerModels reads and parses a ready-to-solve \datakit scenario. In both protocols, GENCO uses batched GPU inference and the PowerModels-based classical solvers use multiple CPU workers, with the same parallelization strategy used to maximize throughput. The only difference in timing scope between the two protocols is whether scenario data loading is included in the measured runtime.\\

The from-disk protocol additionally introduces practical dependencies that are absent from the in-memory protocol, including storage format (e.g., JSON versus other representations), scenario access order (e.g., sequential versus non-sequential access), page-cache state, filesystem type (e.g., local node storage versus GPFS), and system contention. Finally, the solved scenarios differ between the two protocols: the in-memory protocol repeatedly solves the fixed base case, whereas the from-disk protocol solves distinct \datakit scenarios. The use of distinct \datakit scenarios is intentional: the from-disk protocol is designed to represent workloads in which heterogeneous grid snapshots are independently retrieved and solved, rather than repeated perturbations around a single operating point. The resulting runtime difference therefore reflects both data-loading overhead and differences in scenario difficulty and should be interpreted as a comparison of two workload regimes.

\paragraph{Observed effects.}

Loading primarily affects \genco's small-grid speedups as shown in \cref{tab:loading_speedups}. Under the from-disk protocol, \genco is slower than AC-PF on IEEE~14/30/57, but remains faster than AC-OPF on every grid; the large-grid conclusions are unchanged.\\

\begin{table}[H]
\centering
\scriptsize
\setlength{\tabcolsep}{5pt}
\caption{\genco Tiny speedups relative to classical AC-PF and AC-OPF solvers. Speedup is classical solvers' wall-clock time per instance divided by \genco wall-clock time per instance; values above one favor \genco.}
\label{tab:loading_speedups}
\begin{tabular}{@{}l cc cc@{}}
\toprule
& \multicolumn{2}{c}{vs.\ AC-PF} & \multicolumn{2}{c}{vs.\ AC-OPF} \\
\cmidrule(lr){2-3}\cmidrule(lr){4-5}
Grid & In-memory & From-disk & In-memory & From-disk \\
\midrule
IEEE 14     & $0.8\times$  & $0.3\times$  & $18.9\times$  & $6.4\times$ \\
IEEE 30     & $1.5\times$  & $0.4\times$  & $43.0\times$  & $9.5\times$ \\
IEEE 57     & $1.3\times$  & $0.6\times$  & $50.4\times$  & $16.1\times$ \\
IEEE 118    & $1.8\times$  & $1.4\times$  & $88.4\times$  & $50.6\times$ \\
GOC 500     & $2.1\times$  & $2.6\times$  & $114.0\times$ & $108.9\times$ \\
GOC 2{,}000 & $28.2\times$ & $31.4\times$ & $140.0\times$ & $160.8\times$ \\
GOC 10{,}000 & $29.1\times$ & $28.9\times$ & $229.1\times$ & $243.0\times$ \\
\bottomrule
\end{tabular}
\end{table}

\begin{figure}[H]
    \centering
    \includegraphics[width=0.5\columnwidth]{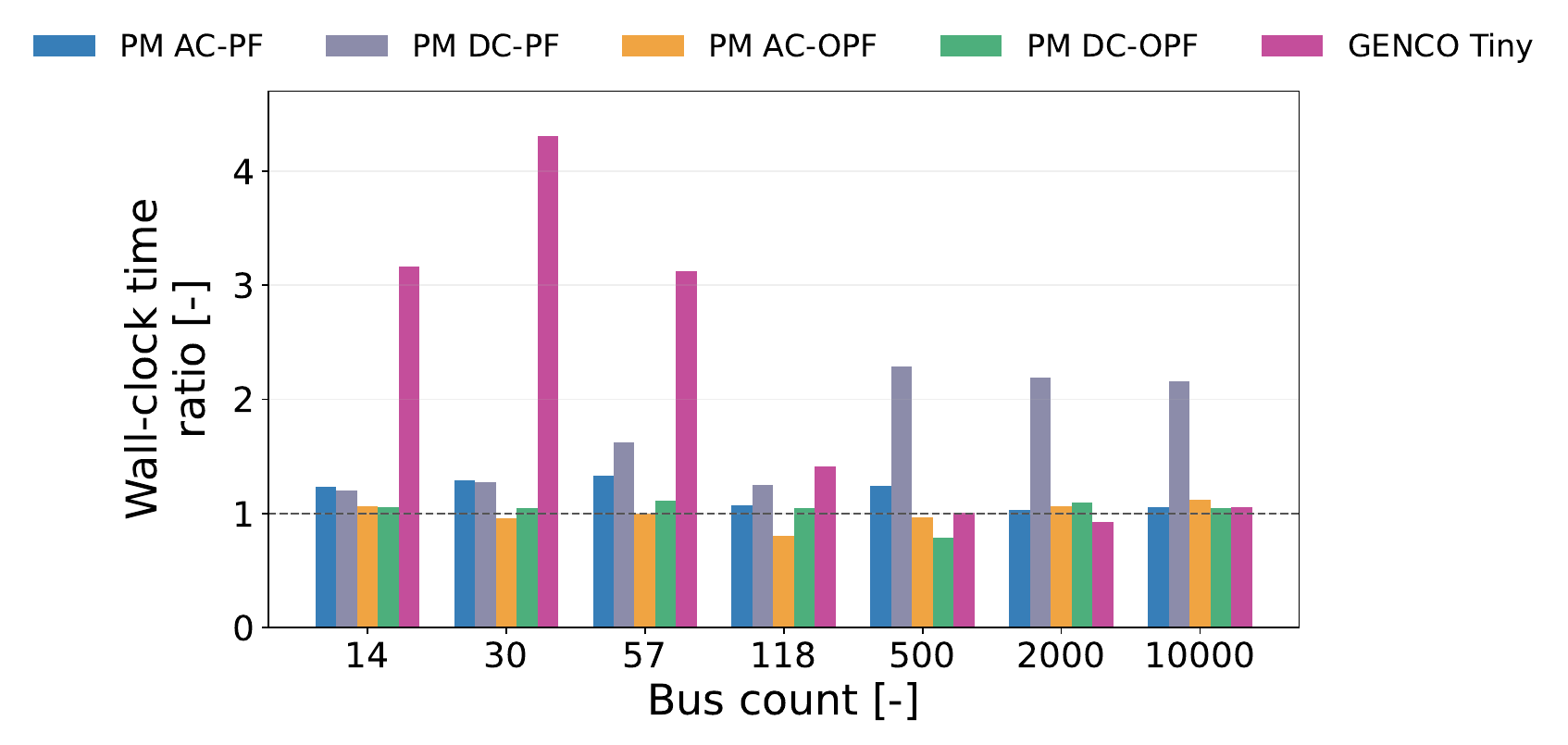}
    \caption{Ratio of from-disk to in-memory wall-clock time per instance at the best configuration selected independently under each protocol. The dashed line marks equal time; values above one indicate loading overhead.}
    \label{fig:loading_overhead_best_config}
\end{figure}

Loading matters most when computation is inexpensive and becomes negligible as model or grid complexity grows as depicted in \cref{fig:loading_overhead_best_config}. For PowerModels, loading affects DC-PF most on large grids because JSON parsing complexity grows faster with grid size than DC-PF solve complexity; AC-PF changes by only about $1.0$--$1.3\times$, while OPF remains near parity because optimization dominates parsing. Ratios below one coincide with lower mean solve times over diverse scenarios than for the repeated base case, indicating that the base load and topology are harder than the average scenario.

\subsection{Additional OPF scaling results}

Detailed results for \genco Base and Small from IEEE~14 through GOC~2{,}000 are provided in \cref{tab:opf_scaling}.
\begin{table}[H]
\caption{Constraint violations and optimality gaps for \genco and DC-OPF, with AC-OPF as the optimality reference. Bold entries denote metrics for which the performance difference exceeds one standard deviation. The metrics are computed as described in Section~\ref{subsec:opf}. \genco Base and Small achieve similar feasibility and optimality improvements across all systems.}
\label{tab:opf_scaling}
\centering
\footnotesize
\setlength{\tabcolsep}{4pt}
\begin{tabular}{@{}lllllclll@{}}
\toprule
& & Optimality & \multicolumn{2}{c}{Thermal limits} & \multicolumn{2}{c}{Power balance} & React.\ gen. bounds \\
\cmidrule(lr){3-3}\cmidrule(lr){4-5}\cmidrule(lr){6-7}\cmidrule(lr){8-8}
System & Model
  & Gap (\%)
  & $S_{ij}(+)$ [MVA]
  & $S_{ij}(-)$ [MVA]
  & $\mathrm{PBRes}_{P}$ [MW]
  & $\mathrm{PBRes}_{Q}$ [MVar]
  & $Q_g$ [MVar] \\
\midrule

\multirow{3}{*}{IEEE 14}
  & \genco Base
  & \textbf{0.10 $\pm$ 0.00}
  & 2.56e-5 $\pm$ 1.94e-5
  & 1.32e-5 $\pm$ 2.93e-6
  & \textbf{4.35e-3 $\pm$ 9.46e-5}
  & \textbf{3.79e-3 $\pm$ 1.11e-4}
  & 1.28e-3 $\pm$ 1.57e-4 \\

  & \genco Small
  & 0.10 $\pm$ 0.01
  & 2.46e-5 $\pm$ 1.83e-5
  & 1.37e-5 $\pm$ 2.37e-6
  & 5.51e-3 $\pm$ 5.28e-4
  & 4.63e-3 $\pm$ 6.25e-5
  & \textbf{1.20e-3 $\pm$ 2.44e-4} \\

  & DC-OPF
  & 5.75 $\pm$ 0.01
  & \textbf{0}
  & \textbf{0}
  & 1.42e+0 $\pm$ 4.56e-3
  & --
  & -- \\
\midrule

\multirow{3}{*}{IEEE 30}
  & \genco Base
  & \textbf{0.21 $\pm$ 0.02}
  & 1.13e-3 $\pm$ 1.74e-4
  & \textbf{1.11e-3 $\pm$ 1.55e-4}
  & \textbf{6.90e-3 $\pm$ 1.05e-3}
  & \textbf{4.95e-3 $\pm$ 2.93e-4}
  & \textbf{7.00e-4 $\pm$ 3.11e-4} \\

  & \genco Small
  & 0.25 $\pm$ 0.01
  & \textbf{1.12e-3 $\pm$ 1.94e-4}
  & 1.12e-3 $\pm$ 1.61e-4
  & 1.06e-2 $\pm$ 1.81e-3
  & 8.03e-3 $\pm$ 6.59e-4
  & 8.95e-4 $\pm$ 3.92e-5 \\

  & DC-OPF
  & 7.83 $\pm$ 0.03
  & 9.83e-2 $\pm$ 1.02e-3
  & 7.89e-2 $\pm$ 8.04e-4
  & 6.34e-1 $\pm$ 4.27e-3
  & --
  & -- \\
\midrule

\multirow{3}{*}{IEEE 57}
  & \genco Base
  & 1.21 $\pm$ 0.18
  & 2.89e-3 $\pm$ 3.64e-4
  & 3.04e-3 $\pm$ 6.22e-5
  & 4.13e-1 $\pm$ 1.59e-1
  & 1.80e-1 $\pm$ 7.74e-2
  & \textbf{6.15e-3 $\pm$ 7.11e-4} \\

  & \genco Small
  & \textbf{0.88 $\pm$ 0.14}
  & \textbf{2.69e-3 $\pm$ 5.02e-5}
  & \textbf{2.87e-3 $\pm$ 3.28e-4}
  & \textbf{3.00e-1 $\pm$ 6.92e-2}
  & \textbf{1.62e-1 $\pm$ 3.38e-2}
  & 7.25e-3 $\pm$ 1.28e-3 \\

  & DC-OPF
  & 5.54 $\pm$ 0.02
  & 1.53e-2 $\pm$ 3.36e-4
  & 1.15e-2 $\pm$ 2.26e-4
  & 1.29e+0 $\pm$ 2.51e-3
  & --
  & -- \\
\midrule

\multirow{3}{*}{IEEE 118}
  & \genco Base
  & 1.66 $\pm$ 0.03
  & \textbf{3.75e-2 $\pm$ 5.62e-3}
  & \textbf{3.74e-2 $\pm$ 5.78e-3}
  & 1.36e+0 $\pm$ 1.12e-1
  & 2.75e-1 $\pm$ 2.36e-3
  & 1.19e-3 $\pm$ 6.65e-4 \\

  & \genco Small
  & \textbf{1.66 $\pm$ 0.03}
  & 4.21e-2 $\pm$ 4.48e-5
  & 4.17e-2 $\pm$ 3.73e-4
  & \textbf{1.12e+0 $\pm$ 6.14e-2}
  & \textbf{2.57e-1 $\pm$ 1.49e-2}
  & \textbf{1.13e-3 $\pm$ 1.83e-4} \\

  & DC-OPF
  & 4.73 $\pm$ 0.02
  & 1.04e-1 $\pm$ 6.49e-4
  & 1.06e-1 $\pm$ 6.25e-4
  & 2.46e+0 $\pm$ 3.11e-3
  & --
  & -- \\
\midrule

\multirow{3}{*}{GOC 500}
  & \genco Base
  & \textbf{0.49 $\pm$ 0.08}
  & \textbf{3.83e-3 $\pm$ 3.57e-4}
  & \textbf{3.86e-3 $\pm$ 1.76e-4}
  & \textbf{7.07e-1 $\pm$ 8.16e-2}
  & \textbf{1.89e-1 $\pm$ 2.18e-2}
  & \textbf{4.90e-4 $\pm$ 1.27e-4} \\

  & \genco Small
  & 0.57 $\pm$ 0.01
  & 4.22e-3 $\pm$ 6.91e-6
  & 4.39e-3 $\pm$ 1.76e-4
  & 8.16e-1 $\pm$ 9.95e-3
  & 2.48e-1 $\pm$ 8.65e-3
  & 6.19e-4 $\pm$ 3.00e-5 \\

  & DC-OPF
  & 1.81 $\pm$ 0.01
  & 4.35e-2 $\pm$ 1.47e-4
  & 8.01e-2 $\pm$ 3.20e-5
  & 1.85e+0 $\pm$ 4.12e-3
  & --
  & -- \\
\midrule

\multirow{3}{*}{GOC 2000}
  & \genco Base
  & \textbf{1.53 $\pm$ 0.03}
  & 3.56e-2 $\pm$ 2.35e-2
  & 3.66e-2 $\pm$ 2.49e-2
  & 2.94e+0 $\pm$ 3.05e-1
  & \textbf{7.93e-1 $\pm$ 1.25e-1}
  & \textbf{5.53e-3 $\pm$ 1.56e-3} \\

  & \genco Small
  & 1.54 $\pm$ 0.00
  & \textbf{7.30e-3 $\pm$ 1.04e-3}
  & \textbf{7.56e-3 $\pm$ 7.94e-4}
  & 2.85e+0 $\pm$ 2.37e-1
  & 8.02e-1 $\pm$ 7.62e-2
  & 5.99e-3 $\pm$ 2.16e-3 \\

  & DC-OPF
  & 2.84 $\pm$ 0.02
  & 5.14e-2 $\pm$ 2.76e-4
  & 9.57e-2 $\pm$ 4.37e-4
  & \textbf{1.08e+0 $\pm$ 7.34e-3}
  & --
  & -- \\
\midrule

\bottomrule
\end{tabular}

\raggedright\footnotesize
$\pm$ denotes standard deviation over three seeds.
\end{table}

\section{\textit{Supplementary Material for \cref{subsec:generalization}: Results -- Robustness \& Generalization}}

\label{appendix:top}

\begin{figure}[H]
\centering

\begin{minipage}{0.4\linewidth}
\centering
\includegraphics[width=\linewidth]{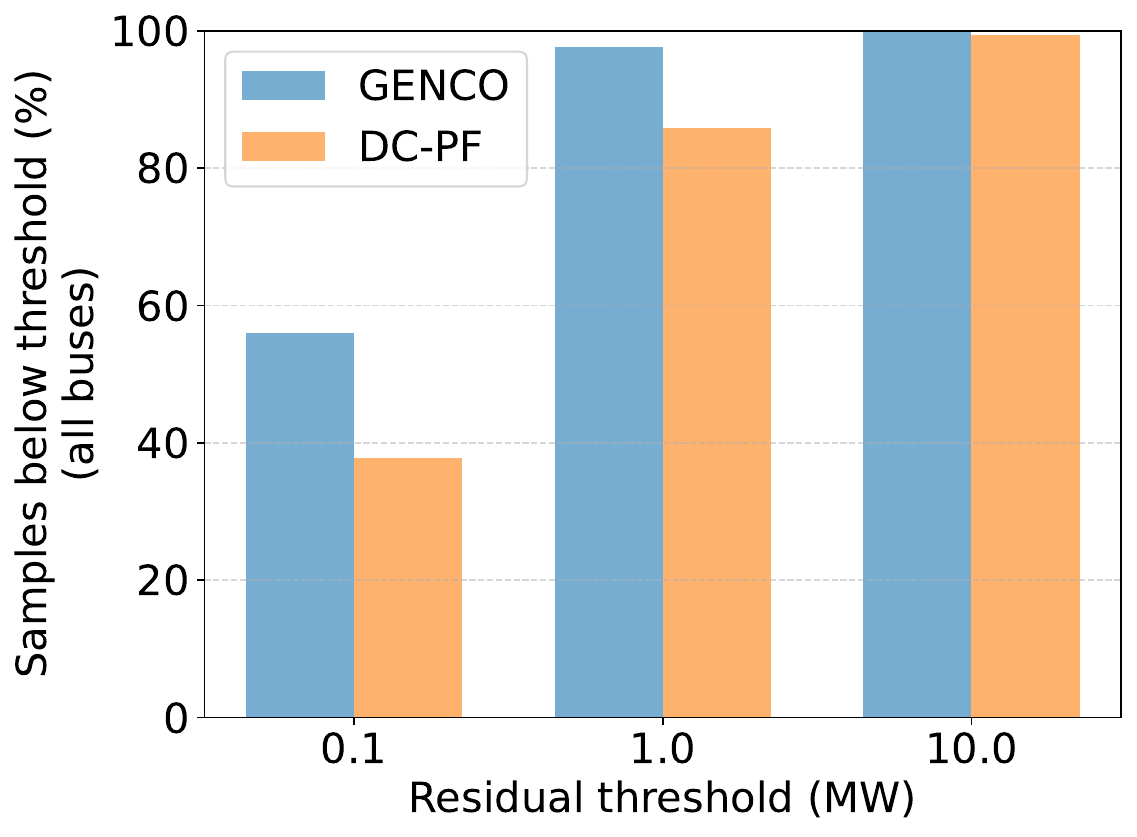}
\caption{Percentage of samples below absolute active power-balance residual thresholds at $k=10$ (all buses). \genco again dominates DC-PF across all thresholds.}
\label{fig:abs_threshold_k10}
\end{minipage}
\hspace{3cm}
\begin{minipage}{0.4\linewidth}
\centering
\includegraphics[width=\linewidth]{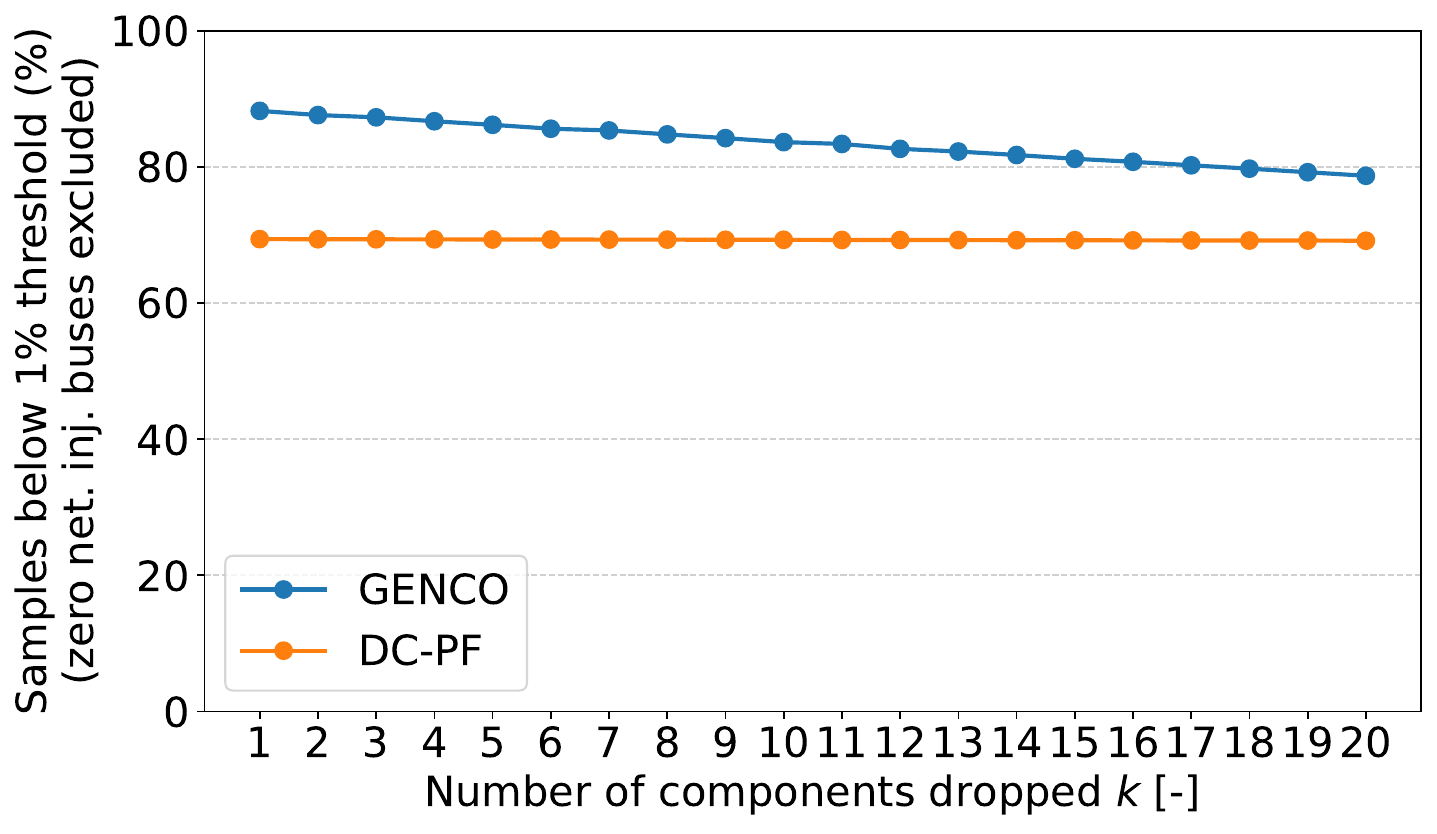}
\caption{Percentage of samples with relative active power-balance residual below 1\% as a function of contingency order $k$ (excluding zero-injection buses). The fraction decreases with $k$ for both methods, but remains consistently higher for \genco.}
\label{fig:rel_threshold_vs_k}
\end{minipage}

\end{figure}

The percentage of samples below
absolute active power-balance residual thresholds is shown in \cref{fig:abs_threshold_k10}. It is similar to \cref{fig:threshold_share_relative_k10} but uses absolute thresholds instead of relative ones, and includes buses with zero net injection.\\

The percentage of samples below 1\% residuals as a function of the contingency order is depicted in \cref{fig:rel_threshold_vs_k}.

\subsection{Residuals at Zero-Net-Injection Buses}
\label{appendix:zeroinj}

\begin{figure}[H]
\centering

\begin{minipage}{0.4\linewidth}
\centering
\includegraphics[width=\linewidth]{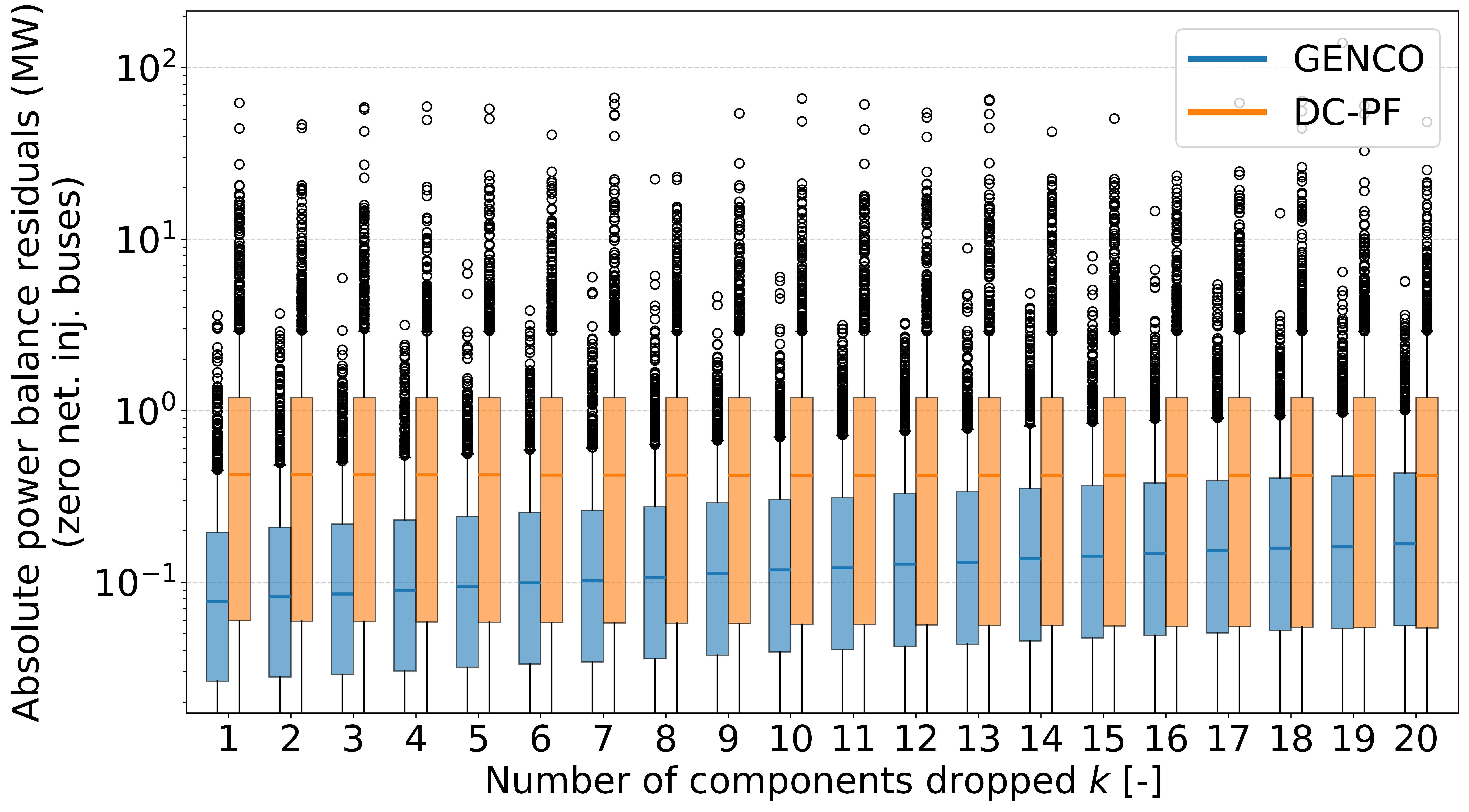}
\caption{Bus-level absolute active power-balance residuals at zero-net-injection buses under $N$-$k$ contingencies ($k=1,\dots,20$). \genco achieves consistently lower residuals than DC-PF across the entire distribution. Whiskers indicate 1.5$\times$IQR.}
\label{fig:zeroinj_boxplot}
\end{minipage}
\hspace{3cm}
\begin{minipage}{0.4\linewidth}
\centering
\includegraphics[width=\linewidth]{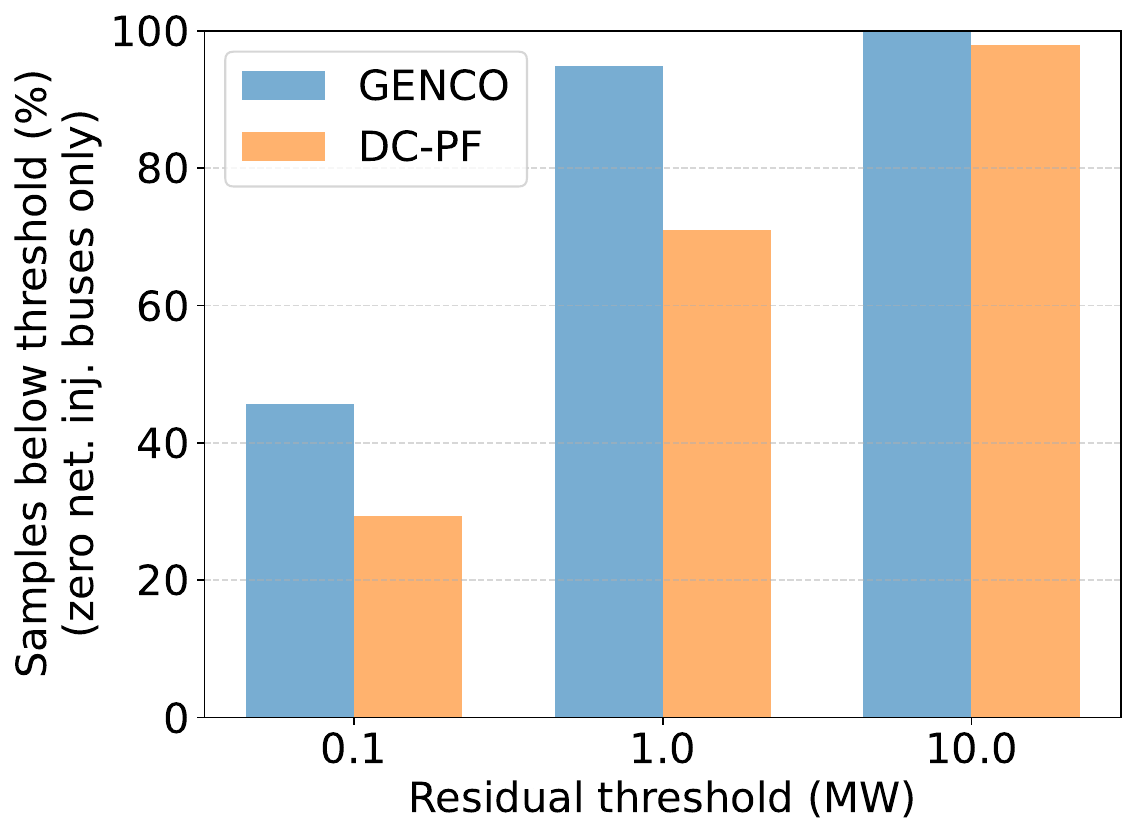}
\caption{Percentage of zero-net-injection buses below absolute active power-balance residual thresholds at $k=10$. \genco consistently achieves a higher fraction of low-residual predictions than DC-PF across all thresholds.}
\label{fig:zeroinj_abs_threshold_k10}
\end{minipage}

\end{figure}

The distribution of bus-level absolute active power-balance residuals at zero-net-injection buses for \genco and DC-PF across $N$-$k$ contingencies with $k=1,\dots,20$ is shown in \cref{fig:zeroinj_boxplot}. Unlike the relative metric in the main text, this view isolates buses for which no normalization by net injection is possible and directly reports the remaining power-balance mismatch in MW.

\cref{fig:zeroinj_abs_threshold_k10} complements this distributional view with the percentage of zero-net-injection buses below practically relevant absolute residual thresholds at $k=10$. 

\section{Voltage Prediction under Out-of-Limit Operating Conditions}
\label{appendix:voltage_violations}

\begin{figure}[H]
\centering

\begin{minipage}{0.4\linewidth}
\centering
\includegraphics[width=\linewidth]{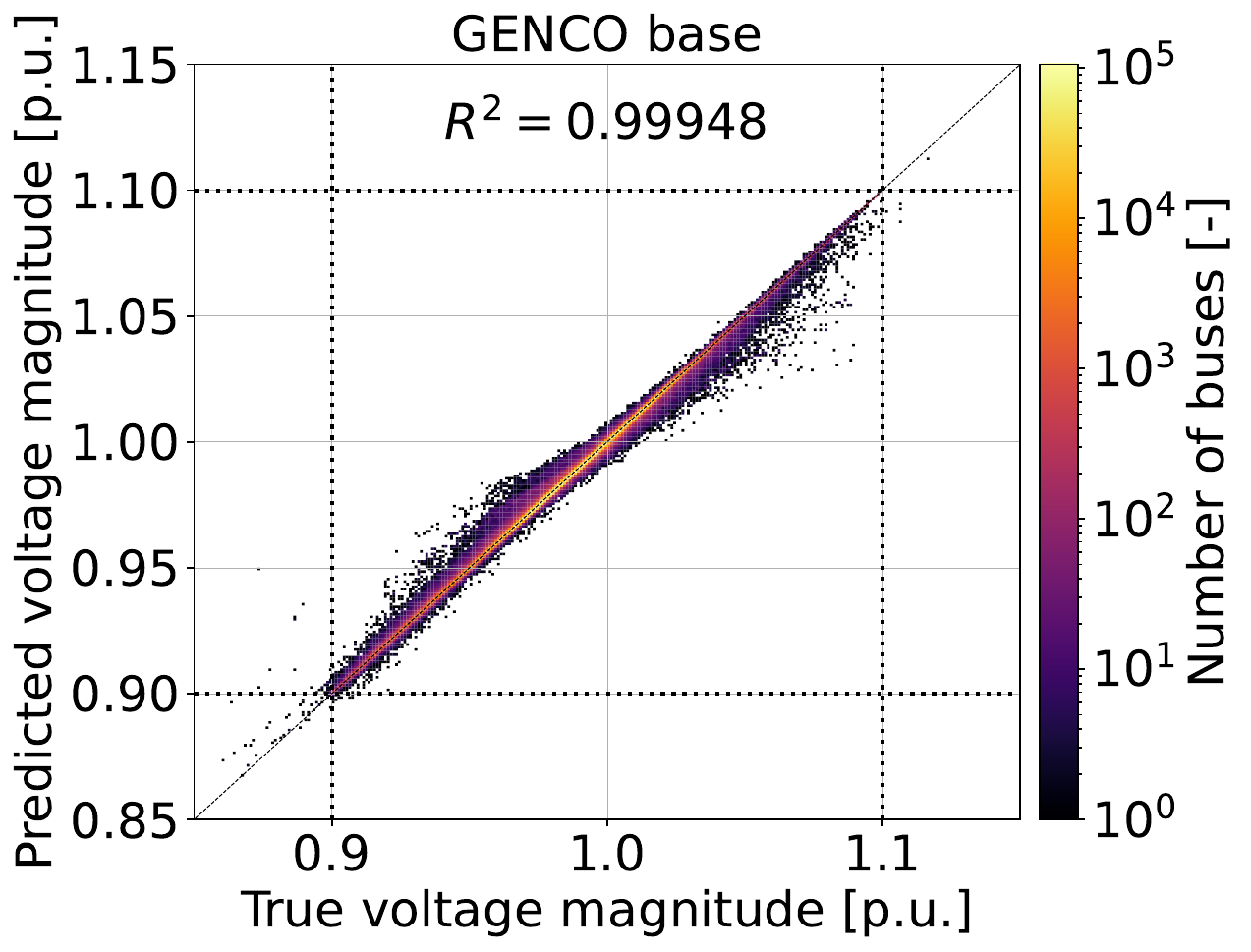}
\caption{Bus voltage predictions obtained with \genco compared to AC-PF voltages.}
\label{fig:genco_voltage}
\end{minipage}
\hspace{3cm}
\begin{minipage}{0.4\linewidth}
\centering
\includegraphics[width=\linewidth]{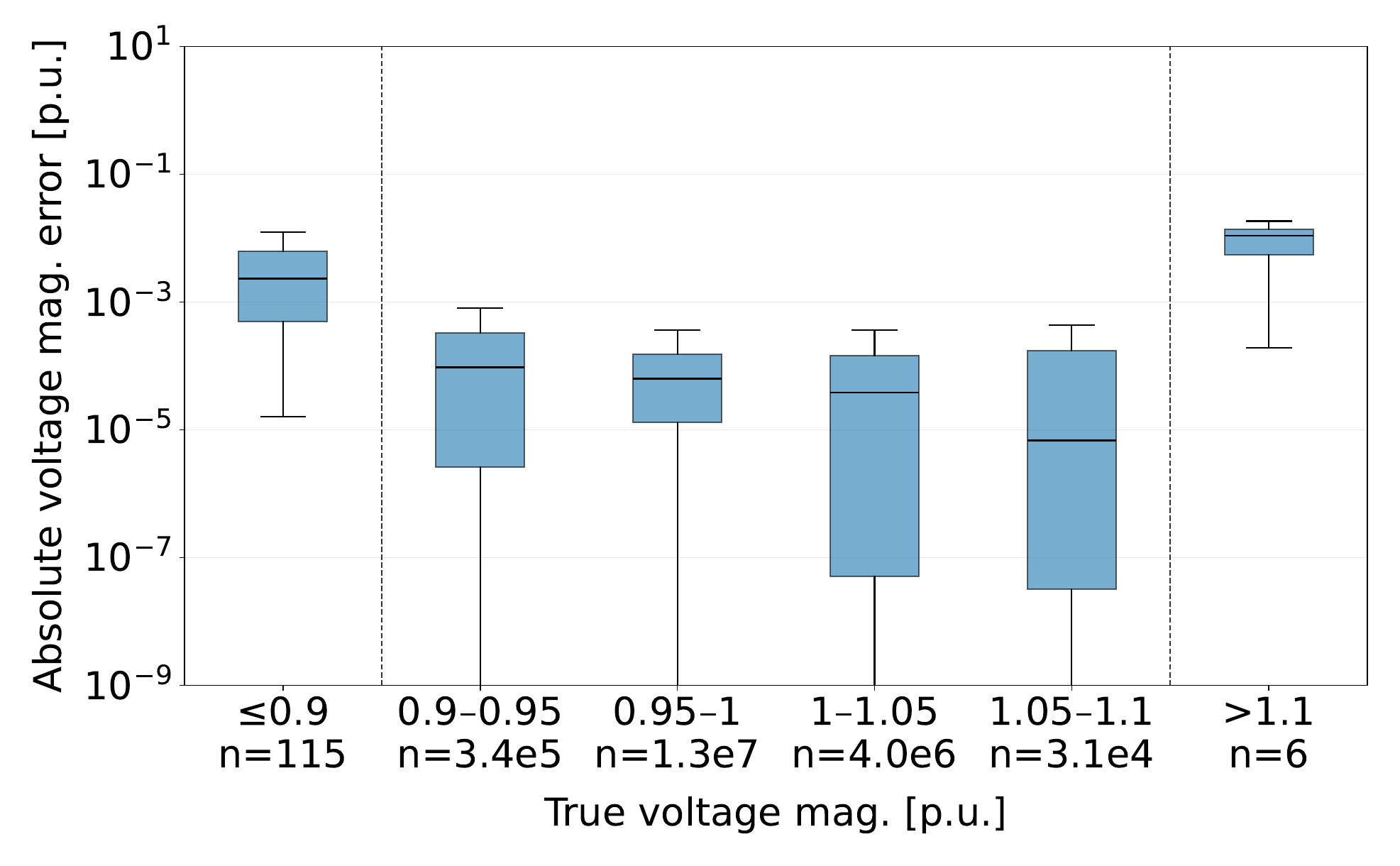}
\caption{Absolute voltage magnitude prediction error as a function of the true voltage magnitude. $n$ indicates the number of test samples in each bin.}
\label{fig:voltage_boxplot}
\end{minipage}

\end{figure}

Voltage violations are rare in the N-2 dataset, with only $6.9 \times 10^{-4}$\% of buses operating outside the nominal voltage range, i.e., 121 buses out of the 17,424,000 buses in the dataset. \cref{fig:genco_voltage} compares \genco voltage magnitude predictions against AC-PF across the full operating range, including undervoltage ($<0.9$~p.u.) and overvoltage ($>1.1$~p.u.) conditions. \genco remains closely aligned with the reference solution even in these rarely observed regimes.\\

Absolute voltage magnitude errors as a function of the true voltage magnitude are depicted in \cref{fig:voltage_boxplot}. Across all voltage ranges, including out-of-limit conditions, \genco maintains errors below 0.015~p.u. These results suggest that the model generalizes well to voltage violations despite their extreme scarcity in the training data, and does not require dedicated oversampling of such scenarios.

\section{\textit{Supplementary Material for \cref{subsec:hq1200}: Results -- Validation on Real Data from the Hydro-Québec Grid}}
\label{appendix:hqgrit}

\paragraph{Additional details about GRIT-HQ.} GRIT-HQ builds on the GRIT architecture~\cite{ma2023graph}, which combines random-walk-based positional encodings, attention layers that jointly update node and edge representations, and explicit incorporation of node degree information at each layer. We adapt the original model based on a large-scale empirical study over 600,000 scenarios with graph sizes ranging from 24 to 240 nodes, which guided the choice of depth, width, and attention configuration. In this setting, we observe that relative random walk positional encodings (RRWP) provide negligible gains over random walk structural encodings (RWSE). We therefore adopt RWSE in the final model, which reduces memory complexity while preserving linear scaling with graph size. The resulting architecture operates on a homogeneous bus-level graph representation, consistent with the original GRIT formulation.

\section{\textit{Supplementary Material for \cref{sec:discussion}: Discussion}}
\label{appendix:future_work}

\subsection{Future Work to Advance GENCO's Performance}
To further strengthen \genco's utility, scalable long-range message-passing and attention mechanisms must be explored, and all OPF constraints must be incorporated into the loss at each correction step. For expansion-planning use cases, robustness to additional grid components and to load and generation increases must also be analyzed. All of these perturbations will be included in the next version of \datakit.

\subsection{Future Work to Advance GENCO's Deployment}
In \cref{subsec:comparison_data_hq}, we successfully validated the realism of synthetic data generated with \datakit against SCADA data from HQ1200 based on the normalized mean feature entropy of the entire grid. In the future, the analysis will be extended to a finer level of granularity, as initial observations from HQ1200 indicate clustered topology changes rather than the uniformly distributed perturbations currently generated by \datakit. Further, analyzing model uncertainty conditioned on specific perturbations opens the door to uncertainty bounds based on conformal prediction, as demonstrated in~\cite{alcantara2026trustworthinesslayerfoundationmodels}.\\

In \cref{subsec:generalization}, we demonstrated \genco's robustness to contingencies of up to $N$-$20$ and out-of-operating-limit scenarios. However, for the most demanding case of entirely unseen grids, even our subgrid-based pre-training strategy (\cref{subsubsec:transfer}) achieved moderate performance, with zero-shot residuals exceeding those of DC solvers. Accordingly, improved training strategies that yield acceptable zero-shot performance will be explored to eliminate the need for model fine-tuning.

\subsection{Future Work to Advance Standardized Development and Benchmarking}
So far, we have benchmarked \genco's runtime against classical solvers and its accuracy against other neural solvers on \pfdelta and OPFData with limited perturbation diversity. In the future, third-party neural solvers must therefore be onboarded to \graphkit and evaluated on our diverse synthetic \datakit benchmark using the hardware configuration reported in this work. \\

Moreover, residuals and violations must be reported using relative metrics rather than absolute quantities such as MW and MVar. For example, power balance errors must be normalized by the corresponding net injections, and branch flow violations by their thermal limits, as absolute errors can have very different implications depending on the scale of the underlying grid element. Finally, reporting violation rates across multiple thresholds and full error distributions, rather than only averages, will enable more meaningful robustness comparisons.

\section{Code \& Reproducibility}

\begin{itemize}
    \item \genco and the OpenGridFM libraries (\graphkit and \datakit) continue to evolve beyond this paper; their latest versions are available on \href{https://github.com/gridfm}{GitHub}.
    \item Step-by-step guides for reproducing the PF and OPF experiments, either from scratch or from released checkpoints, are available in the \href{https://github.com/gridfm/gridfm-paper-genco}{paper repository}. The corresponding SE materials are being prepared and will follow.
    \item All \genco models used for the PF and OPF results in this paper are publicly available on \href{https://huggingface.co/gridfm}{Hugging Face}, including models trained on \pfdelta, OPFData, and \datakit datasets.
\end{itemize}


\end{document}